\PassOptionsToPackage{sort&compress}{natbib}

\renewcommand{\raggedright}{}
\documentclass[preprint,1pt]{elsarticle}

\usepackage{graphicx}%
\usepackage{multirow}%
\usepackage{amsmath,amssymb,amsfonts}%
\usepackage{amsthm}%
\usepackage{mathrsfs}%
\usepackage[title]{appendix}%
\usepackage{xcolor}%
\usepackage{textcomp}%
\usepackage{manyfoot}%
\usepackage{booktabs}%
\usepackage{caption}
\usepackage{array}%
\usepackage{algorithm}%
\usepackage{algorithmicx}%
\usepackage{algpseudocode}%
\usepackage{listings}%
\usepackage{rotating} 
\usepackage{booktabs} 
\usepackage{xcolor} 
\usepackage{pdflscape}   
\usepackage{longtable}   
\usepackage{booktabs}    
\usepackage{array}       
\usepackage{graphicx}
\usepackage{changepage}
\usepackage{makecell}
\usepackage{titlesec}
\usepackage{tabularray}
\usepackage{xcolor}
\usepackage{adjustbox}
\usepackage{longtable}
\usepackage{pdflscape} 
\usepackage{booktabs}  
\usepackage{array}     
\usepackage{geometry}
\usepackage{enumitem}

\usepackage{amssymb}
\usepackage{balance}
\usepackage{ragged2e}
\usepackage{enumitem}
\usepackage{color}
\usepackage{tikz}
\usepackage{textcomp}
\usepackage{amsfonts}
\usepackage{geometry}
\usepackage{enumitem}
\usepackage{booktabs}%
\usepackage[colorlinks=true, linkcolor=blue, citecolor=blue, urlcolor=blue]{hyperref}
\newcommand{\RNum}[1]{\MakeUppercase{\expandafter{\romannumeral #1\relax}}}

\begin{document}
\makeatletter
\def\ps@pprintTitle{%
 \let\@oddhead\@empty
 \let\@evenhead\@empty
 \let\@oddfoot\@empty
 \let\@evenfoot\@empty
}
\makeatother
\begin{frontmatter}

\title{A Deep Dive into Generic Object Tracking: A Survey}

\author[label1]{Fereshteh Aghaee Meibodi}
\ead{fereshtehaghaee@uvic.ca}

\author[label1]{Shadi Alijani}
\ead{shadialijani@uvic.ca}

\author[label1]{Homayoun Najjaran\corref{cor1}}
\ead{najjaran@uvic.ca}

\cortext[cor1]{Corresponding author}

\affiliation[label1]{organization={University of Victoria},
         city={Victoria},
         state={BC},
         postcode={Canada},
         country={V8P 5C2}}

\begin{abstract}
{Generic object tracking remains an important yet challenging task in computer vision due to complex spatio-temporal dynamics, especially in the presence of occlusions, similar distractors, and appearance variations. Over the past two decades, a wide range of tracking paradigms, including Siamese-based trackers, discriminative trackers, and, more recently, prominent transformer-based approaches, have been introduced to address these challenges. While a few existing survey papers in this field have either concentrated on a single category or widely covered multiple ones to capture progress, our paper presents a comprehensive review of all three categories, with particular emphasis on the rapidly evolving transformer-based methods. We analyze the core design principles, innovations, and limitations of each approach through both qualitative and quantitative comparisons. Our study introduces a novel categorization and offers a unified visual and tabular comparison of representative methods. Additionally, we organize existing trackers from multiple perspectives and summarize the major evaluation benchmarks, highlighting the fast-paced advancements in transformer-based tracking driven by their robust spatio-temporal modeling capabilities.}
\end{abstract}

\begin{keyword}
{Generic Object Tracking \sep Siamese-based Trackers \sep Discriminative-based Trackers \sep Transformer-based Trackers.}
\end{keyword}

\end{frontmatter}

\section{Introduction}\label{sec1}

Visual object tracking (VOT) is the task of continuously localizing a target object across frames in a video in computer vision. Over the years, several tracking paradigms have been developed including generic object tracking, multi-object tracking, motion-based tracking, appearance-based tracking, and video object segmentation, among others. In this paper, we focus on generic object tracking (GOT) also known as single object tracking (SOT), which operates in a class-agnostic manner. In this setting, the tracker receives an initial annotation of the target (typically a bounding box) in the first frame and is expected to locate the target in all subsequent frames without any additional supervision.

Generic object tracking based on appearance models presents several fundamental challenges. These include variations in the target’s appearance, scale, and pose, as well as occlusion, deformation, motion blur, and the presence of distractors and background clutter. Despite these difficulties, appearance-based tracking methods have received increasing attention due to their broad applicability in domains such as autonomous transportation, video surveillance, medical diagnostics, and robotic navigation.

\begin{figure*}[h]
\centering
\includegraphics[width=\textwidth]{ 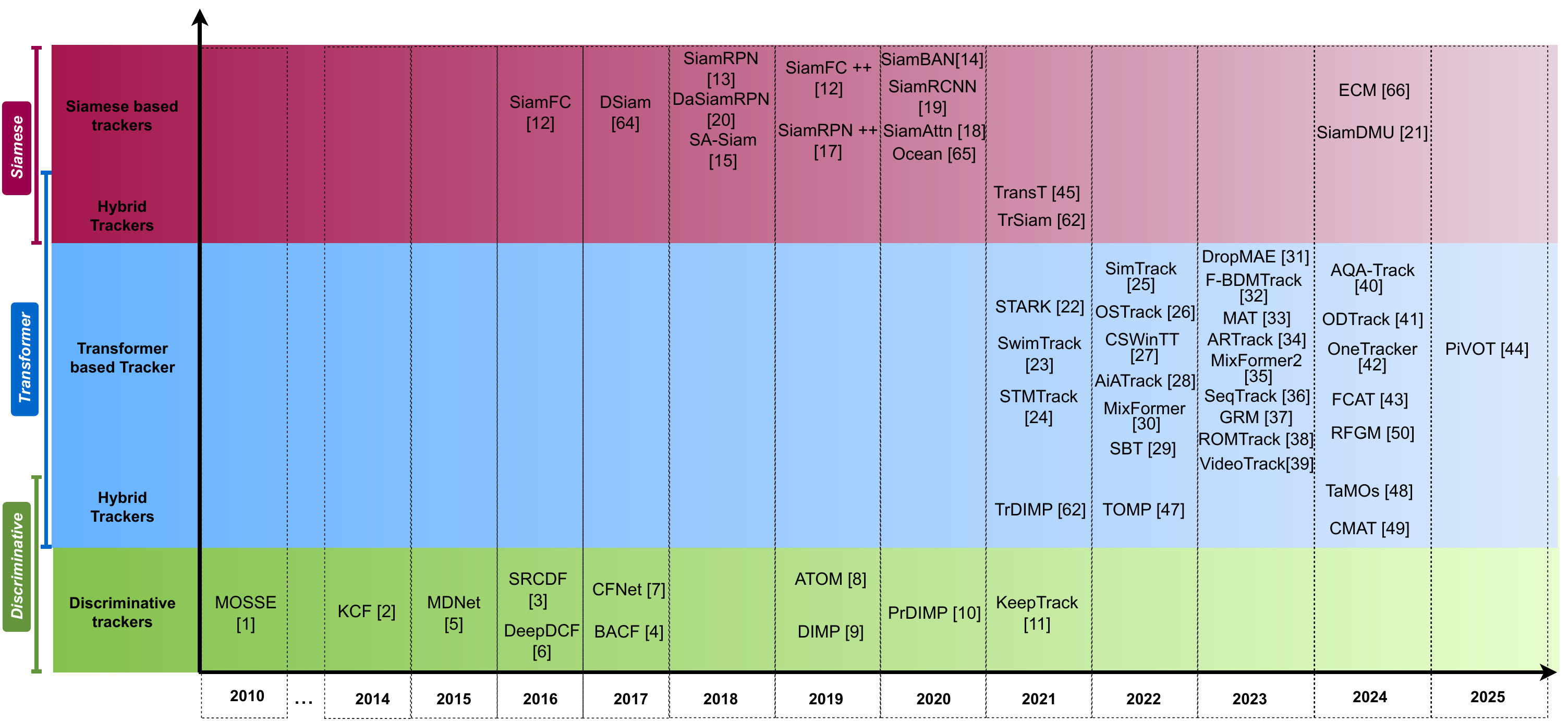}
\caption{A timeline of major breakthroughs in generic object tracking since 2010, with a particular focus on the past decade across Siamese-based, discriminative-based, and transformer-based paradigms.}\label{fig1}
\end{figure*}

Illustrated in Figure.~\ref{fig1}, the evolution of tracking algorithms began with hand-crafted discriminative methods, which relied on correlation filters and online optimization in order to distinguish the target from its background \cite{mosse, kcf,srdcf,bacf}. With the advent of deep learning, discriminative-based trackers began incorporating convolutional neural networks (CNNs) for feature extraction which are often used to train classifiers or regressors that distinguish the target from the background \cite{mdnet, deepdcf, cfnet, atom, dimp, prdimp, keeptrack}. On the other hand, Siamese-based trackers perform template matching between the initial target and candidate regions by computing similarity scores \cite{siamfc, siamrpn, siamban, sasiam, siamfcpp, siamrpnpp, siamattn, siamrcnn, dasiamrpn, siamdmu}. These two paradigms evolved in parallel with significant focus on improving robustness, adaptation, and appearance modeling through deeper backbones \cite{siamrpn}, distractor-aware mechanisms \cite{keeptrack, bacf, siamrpn}, and advanced model update strategies \cite{dimp, keeptrack}.

The field has advanced even more recently with the introduction of transformer architectures. Transformers enable powerful global modeling of spatial and temporal dependencies through self-attention and cross-attention mechanisms. Depicted in the timeline in Figure.~\ref{fig1}, many state-of-the-art trackers now leverage transformers, either as standalone models \cite{stark, swintrack, stmtrack, simtrack, ostrack, cswintt, aiatrack, sbt, mixformer, dropmae, f-bdmtrack, mat, artrack, mixformerv2, seqtrack, grm, romtrack, videotrack, aqa-track, odtrack, onetracker, fcat, pivot} or in hybrid architectures that fuse transformer modules with discriminative or Siamese components \cite{transt,trsiamtrdimp,tomp,tamos,cmat,rfgm}.
In this survey, we review and analyze representative methods from three major families of \RNum{1}. Discriminative-based trackers, \RNum{2}. Siamese-based trackers, and \RNum{3}. Fully Transformer-based and hybrid Transformer-based trackers.

While our emphasis is on recent advancements, we also include foundational earlier works to trace the progression of design strategies and architectural trends. To the best of our knowledge, this is among the first comprehensive survey that jointly reviews and compares these three categories of generic object trackers and recent methods across multiple dimensions, including appearance modeling, design highlight, update strategy, and overall tracking framework. Furthermore, we systematically analyze the challenges addressed by each method, their proposed novelties to overcome these challenges, the potential drawbacks they introduce, and the level of architecture in their model at which they contribute. In addition, to architectural and methodological comparisons, we also analyze the tracking datasets commonly used for training and evaluation. We also conduct a structural comparison by reconstructing standardized architectural diagrams for representative trackers, enabling consistent and direct visual analysis of their design principles and innovations.
\newline
\newline
The main contributions of this work are as follows:\\

   \RNum{1}. \textbf{Comprehensive Categorization of Tracking Paradigms} \\
   We propose a unified taxonomy that systematically categorizes GOT trackers into three core paradigms: Siamese-based, discriminative-based, fully and hybrid transformer-based. To the best of our knowledge, this is the first survey that jointly analyzes both baseline and recent methods across these categories, providing a broader and more inclusive perspective compared to existing reviews.\\

    \RNum{2}. \textbf{Unified Architectural Frameworks for Structural Comparison} \\
    For every representative tracker, including those that only discuss the methodology in theory, we reconstruct standardized visual frameworks to facilitate consistent structural analysis. By highlighting important architectural elements and allowing for a clear understanding of design evolution across paradigms, this unified representation makes it easier to compare tracker designs directly.\\

    \RNum{3}. \textbf{Multi-Dimensional Comparative Analysis and Performance Comparison} \\
    We perform a thorough analysis of trackers using several architectural and functional dimensions, such as appearance model, backbone architecture, design highlights, focus, and novel contributions. We systematically examine the challenges each method addresses, the innovations proposed to overcome them, and the potential drawbacks introduced.  In addition, we examine the tracking datasets used for training and evaluation. Then we compare trackers and illustrate the trade-offs between accuracy and efficiency.\\

The remainder of this paper is organized as follows:
In Section \ref{review} we will review existing survey papers in the field of GOT and highlight how our work differs from them. Section \ref{vot} provides an overview of GOT methods, categorizing them into four main groups: discriminative-based trackers (Section \ref{disc}), Siamese-based trackers (Section \ref{siamese}), transformer-based trackers (Section \ref{transformer}), which are further divided into hybrid and fully transformer-based approaches in Section \ref{hybrid trans} and Section \ref{fully trans}, respectively. In addition to a summary of popular tracking datasets and evaluation metrics, Section \ref{exp} offers an evaluation and comparison of the reviewed trackers in terms of accuracy and efficiency. Section~\ref{discussion} provides a comprehensive discussion of GOT approaches from both architectural and functional perspectives. In this section, recent state-of-the-art designs and emerging trends such as segmentation-assisted tracking are highlighted. Applications of VOT are discussed in Section \ref{app}. The paper is finally concluded in Section \ref{conclusion}, also outlining future research directions in the field.

\section{Background}\label{review}

Generic visual object tracking (GOT) has been extensively studied, and several surveys have reviewed its development from conventional methods to deep learning and beyond as shown in Table~\ref{tab:survey_list}.

\citet{s1} analyzes deep learning-based trackers, including appraches based on Convolutional Neural Networks (CNN), Recurrent Neural Network (RNN), and GAN(Generative Adversarial Network), across multiple dimensions. However, it provides less detailed taxonomy on their architectural design. The work in \cite{s2} offers a timeline-based view, dividing trackers into correlation filter-based and deep learning-based models, including CNNs, RNNs, and Siamese-based trackers. \citet{s3} provides a detailed discussion focusing on how deep learning addresses four core challenges in tracking. It reviews both single-object and multi-object tracking methods. However, it lacks detailed architectural analysis. The survey in \cite{s4} focuses on online updating strategies in trackers, highlighting the importance of adaptability to dynamic environments. However, this survey primarily concentrates on traditional and CNN-based methods in Siamese and discriminative-based trackers, with a particular emphasis on adaptivity during tracking.

The study \cite{s5} focuses particularly on the two dominant paradigms of Discriminative Correlation Filters (DCF) and Siamese Networks. It provides a detailed analysis of shared and specific challenges within these two families only. In addition, the focus of \cite{s6} is narrowed to Siamese-based tracking, thoroughly examining the design principles, strengths, and limitations of this family, without considering discriminative and transformer-based approaches. A broader perspective is adopted by \citet{s7}, which included Siamese-based, discriminative-based, and early transformer-based models. However, it treats different paradigms without a distinct breakdown of architectural and methodological innovations per paradigm.

Further, \citet{s8} offers an experimental analysis of transformer-based trackers, categorizing them into CNN-Transformer models and fully Transformer-based trackers. Nevertheless, it lacks systematic comparison of these trackers with traditional discriminative or Siamese paradigms. Lastly, \citet{s9} explores beyond the conventional approaches, such as autoregressive models, generative models, self-supervised learning, reinforcement learning, and meta-learning in tracking. While it highlights emerging directions, it does not address the evolution of standard tracking architectures or paradigms.

While existing surveys provide important insights into specific families (such as DCF, Siamese, or transformer-based trackers) or focus on emerging learning paradigms, to the best of our knowledge, none offers a unified taxonomy that systematically categorizes GOT trackers across all major paradigms: Siamese-based, discriminative-based, and fully/hybrid transformer-based models. Furthermore, none extensively analyzes trackers across multiple architectural and functional dimensions, including appearance modeling, backbone architecture, template update strategy, novelty contributions, drawbacks, and architecture-level innovation.

In this survey, we bridge this gap by introducing a unified, fine-grained categorization and comparison of recent GOT trackers across all major categories. We provide a consistent structural analysis across paradigms, systematically compare their empirical trade-offs between accuracy and efficiency, and identify trends, challenges, and open research directions in modern object tracking.

\begin{table}[ht]

\centering
\caption{List of existing generic object tracking (GOT) surveys.}
\label{tab:survey_list}
\begin{tabular}{p{1cm}p{13.5cm}}
\toprule
\textbf{Year} & \textbf{Survey Title} \\
\midrule

2021 & Deep Learning for Visual Tracking: A Comprehensive Survey \cite{s1} \\
2021 & Recent Advances of Single-Object Tracking Methods: A Brief Survey \cite{s2}\\
2021 & Deep Learning in Visual Tracking: A Review \cite{s3}\\ 
2021 & A survey on online learning for visual tracking \cite{s4} \\
2021 & Visual Object Tracking With Discriminative Filters and Siamese Networks: A Survey and Outlook \cite{s5} \\
2022 & Siamese Visual Object Tracking: A Survey \cite{s6}\\
2022 & Visual Object Tracking: A Survey \cite{s7} \\
2023 & Transformers in Single Object Tracking: An Experimental Survey \cite{s8} \\

2024 & Beyond Traditional Single Object Tracking: A Survey \cite{s9}\\
\bottomrule
\end{tabular}
\end{table}

\section{Generic Visual Object Tracking}\label{vot}

Detection-based generic visual object tracking aims to estimate the trajectory of an arbitrary target object in a video sequence, given only its initial location in the first frame. Over the past decade, GOT techniques have evolved significantly to cope with key challenges, including occlusions, target deformations, scale variations, illumination changes, background distractors. Consequently, tracking algorithms must consider both short-term and long-term adaptation of their reference target representation in order to remain robust against drastic target appearance changes.

The tracking problem can be formulated as a combination of a classification task and a target state estimation task \cite{atom}. The classification branch aims to robustly determine the coarse location of the target object, while the state estimation branch refines the prediction to accurately determine the full target state, typically represented as a bounding box. A high-performance tracker must learn expressive feature representations and corresponding classifiers that are simultaneously discriminative and generalizable. Being discriminative enables the tracker to differentiate the true target from cluttered or deceptive background regions, while being generalizable allows it to tolerate appearance changes of the tracked object, even when the object category is unknown \cite{sasiam}.

Similar to other fields in computer vision, tracking methods have evolved from relying on hand-crafted features to utilizing deep features and, more recently, transformer-based representations. In this survey, we categorize modern GOT trackers into three major paradigms based on their core architectural principles in order to cover this evolution. \textbf{Discriminative trackers} primarily rely on online learning to construct an appearance model through discriminative formulations, although recent advances have leveraged offline training of more representative features to significantly boost their accuracy. \textbf{Siamese-based trackers}, in contrast, are trained offline to learn feature representations that are robust to appearance variations. During inference, the tracking process involves extracting features from both the template and the search region and applying a fixed matching operation, typically cross-correlation, to localize the target. Recently, attention has shifted toward \textbf{transformer-based} designs, which have advanced tracking performance by modeling long-range dependencies. Transformer modules can be integrated into trackers in a hybrid manner alongside Siamese or discriminative structures, or they can form fully transformer-based tracking architectures.

The underlying architectures play a pivotal role in determining tracking robustness, efficiency, and adaptability. The evolution of methods within each paradigm aims to address critical aspects such as online adaptation, representative feature extraction, accurate target state estimation, robust appearance modeling, effective distractor handling and reliable matching strategies. In the following subsections, we will review representative methods within each category, highlighting their architectural innovations, strengths, and limitations.

\subsection{Discriminative-based Tracking}\label{disc}
Discriminative-based trackers formulate the tracking problem as a binary classification task that distinguishes the target object from the background. In these methods, an appearance model, which can be a correlation filter or convolutional layer, is trained to discriminate between positive samples containing the target and negative samples in background regions by minimizing a discriminative objective function. A key characteristic of discriminative tracking approaches is their focus on online learning and template update during inference, allowing the tracker to adapt to appearance variations, occlusions, and environmental changes in real-time. Early discriminative trackers mostly relied on hand-crafted features and simple classifiers such as support vector machines or ridge regression. Subsequent approaches shifted toward using deep features and optimization-based prediction models. An explanation of the most well-known discriminative trackers is provided below. Together with their matching architectures, they are presented in an unified and organized way to make comparison and analysis simpler. In addition, 
Table~\ref{disctab} provides a detailed specification of these methods, emphasizing their temporal evolution.

Correlation filter (CF)-based trackers have played an important role in advancing discriminative tracking. In these methods, discriminative classifiers are trained online using samples collected during the tracking, helping the tracker to adapt to the changing appearance of the target. Correlation filters efficiently learn a linear template that discriminates the target patch from surrounding background patches by solving a ridge regression problem. The main innovation of CF-based tracking is the use of the Fast Fourier Transform (FFT) to perform calculations in the Fourier domain and take advantage of the properties of circular correlation. This allows for incredibly quick filter training and updating, usually once per frame. During tracking, the correlation filter is applied on a small search window centered around the previous target position and the maximum response in the filter output determines the new location of the target. After every frame, CF-based trackers update the filter weights online, allowing the model to dynamically adapt to photometric and geometric changes in the target's appearance. Furthermore, some CF-based approaches estimate both the target location and scale by selecting the scale corresponding to the highest correlation output. With their introduction, correlation filter-based trackers achieved a breakthrough by offering competitive accuracy compared to the best methods of their time while significantly outperforming them in computational efficiency due to the use of Fourier domain operations.

Minimum Output Sum of Squared Error (MOSSE) tracker \cite{mosse} is one of the earliest CF-based trackers. It proposes a simple and real-time tracking method that is robust to variations in scale, lighting, pose, and non-rigid deformations. In contrast to earlier correlation filter-based approaches, which employed more complicated appearance models and optimization strategies and were relatively slow, MOSSE introduced a much more efficient adaptive tracking framework. It trains the correlation filter using only a single frame, significantly reducing the data requirements compared to previous adaptive CF methods such as ASEF \cite{asef}, which required a large number of training samples. MOSSE can be interpreted as a regularized variant of ASEF, improving stability and robustness by minimizing the output sum of squared error and enabling efficient online adaptation during tracking.

While MOSSE focused on real-time adaptive tracking with simple linear correlation filters, the Kernelized Correlation Filter (KCF) \cite{kcf} continued this direction by introducing a kernelized formulation and multi-channel feature support, such as Histogram of Oriented Gradients (HOG), to improve discriminative power and feature representation. KCF exploits the circulant structure of translated image patches to enable efficient performance. By applying the Discrete Fourier Transform (DFT), it reduces both storage and computational complexity, allowing real-time operation even when using richer feature representations.

MDNet \cite{mdnet} addresses the limitations of hand-crafted features in learning robust target representations by introducing a CNN-based discriminative tracker. Rather than relying on pretrained classification networks as its backbone, which are often ineffective due to the gap between classification and tracking tasks domain, MDNet employs a multi-domain learning framework that separates domain-independent and domain-specific information. During offline training, shared convolutional layers are learned across multiple video sequences, while separate domain-specific branches are trained for binary classification. During inference time, a new domain-specific branch is initialized and fine-tuned online to allow the tracker to adapt effectively to the target's appearance in a new sequence.

Standard DCF-based trackers suffer from boundary artifacts due to the circular convolution assumption. SRDCF \cite{srdcf} (Spatially Regularized Discriminative Correlation Filter) addresses this issue by introducing a spatial regularization term that penalizes filter coefficients based on their spatial location. This enables learning from larger image regions with richer negative samples while focusing on the target. To maintain computational efficiency, the method leverages the sparsity of the regularization in the Fourier domain and employs a Gauss-Seidel solver for online optimization.

DeepDCF \cite{deepdcf} investigates the integration of pretrained convolutional layer activations into correlation filter-based trackers in order to replace traditional hand-crafted features. The study applies these features within both the standard DCF and SRDCF frameworks and shows that shallow convolutional layers, particularly the first layer, offer superior tracking performance compared to deeper ones. This insight highlights the value of spatially detailed and semantically meaningful representations for visual tracking that leads to consistent improvements over conventional features like HOG and Color Names.

Unlike conventional Siamese trackers (Section \ref{siamese}) like SiamFC that match each frame to a static template, CFNet \cite{cfnet} integrates an online correlation filter as a differentiable layer within a shallow Siamese network, enabling end-to-end learning of both the tracking model and the feature representation. To improve adaptability to changes in appearance, this model uses a running average to update the template online. Its key innovation is treating the correlation filter as a closed-form optimization block embedded in the network via back-propagation through the CF solution. This method maintains high speed and efficiency while allowing the network to learn features tailored for correlation-based tracking.

BACF (Background-Aware Correlation Filters) \cite{bacf} addresses a core limitation of traditional CF trackers learning only from circularly shifted target patches and neglecting real background information, which can lead to overfitting and poor discrimination in cluttered scenes. BACF enables the tracker to learn filters that better distinguish the foreground from surrounding distractions by proposing to densely sample real background patches as negative examples. It introduces an efficient ADMM-based optimization to train multi-channel filters with real-time performance, achieving strong accuracy without relying on deep features.

\begin{figure}[!h]
\centering
\includegraphics[width=1.02\textwidth]{ 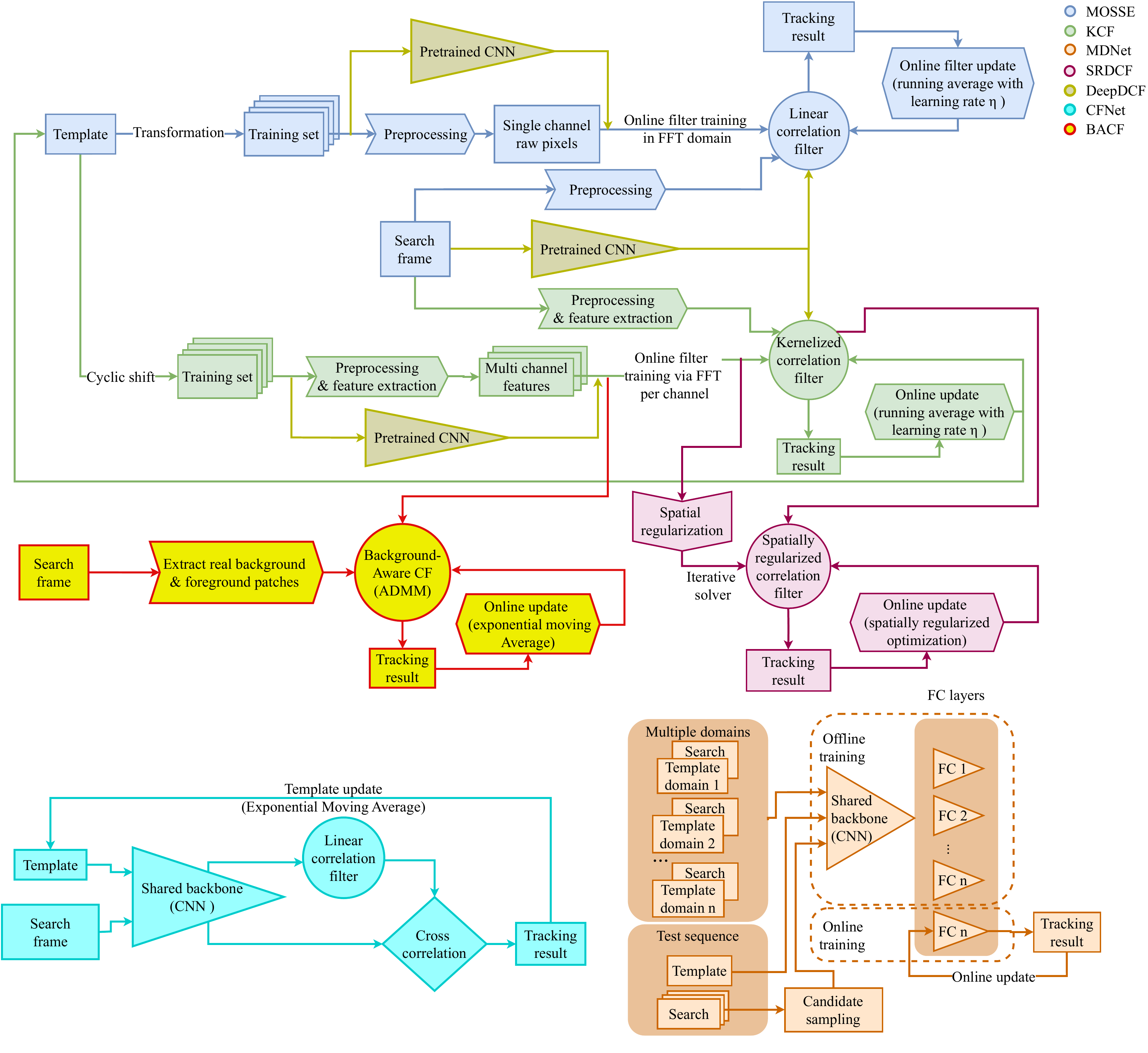}
\caption{Visual overview of frameworks in ealier discriminative trackers. This figure illustrates the progression from early trackers based on hand-crafted features  MOSSE\cite{mosse}, KCF \cite{kcf}, SRDCF \cite{srdcf}, and BACF \cite{bacf} to those leveraging convolutional neural networks (CNNs). The transition began with pre-trained backbones DeepDCF\cite{deepdcf}, CFNet \cite{cfnet}, and MDNet \cite{mdnet}.The diagram also details differences in their appearance modeling approaches and online update strategies.} \label{fig_disc_group1}
\end{figure}

Figure.~\ref{fig_disc_group1} presents a high-level architectural overview of these earlier discriminative-based trackers. It offers a comprehensive visual summary of their core components and highlights key architectural trends across the discussed methods, including variations in feature extraction, classification, update mechanisms, and their novelties.

\begin{figure}[!h]
\centering
\includegraphics[width=1.0\textwidth]{ 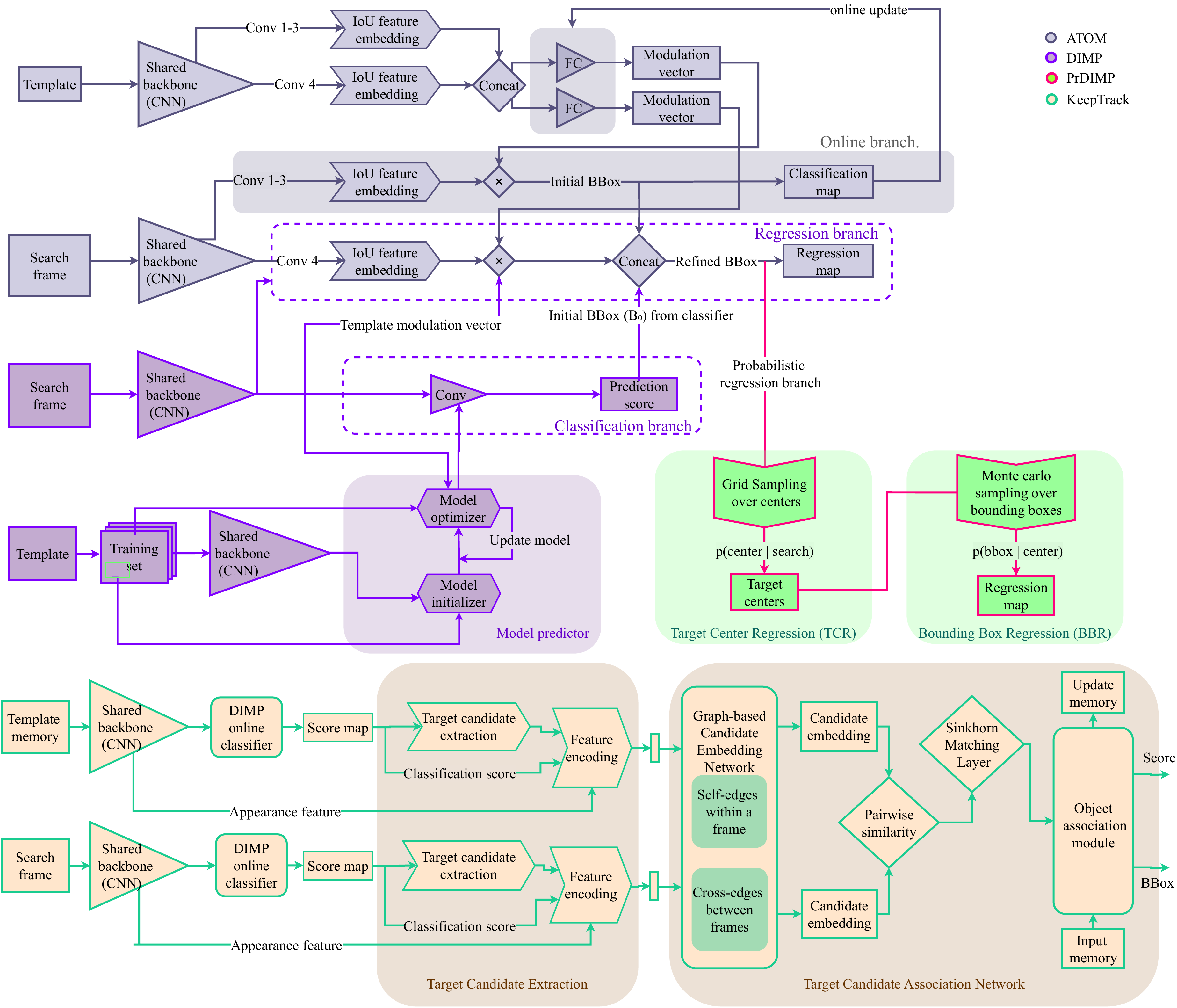}
\caption{Visual overview of more advanced discriminative trackers including more advanced target state estimation in ATOM \cite{atom}, optimization-based discriminative models in DiMP \cite{dimp} and PrDiMP \cite{prdimp}. It also includes discriminative KeepTrack \cite{keeptrack} tracker via learnable target candidate association  .} \label{fig_disc_group2}
\end{figure}

Prior discriminative trackers like \cite{bacf, srdcf, kcf}, rely on multi-scale search without modeling target-specific appearance or aspect-ratio changes. ATOM \cite{atom} shown in Figure.~\ref{fig_disc_group2} addresses this key limitation by introducing a two-stream architecture that decouples target classification and state estimation. Its classification branch is trained online using a lightweight convolutional network optimized with a conjugate-gradient strategy, while the state estimation module is trained offline to predict IoU scores between proposals and the target. Through the use of feature modulation to integrate target-specific features, ATOM provides reliable and accurate bounding box estimation under difficult pose and viewpoint variations.

DiMP \cite{dimp} in Figure.~\ref{fig_disc_group2} improves previous discriminative trackers by improving their ability to distinguish the target from background distractors, which is often hindered by limited use of background information. It formulates target model learning as an optimization problem derived from a discriminative loss, where the target model is represented as a convolutional layer updated through an iterative steepest-descent procedure. A meta-learned optimizer, trained offline, is used to adapt this model online in a few gradient steps using both positive and densely sampled negative examples from the current frame. This enables DiMP to construct a robust, target-specific classifier that generalizes well to appearance changes and unseen targets, while maintaining strong target-background separation throughout tracking. Additionally, DiMP incorporates a parallel IoU-prediction branch for accurate bounding box estimation.

PrDiMP \cite{prdimp} enhances the DiMP tracker \cite{dimp} by reformulating both target center localization and bounding box regression as probabilistic regression tasks. Unlike confidence-based methods that predict scalar scores, PrDiMP models the conditional probability density of the target state directly through the network architecture, without assuming a predefined distribution. This enables the tracker to represent uncertainty in the annotation itself as well as in the target state. The model is trained by minimizing the Kullback-Leibler divergence between predicted and label distributions, enabling it to reason about ambiguities and label noise. This probabilistic formulation improves robustness in challenging scenarios with occlusion, blur, or similar distractors. The architecture od this paper is illustrated in Figure.~\ref{fig_disc_group2}.

Another paper working on robustness against distractors is KeepTrack \cite{keeptrack} which introduces an explicit target candidate association mechanism, rather than relying solely on a more powerful appearance model. It extends the DiMP \cite{dimp} framework by incorporating the target classifier from DiMP and the probabilistic bounding box regressor from PrDiMP \cite{prdimp}. Shown in Figure.~\ref{fig_disc_group2}, a learned Target Candidate Association Network is used to propagate candidate identities across frames by associating each candidate using features like position, score, and appearance. To enable distractor-aware learning in the absence of ground-truth annotations, the paper combines partial labels with a self-supervised training strategy. A graph-based Candidate Embedding Network is employed to capture relationships among nearby candidates. Furthermore, during online updates, a memory sample confidence mechanism evaluates the reliability of training samples to reduce the influence of unreliable samples and improve adaptability in the presence of distractors.

\begin{sidewaystable}[htbp]
\tiny
\centering
\caption{A Detailed Comparison of Discriminative-Based Trackers.}
\label{disctab}
\begin{tabular}{%
    >{\centering\arraybackslash}p{0.5cm}|  
    p{1.3cm}  
    p{0.5cm}  
    p{1.3cm}  
    p{2.5cm}  
    p{2.5cm}  
    p{2.5cm}  
    p{2.5cm}  
    p{1cm}    
    p{2cm}    
}
\midrule
\multirow{12}{*}{\rotatebox{90}{\parbox[c]{10cm}{\centering \textbf{Discriminative-Based} \\ \textbf{Appearance Model}}}} & 
\textbf{Method} & \textbf{Year} & \textbf{Backbone} & \textbf{Design Highlight} & \textbf{Focus} & \textbf{Novelty} & \textbf{Drawbacks} & \textbf{Template Update} & \textbf{Architectural-Level of Contribution} \\
\cmidrule{2-10}
& MOSSE \cite{mosse} & 2010 & None & Adaptive MOSSE filter; FFT optimization; One-frame training & Real-time CF tracking with online updates & First adaptive CF; Real-time updates from one frame & Grayscale only; no deep features & Yes & Appearance model \\
\cmidrule{2-10}
& KCF \cite{kcf} & 2014 & None (hand-crafted features) & Kernelized CF with DFT; Multi-scale via hand-crafted features & Fast discriminative CF with HOG/raw input & Efficient kernel trick; circular shift formulation & Sensitive to lighting and deformation & Yes & Appearance model \\
\cmidrule{2-10}
& MDNet \cite{mdnet} & 2015 & VGG-M \cite{vgg}  & Multi-domain CNN; Offline+online adaptation; Hard negative mining & Feature generalization; domain-specific adaptation & Separate domain-specific/shared layers; fine-tuning & Slow due to SGD updates; costly training & Yes & Training procedure \\
\cmidrule{2-10}
& SRDCF \cite{srdcf} & 2016 & None & Spatial regularization in DCF; extended search region & Improves background modeling; mitigates boundary effects & Spatial regularization; real-time Gauss-Seidel updates & Complex tuning; high computational cost & Yes & Appearance model \\
\cmidrule{2-10}
& DeepDCF \cite{deepdcf} & 2016 & VGG-M \cite{vgg} & Deep features in DCF/SRDCF; shallow layer effectiveness & Improve robustness over HOG/CN features & Demonstrates CNN layer value for tracking & No end-to-end training; fixed pretrained CNN & Yes & Feature representation \\
\cmidrule{2-10}
& CFNet \cite{cfnet} & 2017 & Siamese Net & End-to-end trainable CF layer; Running-average update & Add CF adaptability to Siamese tracking & First trainable CF layer within CNN & Shallow architecture; no bbox regression & Yes & Feature representation; Online update \\
\cmidrule{2-10}
& BACF \cite{bacf} & 2017 & None & Real negative sampling; multi-channel hand-crafted features & Model background clutter; Improve adaptability & Dense real background sampling; ADMM optimizer & Limited long-term memory; not deep-learned & Yes & Appearance model; Sampling strategy \\
\cmidrule{2-10}
& ATOM \cite{atom} & 2019 & ResNet-18 \cite{resnet} & Two-branch model: IoU-based estimation + classification; CG-based update & Accurate target state estimation; distractor handling & Combines modulation and IoU scoring & Multiple hyperparams; relatively expensive & Yes & Target state estimation \\
\cmidrule{2-10}
& DiMP \cite{dimp} & 2019 & ResNet-18 / ResNet-50 \cite{resnet} & Meta-learned optimizer; Steepest descent; IoU branch from ATOM & Discriminative end-to-end model update & Optimization-based learning with meta-training & High training and inference cost; complex tuning & Yes & Appearance model \\
\cmidrule{2-10}
& PrDiMP \cite{prdimp} & 2020 & ResNet-18 / ResNet-50 \cite{resnet} & Probabilistic regression; KL-loss for uncertainty modeling & Improve robustness to ambiguous cases & Predicts distributions over location and box & Complex training; label noise sensitivity & Yes & Target state estimation; Training procedure \\
\cmidrule{2-10}
& KeepTrack \cite{keeptrack} & 2021 & ResNet-18 / ResNet-50 \cite{resnet} & Candidate Association Net + Memory confidence + Graph learning & Distractor-aware online updates & Identity-aware association; memory-based selection & Sensitive to candidate quality; complex graph module & Yes & Prediction head; Online update \\
\midrule
\end{tabular}
\end{sidewaystable}

\subsection{Siamese-based Tracking}\label{siamese}
Siamese-based trackers represent a prominent paradigm in generic object tracking, where the task is formulated as a similarity matching problem between a target template and a search region. A typical Siamese network consists of two shared-weight branches: the template branch, which processes the target patch from the first frame, and the search branch, which processes a region from the current frame. Both branches embed their inputs into a common feature space using a shared backbone, and the similarity between the two is computed to localize the target. Different types of fusion mechanisms have been proposed for comparing these embeddings, ranging from fully connected layers (e.g., in GOTURN \cite{goturn}) to depth-wise and point-wise cross-correlation tensors in more advanced models. These trackers are trained offline on large-scale datasets to learn general matching functions, which provide a fast and efficient online inference without extensive adaptation. Siamese-based models have progressively improved in robustness and accuracy over time through innovations such as novel regression heads, update mechanisms, deeper backbones, and attention modules. The ability of Siamese trackers to balance high-speed inference with competitive accuracy makes them one of the key components in modern tracking systems. Below is a description of the most well-known Siamese trackers along with their corresponding architectures presented in a unified manner to facilitate tracker comparison. Furthermore, the structured details of the reviewed Siamese-based algorithms over the course of time is represented in Table. \ref{siamtab}.

SiamFC \cite{siamfc} introduced a fully convolutional Siamese network trained end-to-end on large-scale video datasets to learn a general-purpose similarity function for tracking. The network consists of two identical branches that extract embeddings from the target template and search region, followed by a cross-correlation layer that produces a dense response map indicating the target's location. This architecture provides efficient sliding-window matching in a single forward pass without requiring online model updates and handles scale variation by applying multi-scale evaluation using a search pyramid. Additionally, a cosine window is applied to the response map to suppress distractors and encourage smoother localization. Despite its simplicity and lack of online adaptation, SiamFC achieved strong real-time performance and established the foundation for subsequent Siamese-based tracking architectures.

DSiam \cite{dsaim} improves SiamFC by adding dynamic adaptability to changes in appearance over time and background clutter. It incorporates a fast online transformation learning module that adjusts the target template and search features using learned convolutional mappings, allowing real-time adaptation without replacing the template. The appearance variation transformation and background suppression transformation are learned efficiently in the frequency domain. Besides, to improve localization and robustness, DSiam integrates element-wise multi-layer feature fusion to leverage both deep and shallow layers. Unlike typical Siamese trackers trained on image pairs, DSiam is jointly trained on full video sequences, enabling it to exploit spatial-temporal dynamics. This method significantly outperforms static Siamese models like SiamFC in challenging scenarios by providing the balance between online adaptability and real-time speed.

SA-Siam \cite{sasiam} introduces a twofold Siamese network to improve the generalization of SiamFC by incorporating complementary appearance and semantic features. It consists of two separate appearance and semantic branches, each of them trained independently to preserve feature heterogeneity. The appearance branch retains the structure of SiamFC and focuses on similarity learning, while the semantic branch extracts high-level semantic features from a pretrained classification network. These branches are fused only at inference to generate a combined similarity score. To enhance target-specific representation in the semantic branch, SA-Siam employs a channel-wise attention mechanism that assigns weights to feature channels based on both target and surrounding context, enabling minimal but effective target adaptation. While performing in real-time, this model increases robustness against changes in appearance.

\begin{figure}[phtb]
\centering
\includegraphics[width=1\textwidth]{ 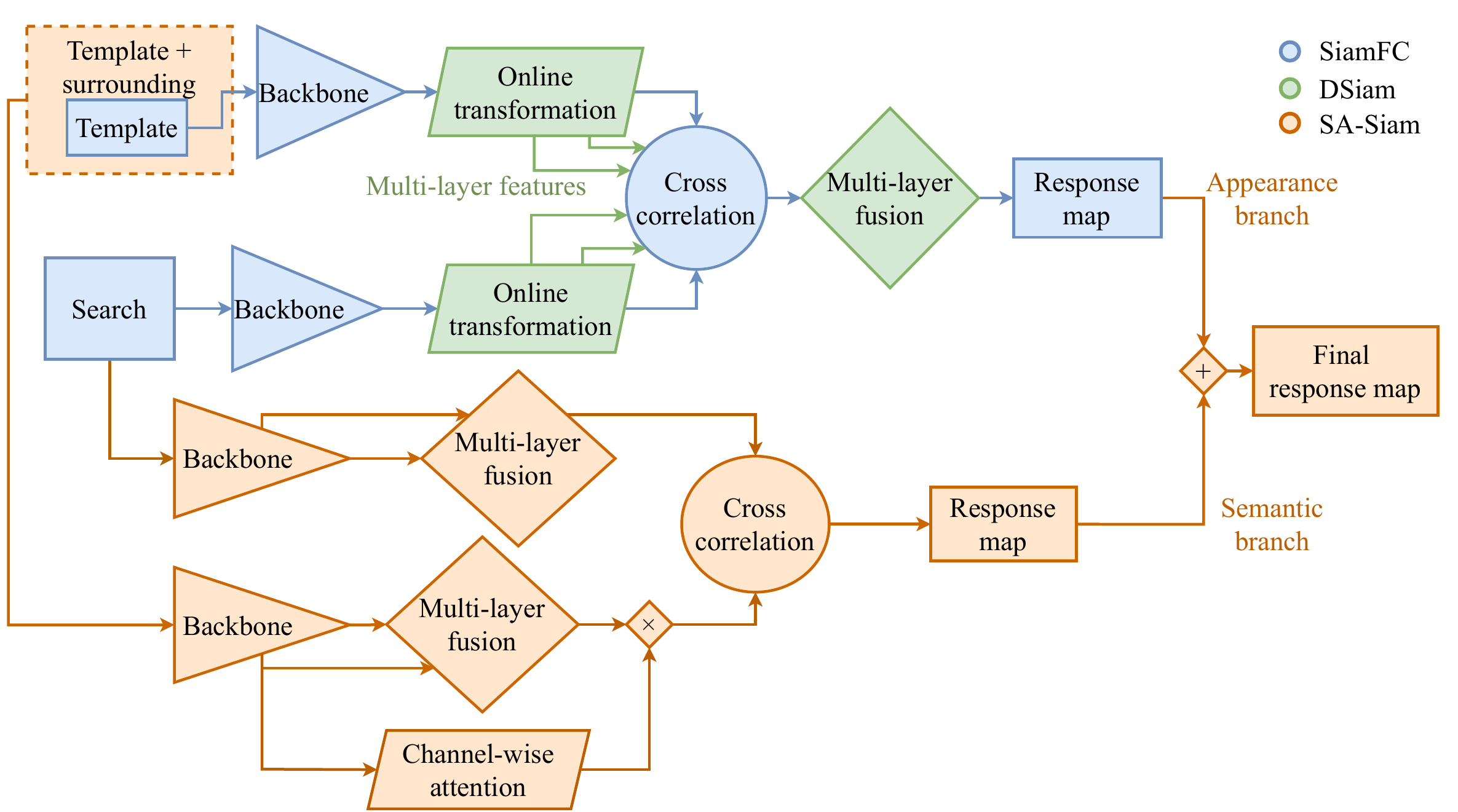}
\caption{Visual overview of early classification-based Siamese-based tracking frameworks namely SiamFC \cite{siamfc}, DSima \cite{dsaim}, and SA-Siam \cite{sasiam}.}\label{siam_group1}
\end{figure}

A high-level architectural comparison of above classification-based Siamese-based trackers \cite{siamfc, dsaim, sasiam} is provided in Figure.~\ref{siam_group1}, which highlights their progression and key innovations including multi-level feature fusion, attention modules, and online refinement mechanisms. 

SiamRPN \cite{siamrpn} introduces a Region Proposal Network (RPN) into the Siamese framework to enhance tracking accuracy and robustness. Adding RPN into the template and search branch enables accurate foreground-background classification and bounding box regression to achieve a more precise scale and aspect ratio estimation. The model eliminates the need for multi-scale search strategies used in earlier Siamese trackers like SiamFC. Moreover, it formulates tracking as a local one-shot detection task, where the template branch acts as a meta-learner to generate detection kernels for the search branch. This end-to-end offline training approach, combined with proposal refinement, results in a compact and highly efficient tracking pipeline.

The crucial problem of data imbalance between semantic and non-semantic backgrounds in generic object tracking, specifically the under representation of semantic distractors compared to non-semantic backgrounds during training is addressed by DaSiamRPN \cite{dasiamrpn}. It introduces a distractor-aware sampling strategy during offline training by incorporating semantic negative pairs from both the same and different categories in order to enable the network to learn more discriminative representations. During inference, a distractor-aware module uses hard negative mining along with a modified similarity function in order to incrementally learn how to adaptively suppress distractions. It employs a local-to-global search strategy for long-term tracking by gradually expanding the search area to re-detect targets that are occluded or out of view. These innovations enhance the short-term accuracy and long-term robustness of Siamese-based trackers.

Early Siamese trackers were limited by their inability to use deep backbones like ResNet \cite{resnet} because of their strict translation invariance and symmetric structural requirements. SiamRPN++ \cite{siamrpnpp} solves these issues by introducing a spatial-aware sampling strategy to break translation invariance, which enables end-to-end training with deeper networks. In addition, multi-level feature aggregation is employed across multiple ResNet layers to enhance robustness during appearance variations such as motion blur and deformation. These aggregated features are passed through three Siamese RPN modules and then fused with distinct weights for classification and regression. Furthermore, to resolve the parameter imbalance introduced by up-channel cross-correlation in SiamRPN, this paper proposes a depthwise cross-correlation module. This lightweight design reduces parameter count, stabilizes training, and yields higher accuracy by producing semantically meaningful, channel-separated similarity maps.

SiamFC++ \cite{siamfcpp} refines the original SiamFC framework by introducing a set of practical guidelines for accurate target state estimation in generic object tracking. The model separates classification and regression branches to decouple coarse target localization from precise bounding box prediction and eliminate the need for brute-force multi-scale search. Then it adopts an anchor-free, per-pixel estimation strategy that avoids ambiguity and dependency on prior knowledge of object scale and aspect ratio. To further improve precision, a quality assessment branch is introduced to estimate the reliability of bounding box predictions in order to address the mismatch that can occur between high classification confidence and poor localization. This branch outputs a parallel quality score map and is used to modulate the final tracking decision. SiamFC++ achieves high tracking accuracy in real time while maintaining architectural simplicity and generality.

\begin{figure}[phtb]
\centering
\includegraphics[width=1.0\textwidth]{ 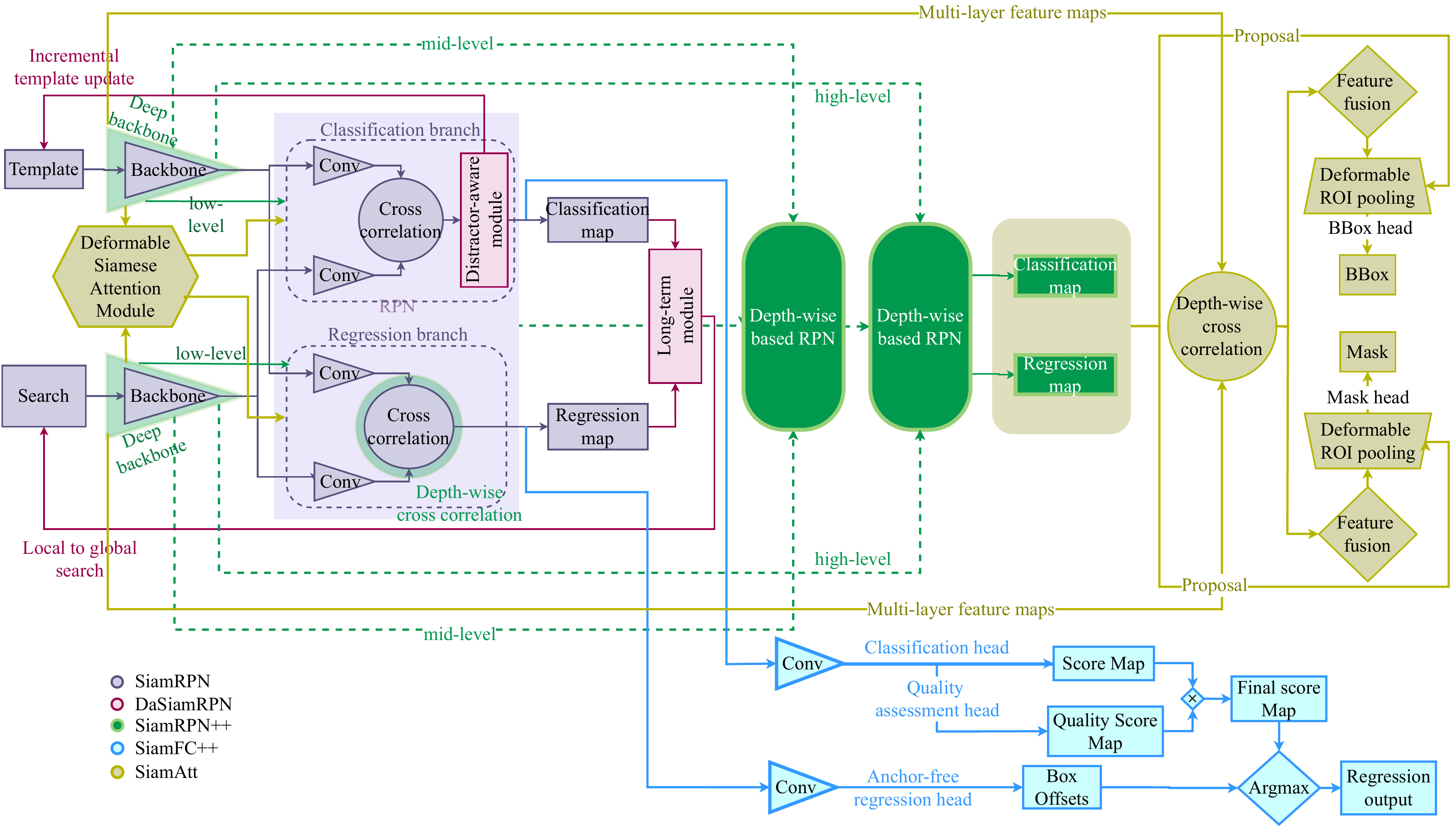}
\caption{Visual overview of Siamese-based tracking frameworks. This figure highlights more advanced Siamese approaches that improve localization accuracy through regression head. It also illustrates how Siamese trackers incorporate online update mechanisms in DaSiamRPN \cite{dasiamrpn} and SiamFC++ \cite{siamfcpp}. Additional architectural innovations and contributions of these methods, such as depth-wise correlation filters in SimaRPN++ \cite{siamrpnpp} and multi-level feature fusion and attention modules in SiamAtt \cite{siamattn}, can also be inferred from the figure.}\label{siam_group2}
\end{figure}

A high-level architectural comparison of Siamese trackers with localization head is provided in Figure.~\ref{siam_group2}, which highlights their key innovations including multi-level feature fusion, various types of cross-correlation, regression heads, and online update mechanisms. This visual overview shows how the functionality and complexity of Siamese-based tracking architectures have increased to meet existing challenges such as accurate localization, online adaptation, and distractor handling. 

Another paper, SiamBAN \cite{siamban}, addresses the challenges of accurate target state estimation in visual tracking by eliminating the need for predefined candidate boxes or multi-scale search schemes. The model predicts a foreground-background score and a 4D offset vector at each spatial location in the correlation feature maps, which describes the associated bounding box. By avoiding the tedious design of anchor parameters, this anchor-free approach makes SiamBAN more flexible and general. It also adopts multi-level prediction and depth-wise cross-correlation to enhance both efficiency and accuracy in order to achieve end-to-end offline training. The no-prior box design reduces the need for hyperparameters, enabling the tracker to adapt better to various scales and aspect ratios.

Siam R-CNN \cite{siamrcnn} introduces a two-stage Siamese re-detection framework for long-term visual tracking by leveraging a full-image search and a novel Tracklet Dynamic Programming Algorithm (TDPA). Unlike prior Siamese trackers that rely on local search windows around prior predictions, Siam R-CNN performs global re-detection across the entire frame. The second stage of the architecture compares ROI-aligned features of candidate regions with a first-frame template to determine object similarity using a three-stage cascade re-detection head. TDPA jointly considers re-detections from both the first frame and the previous frame to form spatio-temporal tracklets to allow for robust target association and distractor suppression over time. In addition, Siam R-CNN introduces a hard negative mining strategy, which retrieves visually similar objects from other videos to improve re-detection discriminability. This offline training strategy is highly effective for long-term tracking scenarios because of its robustness against significant appearance changes and occlusions.

To address the limitations of fixed template representations and independent feature extraction in Siamese trackers, SiamAttn \cite{siamattn} introduces a Deformable Siamese Attention (DSA) module into the Siamese architecture, which integrates deformable self-attention and cross-attention to enhance feature representations. The self-attention models intra-frame context via channel-wise and spatial operations. This is while the cross-attention aggregates interdependencies between the template and search regions to adaptively refine the target template. This implicit template update improves robustness against appearance variations, occlusions, and background clutter. Furthermore, SiamAttn introduces a region refinement module that performs depth-wise cross-correlation on attention-enhanced features and fuses them to refine both bounding box and segmentation mask predictions.

Ocean \cite{ocean} introduces a novel object-aware anchor-free tracking framework to overcome the limitations of anchor-based Siamese trackers, which often struggle when predefined anchor boxes poorly overlap with target objects. Instead of refining offsets from anchors, Ocean directly regresses the position and scale of the target using a dense prediction over all pixels within the ground truth bounding box to improve localization accuracy even in weak prediction scenarios. The method incorporates an object-aware feature alignment module, which aligns feature sampling with predicted bounding boxes. The method produces global and discriminative features to enhance classification reliability in parallel with a regular-region feature that captures localized detail. It then fuses both of these features to obtain robust target representations. In order to cope with appearance variations during inference, Ocean additionally supports online model updates. The combination of anchor-free regression and object-aware classification enables Ocean to achieve high robustness in cluttered and dynamic environments while maintaining real-time performance.

Traditional cross-correlation modules in Siamese trackers do not adequately account for channel importance or the local spatial information of the target, which limits the quality of similarity estimation and contributes to poor target representation under appearance variation or background clutter. ECIM \cite{ecim}  proposes an effective Efficient Correlation Information er, which decomposes the cross-correlation into Depthwise Cross-Correlation (DCC) and Pointwise Cross-Correlation (PCC) to capture both channel-wise semantic information and fine-grained local context. A novel correlation information er then fuses these two types of correlation maps via channel and spatial ing mechanisms to enhance the final representation for classification and anchor-free target state estimation. This approach improves the robustness and discriminability, particularly under complex scenes, while keeping computational cost low.

Most Siamese trackers keep the initial template fixed throughout the tracking sequence. As a result, they struggle to adapt to significant appearance variations of the target, often leading to tracking failure. In order to improve representation quality and adaptability, SiamDMU \cite{siamdmu} suggests a dual-mask template update approach. It builds upon the SiamRPN++ framework and consists of a Siamese Matching Block and a Template Updating Module (TUM). The TUM is composed of a Mask Enhancing Block (MEB) and a Template Updating Block (TUB). MEB refines the basic template and tracking outputs at predetermined intervals by utilizing semantic segmentation and long-term motion information. TUB then updates the template at the image level using these enhanced representations, thereby preserving high-resolution spatial details that are typically lost in feature-level updates. This approach facilitates robust tracking under severe appearance changes while remaining lightweight and easy to train. The final tracking result is obtained via a region proposal network head that performs pair-wise correlation.

\begin{figure}[h!]
\centering
\includegraphics[width=1.0\textwidth]{ 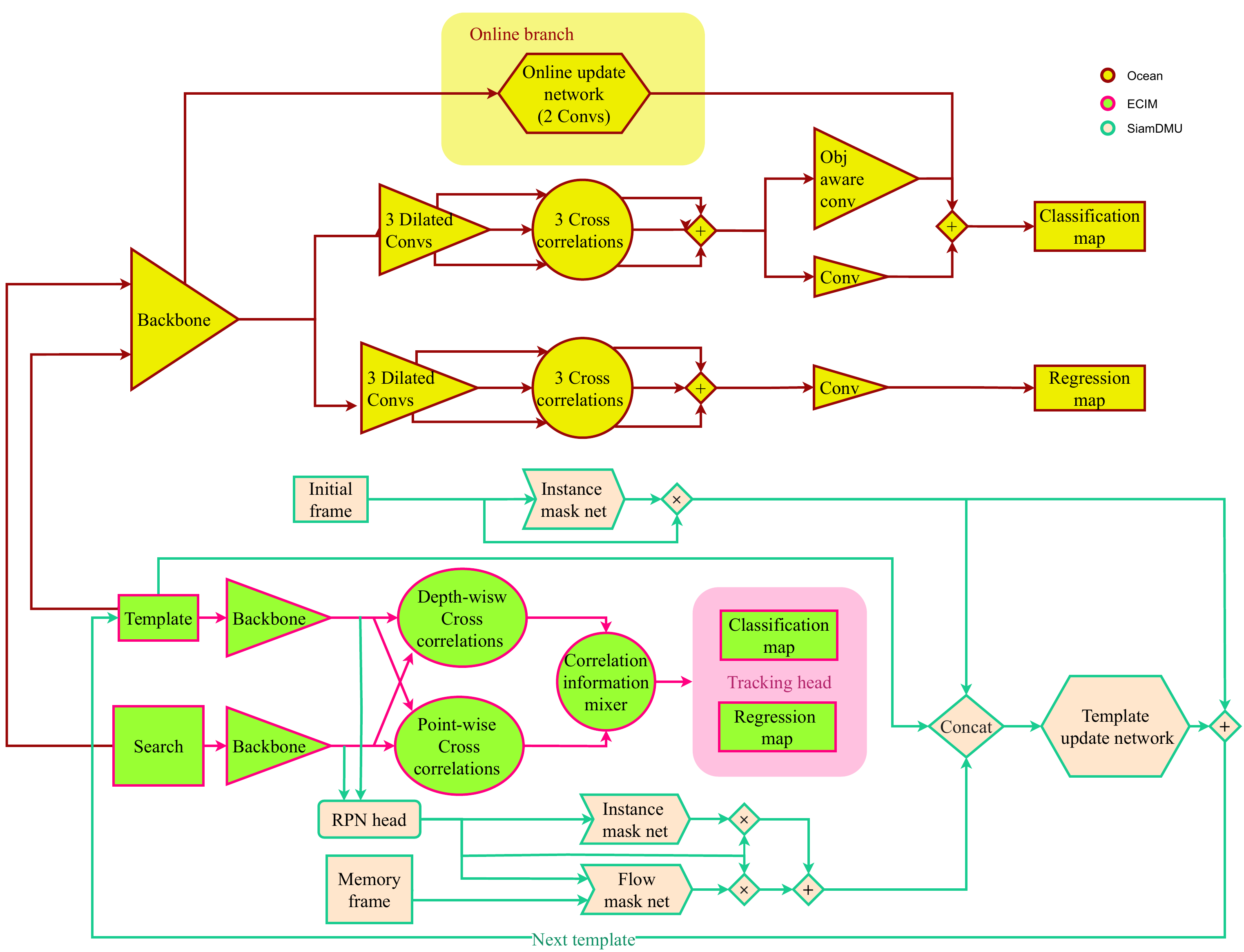}
\caption{Visual overview of more advanced Siamese-based tracking frameworks. This figure highlights the progression of Siamese trackers via incorporating better online update mechanisms in Ocean \cite{ocean}, appying memory in SiamDMU \cite{siamdmu}, and novel correlation-filter operations in ECIM \cite{ecim}. Additional architectural innovations and contributions of these methods can also be inferred from the figure.}\label{siam_group3}
\end{figure}

A high-level architectural comparison of more advanced reviewed Siamese-based trackers is provided in Figure.~\ref{siam_group3}, which highlights their progression and key innovations including various types of cross-correlation, memory integration, and online update mechanisms. This visual overview shows how the functionality and complexity of Siamese-based tracking architectures for more accurate online adaptation, and improving discriminability of the model. 

\begin{sidewaystable}
\tiny
\centering
\caption{A Detailed Comparison of Siamese-Based Trackers.}
\label{siamtab}
\begin{tabular}{%
    >{\centering\arraybackslash}p{0.5cm}|  
    p{1.4cm}  
    p{0.5cm}  
    p{1.7cm}  
    p{2.5cm}  
    p{2.5cm}  
    p{2.5cm}  
    p{2.5cm}  
    p{1cm}    
    p{2cm}    
}
\midrule
\multirow{13}{*}{\rotatebox{90}{\parbox[c]{12cm}{\centering \textbf{Siamese-Based} \\ \textbf{Appearance Model}}}} & 
\textbf{Method} & \textbf{Year} & \textbf{Backbone} & \textbf{Design Highlight} & \textbf{Focus} & \textbf{Novelty} & \textbf{Drawbacks} & \textbf{Template Update} & \textbf{Architectural-Level of Contribution} \\
\cmidrule{2-10}
& SiamFC \cite{siamfc} & 2016 & AlexNet \cite{alexnet} & Cross-correlation fusion; End-to-end offline training; Cosine window suppression; Multi-scale search & Generic similarity learning; addresses online-only limitations & First cross-correlation in Siamese networks & No update; no bbox regression; weak to appearance/scale variation & No & Appearance model \\
\cmidrule{2-10}
& DSiam \cite{dsaim} & 2017 & AlexNet\cite{alexnet}/ VGG19 \cite{vgg} & Dynamic transformation learning; FFT-based update; Elementwise feature fusion & Appearance variation and clutter handling & Online target/background transform learning; deep feature fusion & No bbox regression; shallow backbone & Yes & Online update; Feature representation \\
\cmidrule{2-10}
& SiamRPN \cite{siamrpn} & 2018 & AlexNet \cite{alexnet} & RPN-based regression; Meta-learning perspective; No multi-scale needed & Scale-aware proposals; Robust one-shot detection & End-to-end meta-detection tracking & Anchor sensitivity; complex hyperparams; lacks update & No & Target state estimation; Online update \\
\cmidrule{2-10}
& DaSiamRPN \cite{dasiamrpn} & 2018 & AlexNet \cite{alexnet} & Distractor-aware sampling/training; Local-to-global re-detection & Handle distractors and long-term tracking & Negative pair sampling; online distractor suppression & Shallow features; weak to extreme appearance changes & Yes & Appearance model; Feature representation; Online update \\
\cmidrule{2-10}
& SA-Siam \cite{sasiam} & 2018 & AlexNet \cite{alexnet} & Dual-branch Siamese; Semantic attention; Score fusion & Improve generalization via appearance + semantics & Channel-attentive semantic fusion & No update in appearance branch; shallow features & No & Feature representation; Online update \\
\cmidrule{2-10}
& SiamRPN++ \cite{siamrpnpp} & 2019 & ResNet-50~\cite{resnet}/ MobileNet \cite{mobilenets} & Deep backbone; Depthwise correlation; Multi-level aggregation & Translation invariance; Deep feature generalization & Deep Siamese with spatial-aware sampling & Anchor prior needed; no update; high complexity & No & Feature representation; Appearance model \\
\cmidrule{2-10}
& SiamFC++ \cite{siamfc} & 2019 & AlexNet\cite{alexnet}/ GoogLeNet \cite{googlenet} & Anchor-free design; Decoupled cls/reg; Quality assessment & Eliminate anchor ambiguity; improve reliability & Quality-aware anchor-free Siamese tracking & No update; limited occlusion/deformation handling & No & Target state estimation \\
\cmidrule{2-10}
& SiamBAN \cite{siamban} & 2020 & ResNet-50 \cite{resnet} & Fully conv anchor-free regression; 4D offset map; Multi-level output & Eliminate anchor tuning; scale generalization & Unified end-to-end box adaptive network & No template update; occlusion-sensitive & No & Target state estimation \\
\cmidrule{2-10}
& Siam R-CNN \cite{siamrcnn} & 2020 & ResNet-101-FPN \cite{resnet} & Two-stage cascade with TDPA; Hard negative mining; Full-image re-detection & Long-term tracking; drift/distractor suppression & Tracklet DP; cross-video hard sample mining & High computational cost; not real-time & No & Appearance model \\
\cmidrule{2-10}
& SiamAttn \cite{siamattn} & 2020 & ResNet-50 \cite{resnet} & Deformable attention with spatial/channel modules; Depth-wise correlation & Robustness to deformation and occlusion & Implicit template update via cross-attention & Increased complexity; slightly slower & Yes & Feature representation; Online update \\
\cmidrule{2-10}
& Ocean \cite{ocean} & 2020 & ResNet-50 \cite{resnet}  & Object-aware regression; Feature alignment; Dual-branch classification & Overcome anchor dependency; adaptive scale matching & Object-aware regression + aligned classification & Added complexity and training cost & Yes & Target state estimation \\
\cmidrule{2-10}
& ECIM \cite{ecim} & 2024 & Inception V3 \cite{inceptionv3} & Depthwise and pointwise correlation mixers; fusion via spatial/channel mixing & Address channel-insensitive similarity fusion & Correlation mixer for channel and local awareness & High computation; heuristic bbox regression & No & Appearance model; Feature representation \\
\cmidrule{2-10}
& SiamDMU \cite{siamdmu} & 2024 & ResNet \cite{resnet} & Dual-mask Template Update Module; RPN-head + semantic/motion cues & Handle static templates with motion/semantic updates & Image-level dual-mask update with DeepMask + FlowNet-C & Drift under distractors; RPN still correlation-based & Yes & Online update \\
\midrule
\end{tabular}
\end{sidewaystable}

\subsection{Transformer-based Tracking}\label{transformer}

Following our discussion of discriminative-based and Siamese-based trackers, We now discuss the expanding Transformer-based tracking technique family, which has experienced significant growth in recent years. Since their introduction in natural language processing for tasks like machine translation, transformers have shown remarkable results in a variety of vision applications, such as semantic segmentation, object detection, image classification, and point cloud analysis \cite{s8}. While Siamese-based trackers primarily focus on spatial information for tracking, and online methods incorporate historical predictions for model updates, both approaches lack an explicit mechanism to jointly model spatial and temporal relationships \cite{stark}. The ability of transformers to model both intra-frame and inter-frame dependencies through attention mechanisms makes them especially well-suited for visual tracking. Transformers employ global attention to capture long-range contextual information, in contrast to CNNs, which rely on local receptive fields \cite{alijani}. Transformer-based tracking uses key components such as encoder–decoder architectures, self-attention, and cross-attention to enhance feature representation and target localization. For further details on these components, we refer readers to \cite{vit, detr, s8}. We divide Transformer-based trackers into two primary categories: fully Transformer-based trackers, which offer completely new architectures based on Transformer principles, going beyond traditional tracking paradigms, and hybrid Transformer-based trackers, which expand upon Siamese or discriminative frameworks by adding Transformer modules to improve performance.

\subsubsection{Hybrid Transformer-based Trackers}\label{hybrid trans}

Transformer architectures have demonstrated outstanding performance across various vision tasks in recent years, motivating their integration into existing tracking frameworks. In the field of GOT, several approaches have emerged that enhance Siamese-based or discriminative-based trackers with transformer components, which are referred to in this section as hybrid transformer-based trackers. By including transformer blocks into various model stages like feature fusion and prediction model, these techniques aim to address challenges in CNN-based designs, such as limited receptive fields, limited global context modeling, or weak feature interactions. As a result, they achieve robustness to distractors and occlusions, better target-background discrimination, and better long-range dependency modeling. In this section, we analyze key hybrid transformer-based trackers by highlighting how they integrate transformers into tracking pipelines, what challenges they address, and their novelties, followed by their architectural illustrations. Besides, the important features of hybrid transformer-based trackers are summarized in Table \ref{hybridtab}.

\begin{figure}[hbt!]
\centering
\includegraphics[width=1.0\textwidth]{ 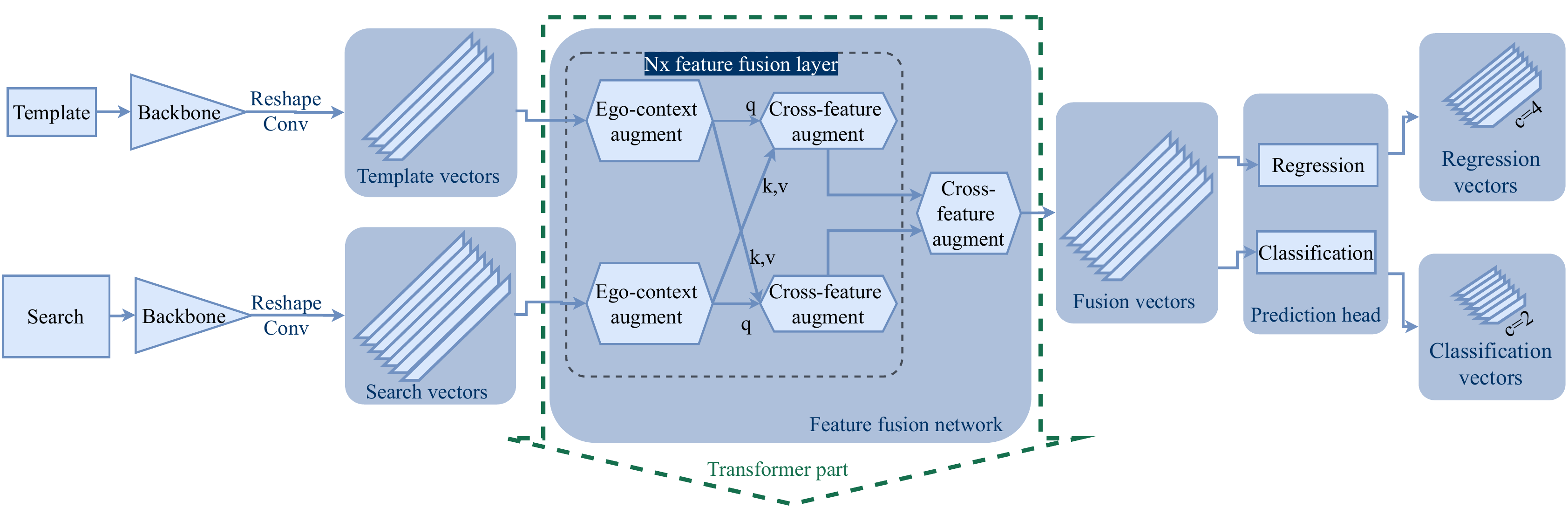}
\caption{Visual overview of TransT \cite{transt}, a hybrid transformer-based tracking frameworks, incorporating transformer into relation modeling stage of Siamese-based architecture.}\label{transt}
\end{figure}

TransT \cite{transt} shown in Figure.~\ref{transt} is an early effort to incorporate transformer architectures into the field of GOT. It fully replaces the traditional correlation-based feature fusion in Siamese frameworks with a pure attention-based design to better capture global context and preserve semantic information during the integration of template and search region features. The core idea of TransT lies in its feature fusion network, which is composed of ego-context augment (ECA) modules based on multi-head self-attention and cross-feature augment (CFA) modules utilizing multi-head cross-attention. These components are applied repeatedly to progressively enhance localization and boundary awareness. The ECA modules enrich feature representations within each branch, while the CFA modules enable deep interaction between template and search features. This design allows TransT to achieve robust performance under occlusions, appearance changes, and similar object interference.

Conventional trackers often treat video frames independently or rely on weak heuristics such as cosine windows or frame-wise updates to apply temporal information, which fail to capture deep temporal dependencies. TrDiMP and TrSiam \cite{trsiamtrdimp} solve this problem by extending discriminative and Siamese trackers with transformer architecture to model rich temporal dependencies across video frames. This paper designs a parallel encoder-decoder transformer framework in which the encoder applies self-attention to enhance template features across multiple frames, while the decoder propagates both spatial masks and features from the historical templates to the current search region. To maintain consistency between branches, attention weights are shared across the encoder and decoder, and a lightweight single-head attention design ensures computational efficiency. This architecture generalizes well across both Siamese and discriminative tracking pipelines, yielding TrSiam and TrDiMP variants. Both trackers improve robustness to appearance changes and occlusion, benefit from improved temporal modeling, online template updating, and fully end-to-end training.

Optimization-based discriminative trackers like DiMP \cite{dimp} rely on rigid objective minimization over limited past frames, which constrains model flexibility due to inherent inductive biases and also prevents the incorporation of test frame information during model prediction. ToMP \cite{tomp} addresses these limitations by replacing the traditional model optimizer with a transformer-based model predictor capable of modeling global context across both training and test frames. This design enables transductive target model prediction and facilitates richer feature representation through attention-based reasoning. In addition, by encoding target location and extent, ToMP injects spatial priors into the training features, allowing the transformer to more effectively model the target from background regions. Furthermore, it jointly predicts the weights for both target classification and bounding box regression through a unified transformer decoder in parallel. These weights are then applied to globally enhanced test frame features, resulting in robust localization and precise target estimation. The architecture achieves significant improvements over prior optimization-based methods and other transformer-enhanced trackers.

\begin{figure}[hbt!]
\centering
\includegraphics[width=1.0\textwidth]{ 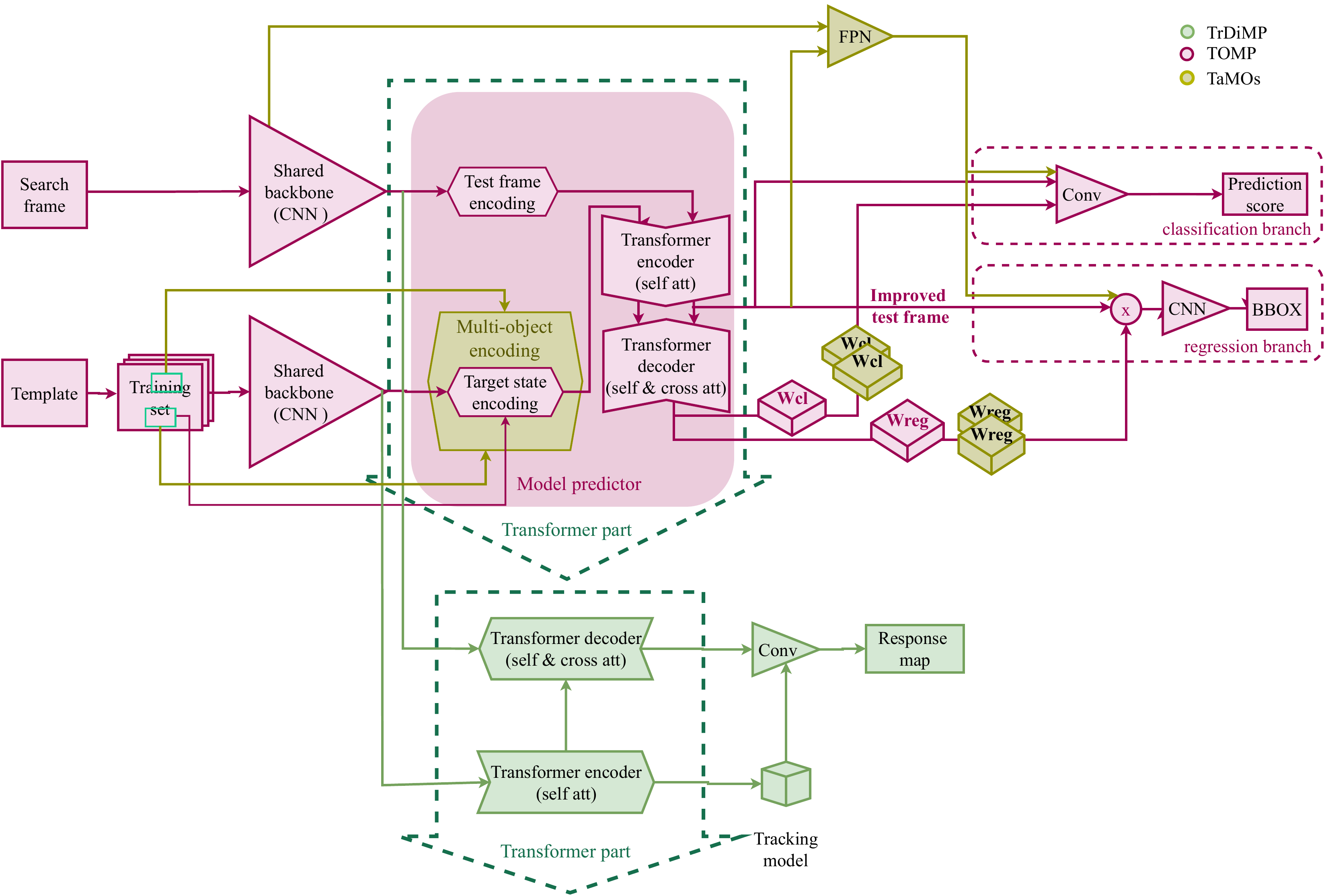}
\caption{Visual overview of TrDiMP \cite{trsiamtrdimp}, TOMP \cite{tomp}, and TaMOs \cite{tamos} hybrid transformer-based trackers. This figure illustrates how the reviewed methods incorporate transformers into discriminative-based architectures within the relation modeling stage. It also highlights the extension of  single generic object tracking to multi generic object tracking by defining multiple target models and applying multi-object encoding \cite{tamos}.} \label{hybrid_group2}
\end{figure}

In many real-world applications and scenarios, it is required to track multiple arbitrary objects simultaneously. TaMOs \cite{tamos} addresses this challenge by extending ToMP \cite{tomp} to multiple generic object tracking. This model introduces a transformer-based architecture capable of handling full-frame inputs and jointly predicting multiple target models through shared computation. Besides, TaMOs applies a global search strategy by constructing a unified feature representation for all targets instead of relying on localized crops for each object. To improve localization accuracy, particularly for small objects, it enhances the transformer encoder output using a Feature Pyramid Network (FPN), which fuses low-resolution test frame features with high-resolution backbone features. In addition, TaMOs proposes a novel multi-object encoding strategy, where every target is associated with a unique learnable embedding. The transformer decoder is then conditioned by these embeddings to predict target-specific models in a single forward pass. This shared tracking pipeline enables robust inter-object reasoning, reduces computational redundancy, and improves resilience against distractors in cluttered scenes. The authors of this paper also introduce a large-scale benchmark for multiple generic object tracking, LaGOT, which is based on the GOT framework \cite{got10k} to enable development of efficient trackers in diverse, real-world scenarios. Figure.~\ref{hybrid_group2} shows how methods apply transformers into discriminative-based trackers. 

\begin{figure}[t!]
\centering
\includegraphics[width=0.9\textwidth]{ 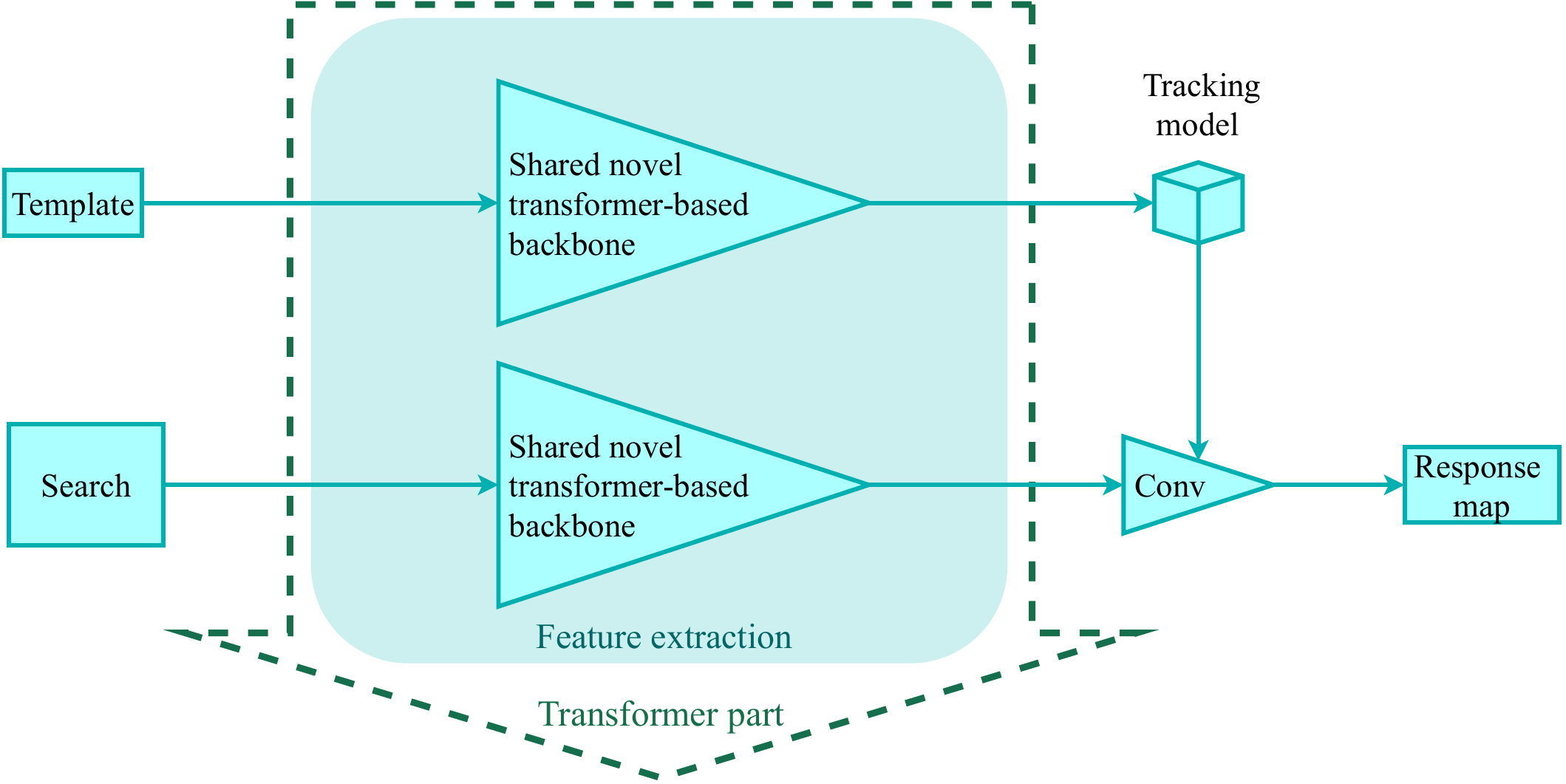}
\caption{Architecture overview of CMAT \cite{cmat} hybrid transformer-based tracking framework, which shows  This figure illustrates how it applies transformers in feature extraction and relation modeling in unified manner.}\label{cmat}
\end{figure}

CMAT \cite{cmat} shown in Figure.~\ref{cmat} proposes a novel feature extraction backbone for visual tracking by integrating CNN and transformer paradigms in a unified architecture in order to benefit from their complementary aspects. It proposes an aggregation module called CMAagg to integrate the strengths of convolutional layers in capturing local information and self-attention in modeling global dependencies. CMAT includes a convolutional mixer, which is built upon depthwise and pointwise convolutions to minimize local redundancy and improve efficiency. It also avoids redundant computation and improves representational quality by sharing the projection operation across both template and search branches. Afterwards, the outputs of the convolutional and self-attention paths are fused using learnable weights, and a dropout layer is added to enhance generalization and avoid overfitting. The resulting architecture effectively extracts both fine-grained local and broad contextual features without requiring online updates or adaptive model tuning during tracking.

 This section highlights how transformer modules have been used to enable more adaptable, context-aware, and scalable tracking architectures to address the fundamental drawbacks of previous trackers, including static model weights, limited temporal context, and ineffective per-target computation. 

\begin{sidewaystable}
\tiny
\centering
\caption{A Detailed Comparison of Hybrid Transformer-Based Trackers}
\label{hybridtab}
\begin{tabular}{%
    >{\centering\arraybackslash}p{0.5cm}|  
    p{1.4cm}  
    p{0.5cm}  
    p{1.3cm}  
    p{2.5cm}  
    p{2.5cm}  
    p{2.5cm}  
    p{2.5cm}  
    p{1cm}    
    p{2cm}    
}
\midrule
\multirow{5}{*}{\rotatebox{90}{\parbox[c]{7cm}{\centering \textbf{Hybrid Transformer-Based} \\ \textbf{Appearance Model}}}} & 
\textbf{Method} & \textbf{Year} & \textbf{Backbone Network} & \textbf{Design Highlight} & \textbf{Focus} & \textbf{Novelty} & \textbf{Drawbacks} & \textbf{Template Update} & \textbf{Architectural-Level of Contribution} \\
\cmidrule{2-10}
& TransT \cite{transt} & 2021 & ResNet-50 \cite{resnet} & Siamese-based feature extraction + Transformer-based fusion; Multi-head self/cross-attention with positional encoding & Global attention replaces correlation for long-range interaction; Robustness to distractors & Attention-only fusion addressing correlation limitations & No template update; heavier fusion; not fully transformer-based & No & Appearance model \\
\cmidrule{2-10}
& TrDiMP/ TrSiam \cite{trsiamtrdimp} & 2021 & ResNet-50 \cite{resnet} & Unified framework (Siamese + Discriminative); Transformer encoder-decoder for temporal context; Shared attention; Masking for background suppression & Temporal context modeling between frames & Online template update; attention-based propagation; background suppression & Slight increase in complexity & Yes & Feature representation \\
\cmidrule{2-10}
& ToMP \cite{tomp} & 2022 & ResNet-50 / ResNet-101 \cite{resnet} & Replaces DiMP optimizer with transformer predictor; Joint prediction via decoder; Transductive context usage & Address inductive bias and fixed modeling in DiMP & Learns classifier/regressor via transformer; test-frame-aware encoding & High memory; lacks distractor-specific handling & Yes & Feature representation, appearance model \\
\cmidrule{2-10}
& TaMOs \cite{tamos} & 2024 & ResNet-50 \cite{resnet} / Swin-Base \cite{swintransformer} & Extends ToMP for MOT; Full-frame encoding; joint multi-target decoding; FPN for resolution & Multi-target generic tracking & Multi-target encoder-decoder; target embedding pool; full-frame reasoning & High joint complexity & Yes & Feature representation, appearance model \\
\cmidrule{2-10}
& CMAT \cite{cmat} & 2024 & ResNet-50 \cite{resnet} & CMAagg: ConvMixer + self-attention; joint local-global modeling; shared projection & Efficient local-global representation with generalization & Hybrid convolution-attention backbone; reduces redundancy & No update; lacks template adaptation & No & Feature representation \\
\midrule
\end{tabular}
\end{sidewaystable}

\subsubsection{Fully Transformer-based Trackers}\label{fully trans}
Unlike hybrid trackers that apply transformer modules to conventional Siamese and discriminative-based tracking architectures, fully transformer-based trackers are not derived from these prior paradigms. Instead, they are built upon standalone transformer architectures designed from the ground up. While some of these methods may incorporate convolutional layers, they do not rely on the structural principles of Siamese matching or discriminative learning frameworks. These fully transformer-based trackers leverage the attention mechanism in self-attention and cross-attention as a fundamental building block throughout the tracking pipeline, such as feature encoding, relation modeling, feature fusion, and prediction. Based on their architectural design, fully transformer-based trackers can be broadly divided into two categories: \RNum{1}. Convolution-attention trackers, which combine convolutional priors with transformer-based reasoning, and \RNum{2}. Pure attention-based trackers, which rely exclusively on attention mechanisms. In this section, we review both categories in detail, highlighting their design choices, target representation strategies, and relation modeling techniques.

\vspace{18pt}

\textbf{Convolution-Attention Transformer Trackers:} The best-known methods in the convolution-attention transformer tracker field are described below with their corresponding structures in a cohesive and organized way. Furthermore, a detailed comparison of these fully convolution-attention transformer-based trackers is provided in Table~\ref{cnn attention tab}.

\begin{figure}[h!]
\centering
\includegraphics[width=1.01\textwidth]{ 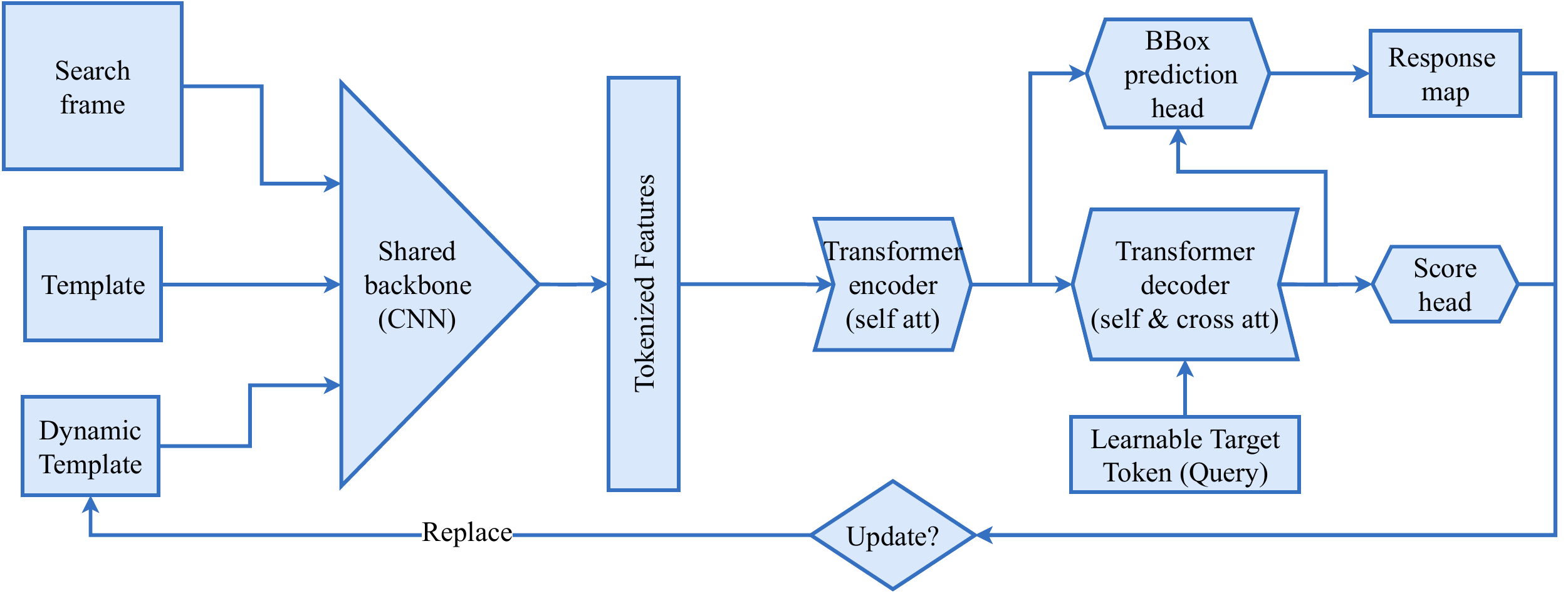}
\caption{Visual architecture of STARK \cite{stark} as a convolution-attention based Fully transformer trackers emphasizing its online update and transformer-based relation modeling.}\label{stark}
\end{figure}

Convolution-based trackers are only effective at modeling the local spatial or temporal neighborhood information but struggle to capture long-range dependencies. This limits their robustness under large-scale object variation, occlusion, and frequent appearance-reappearance. \citet{stark}, shown in Figure.~\ref{stark} addresses these limitations by introducing the STARK model with encoder-decoder transformer architecture. The encoder aims at reinforcing original features with long-range spatio-temporal encoding by jointly processing features from the initial template, a dynamically updated template, and the current search region, capturing global contextual relationships through multi-head self-attention. A lightweight decoder learns a single query embedding that attends to the encoded features to predict spatial position. For bounding box prediction, STARK proposes a fully convolutional corner-based head in order to directly estimate the probability distribution of the top-left and bottom-right corners. This strategy eliminates the need for proposals, predefined anchors, and the complicated post-processing with hyperparameters. A confidence-based score head controls the dynamic update of the template, ensuring adaptation only when reliable. This end-to-end framework simplifies tracking pipelines while improving accuracy and speed.

\begin{figure}[t!]
\centering
\includegraphics[width=1.0\textwidth]{ 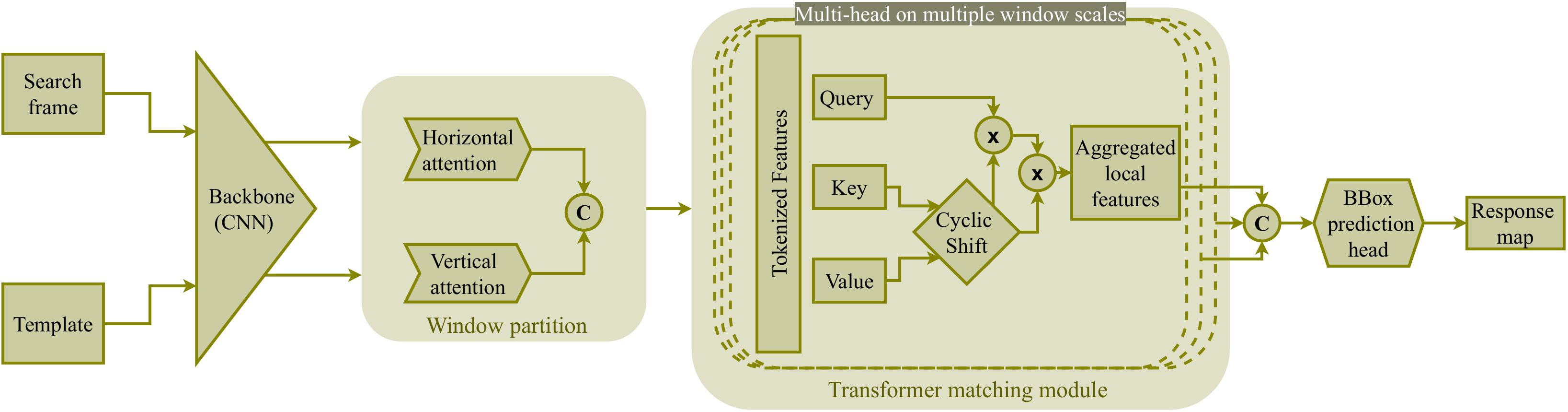}
\caption{Visual architecture of CSWinTT \cite{cswintt} with window-wise attention mechanism for object-oriented relation modeling.}\label{CSWinTT}
\end{figure}

Pixel-level attention in existing transformer-based trackers often breaks object integrity and loses relative positional information which makes it difficult to accurately match targets in cluttered scenes. To overcome these limitations, CSWinTT \cite{cswintt} in Figure.~\ref{CSWinTT} introduces a multi-scale cyclic shifting window attention mechanism that elevates attention computation from the pixel to the window level. Inspired by Swin Transformer \cite{swintransformer}, CSWinTT partitions template and search features into windows and performs attention between entire windows, thereby preserving object structure and enabling more localized yet robust matching at different scales. Each transformer head operates on a specific window scale, supporting fine-to-coarse matching granularity. To further improve accuracy, CSWinTT proposes a cyclic shifting strategy that generates diverse window samples by circularly translating windows. This is while a spatially regularized attention mask suppresses boundary artifacts caused by this shifting. Additionally, the model eliminates redundant computation through three efficiency-driven optimizations, enabling real-time tracking. The fused multi-scale features are passed through a corner-based prediction head to produce the final bounding box.

\begin{figure}[t!]
\centering
\includegraphics[width=1.01\textwidth]{ 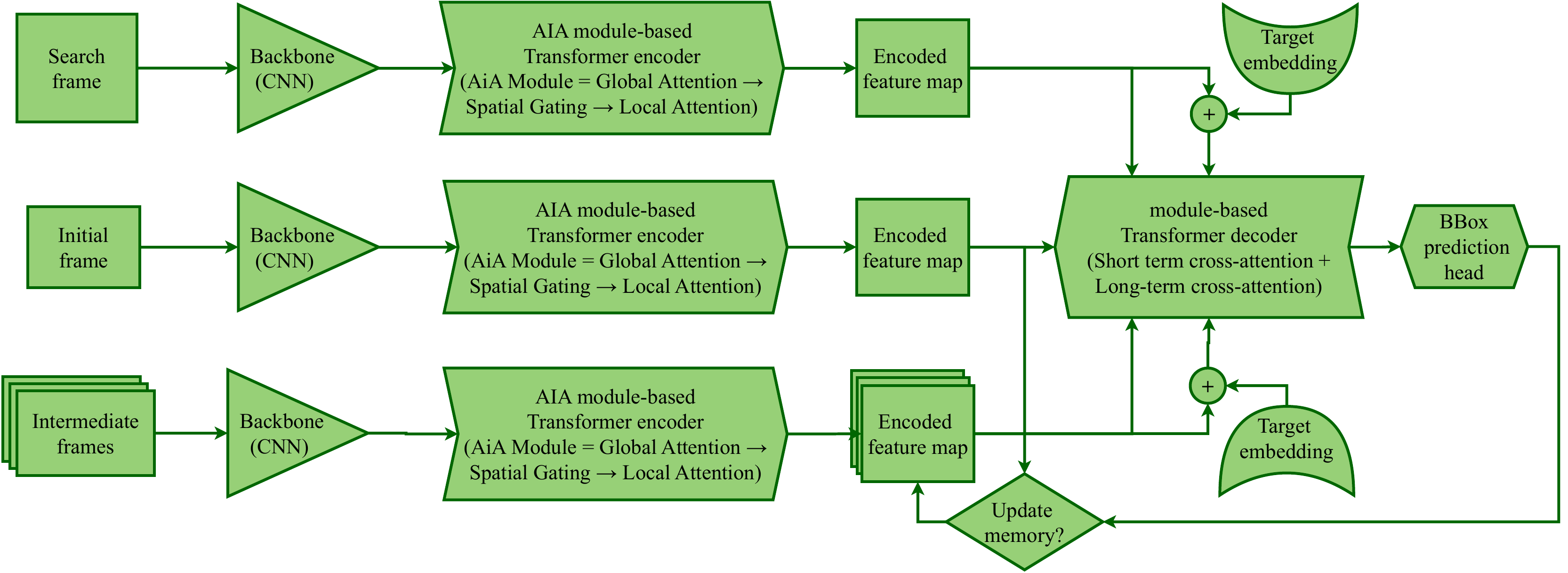}
\caption{Visual architecture of discriminative AiATrack \cite{aiatrack} with novel Attention-in-Attention (AiA) module with both short-term and long-term templates.}\label{aiatrack}
\end{figure}

Transformer-based trackers often suffer from noisy and ambiguous attention weights due to the independent computation of query-key correlations in attention mechanisms. Therefore, they fail to capture the contextual relationships among different query-key pairs which leads to unreliable attention especially in scenes with background clutter or imperfect feature representations. To overcome this limitation, AiATrack \cite{aiatrack} in Figure.~\ref{aiatrack}  introduces a novel Attention-in-Attention (AiA) module that enhances conventional attention by embedding an inner attention mechanism to refine the raw correlation maps. The AiA module operates on correlation vectors in order to find consensus among them, effectively amplifying reliable associations and suppressing erroneous ones. This module is integrated into both self-attention blocks to improve feature aggregation and cross-attention blocks to strengthen information propagation. In addition, AiATrack adopts an efficient feature reuse strategy to avoid repeated computations during online updates. It also incorporates a target-background embedding assignment mechanism that explicitly distinguishes the foreground target from the background while preserving contextual information. The tracker maintains a long-term template extracted from the initial frame, as well as a short-term template dynamically updated based on an IoU prediction head.

\begin{figure}[h!]
\centering
\includegraphics[width=1.0\textwidth]{ 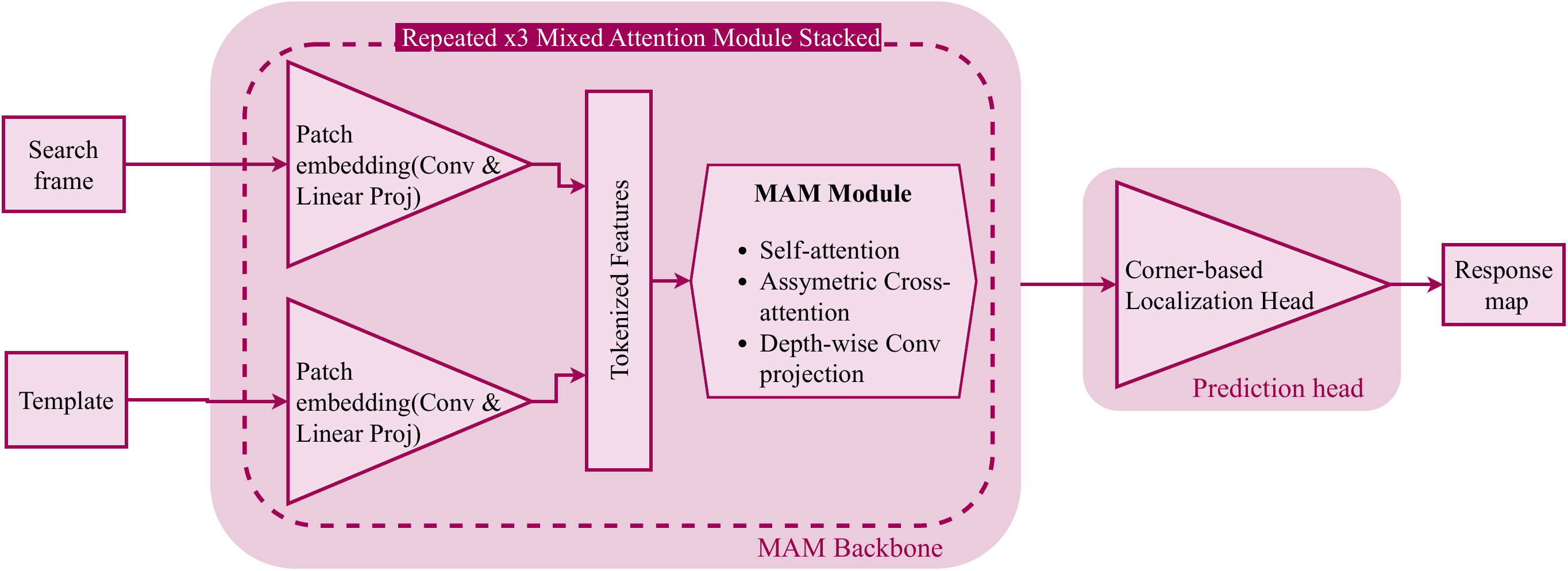}
\caption{End-t-end MixFormer visual architecture \cite{mixformer} with novel Mixed Attention Module (MAM).}\label{MixFormer}
\end{figure}

MixFormer \cite{mixformer} introduces a compact end-to-end architecture with unified tracking stages to solve high complexity and limited adaptability in dominant tracking frameworks, which often relied on multi-stage pipelines with separate modules for feature extraction, information integration, and localization. Central to this design is the Mixed Attention Module (MAM) that concurrently performs self-attention and cross-attention operations, enabling the extraction of long-range intra-frame dependencies while integrating target-specific information between the template and the search region. MixFormer applies CvT \cite{cvt} as its backbone, which utilizes a combination of transformers and convolutional layers to efficiently model both local and global representations. For better efficiency and distractor handling, an asymmetric attention scheme is introduced that selectively excludes cross-attention from the template to the search area. Shown in Figure.~\ref{MixFormer}, the overall framework consists of only a stacked MAM-based backbone and a lightweight corner-based localization head. During the inference, MixFormer incorporates a confidence-guided score prediction module that dynamically selects high-quality online templates to enhance robustness to appearance changes and occlusions.

\begin{sidewaystable}
\tiny
\centering
\caption{Detailed Comparison Of Fully Transformer-Based Trackers with Convolution-Attention Architectures}
\label{cnn attention tab}
\begin{tabular}{%
    >{\centering\arraybackslash}p{0.5cm}|  
    p{1.4cm}  
    p{0.5cm}  
    p{1.3cm}  
    p{2.5cm}  
    p{2.5cm}  
    p{2.5cm}  
    p{2.5cm}  
    p{1cm}    
    p{2.3cm}    
}
\midrule
\multirow{22}{*}{\rotatebox{90}{\parbox[c]{9cm}{\centering \textbf{Transformer-based Appearance Model} \\ \textbf{Convolution-Attention Architectures}}}} & 
\textbf{Method} & \textbf{Year} & \textbf{Backbone Network} & \textbf{Design Highlight} & \textbf{Focus} & \textbf{Novelty} & \textbf{Drawbacks} & \textbf{Template Update} & \textbf{Architectural-Level of Contribution} \\
\cmidrule{2-10}
& STARK \cite{stark} & 2021 & ResNet-50 / ResNet-101 \cite{resnet} & Spatio-temporal transformer encoder-decoder; corner-based box prediction; dynamic template update via a confidence-aware score head & Unified modeling of spatial and temporal information for robust long-term tracking & Introduction of NOTU largescake benchmark; Global spatio-temporal attention; Anchor-free corner prediction & No explicit memory; Limitation of CNN-based feature representation & Yes & \begin{tabular}[c]{@{}l@{}}Online update\\ Feature representation\\ Target state estimation\end{tabular}\\
\cmidrule{2-10}
& CSWinTT \cite{cswintt} & 2022 & ResNet-50 \cite{resnet} & Multi-scale cyclic shifting window attention; Spatially regularized attention mask for boundary effect; Applying dynamic and static templates & Window-level similarity modeling with robust object integrity preservation & Elevates attention from pixel to window level; Multi-scale matching with cyclic shift sampling; Spatial mask alleviates boundary artifacts & Relatively high computational complexity & Yes & Relation modeling \\
\cmidrule{2-10}
& AiATrack \cite{aiatrack} & 2022 & ResNet-50 \cite{resnet} & Introduces the Attention-in-Attention (AiA) module to refine attention maps; Applied AiA to both self- and cross-attention; Reuses cached encoded features instead of frames in online tracking; An IoU-guided short-term memory mechanism & Robust and efficient attention refinement to avoid noisy results; Temporal feature utilization in cluttered and dynamic scenes & AiA module for consensus-based correlation refinement; Target-background embedding assignment; Feature reuse strategy for efficient online adaptation; Dual-branch cross-attention with IoU-guided short-term reference updates & Sensitive to incorrect IoU estimation during template update & Yes & Feature representation; online update \\
\cmidrule{2-10}
& MixFormer \cite{mixformer} & 2022 & CVT \cite{cvt} & Joint feature learning \& relation modeling via MAM module; Solely based on multiple stacked MAM modules as backbone; Asymmetric attention to lower computational cost \& a Corner Based prediction Head; Score based template update & Joint spatial and temporal feature integration via transformers with minimal computational overhead to improve discriminability & Handle multiple target templates during online tracking; Mixed attention for joint target integration and feature extraction; Query-aware template filtering via score prediction & High Training Overhead; Computational complexity & Yes & Feature representation; Relation modeling \\
\midrule
\end{tabular}
\end{sidewaystable}

\vspace{12pt}

\textbf{Pure Attention-Based Transformer Trackers:} Existing transformer-based trackers often rely on CNN backbones for feature extraction, limiting the full potential of transformers in representation learning. In this section, we discuss transformer-based trackers which are fully based on transformers and attention layers aiming at fully leverage their spatio-temporal modeling capabilities for better performance. The pure attention-based transformer trackers are categorized into one-stream, two-stram, box based token based, video transformer based, memory based and prompt based methods will be explained in order.
Existing transformer-based trackers often rely on CNN backbones for feature extraction, which restricts the representational capacity and end-to-end modeling potential of transformers. In this section, we focus on fully transformer-based trackers that eliminate CNN components and rely entirely on attention mechanisms. These methods aim to fully exploit the spatio-temporal modeling capabilities of transformers for improved tracking performance.

In the following, the most prominent methods in the field of pure attention-based transformer trackers are explained. This is along with their corresponding architectures in a unified and structured manner to facilitate easier comparison and analysis. Furthermore,
a detailed comparison of reviewed fully pure attention transformer-based trackers is provided in Table~\ref{fullytab}.

\begin{figure}[t!]
\centering
\includegraphics[width=1.0\textwidth]{ 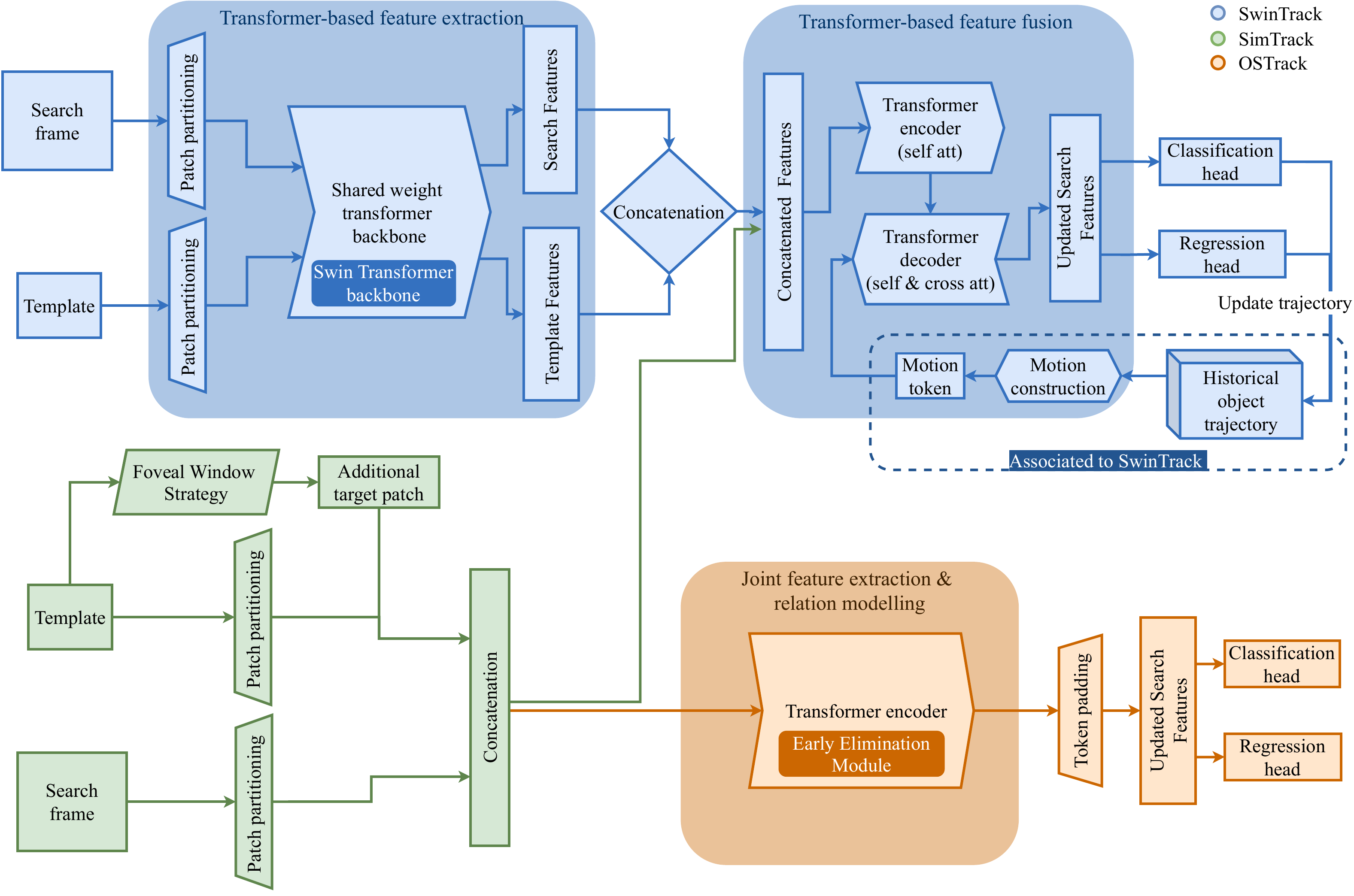}
\caption{Visual overview of earlier pure attention-based fully transformer trackers emphasizing fully tranformer-based feature extraction and relation modeling in SwinTrack \cite{swintrack} and joint feature extraction and fusion in simtrack \cite{simtrack} and OSTrack \cite{ostrack}. This figure also highlights additional components such as motion information, central object tokens, and elimination modules providing more accurate and efficient trackers.} \label{puregroup1}
\end{figure}

SwinTrack \cite{swintrack} in Figure.~\ref{puregroup1} proposes a fully attentional tracking framework built on the Swin Transformer architecture, in which both feature representation learning and fusion are conducted using attention mechanisms. This leads to more compact and semantic-aware feature representation to localize the target object. Within a simplified framework, template and search region features are concatenated and passed through a shared Swin Transformer backbone to enable joint modeling. To further enhance robustness without explicit online updates, SwinTrack introduces a motion token that captures the historical trajectory of the target within a local temporal window. During inference, this token is added to the attention mechanism of the decoder to improve temporal awareness and make it easier to find the target under motion. The lightweight decoder is applied for vision-motion fusion and a dual-branch prediction head. Notably, SwinTrack avoids complex designs like multi-scale features or query-based decoders, offering simplicity, efficiency, and strong performance.

Instead of relying on complex architectures with separate feature extraction and interaction stages, SimTrack \cite{simtrack}, shown in Figure.~\ref{puregroup1}, introduces a simplified transformer-based architecture that unifies joint feature learning and interaction within a one-branch transformer backbone to improve model flexibility and efficiency. By serializing and concatenating the exemplar and search images before feeding them into the backbone, the model allows bidirectional attention across all layers, enabling multi-level and more comprehensive interaction between them. To prevent information loss as a result of patch downsampling, SimTrack proposes a foveal window strategy that emphasizes the central region of the exemplar by sampling diverse, target-focused patches. This significantly improves tracking accuracy while maintaining computational efficiency. The architecture removes specialized modules, reduces training complexity, and generalizes well across tracking tasks.

Two-stage trackers extract features from the template and search regions independently and fuse them later for relation modeling, leading to weak target awareness and limited target-background discriminability. To address this, OSTrack \cite{ostrack} proposes a one-stream, one-stage transformer framework that unifies feature extraction and relation modeling by allowing bidirectional information flow between the template and search at the earliest stage efficiently visualized in Figure.~\ref{puregroup1}. By directly concatenating both inputs, the model enables simultaneous learning of target-aware features through self-attention which eliminates the need for separate cross-attention modules. In addition, an early candidate elimination module is integrated into selected encoder layers to enhance efficiency via identifying and discarding background tokens based on a free similarity score derived from attention weights. This reduces computational cost and suppresses distractor interference. A restoration mechanism reorders remaining tokens and pads discarded ones to preserve spatial alignment for bounding box prediction. 

\begin{figure}[t!]
\centering
\includegraphics[width=0.8\textwidth]{ 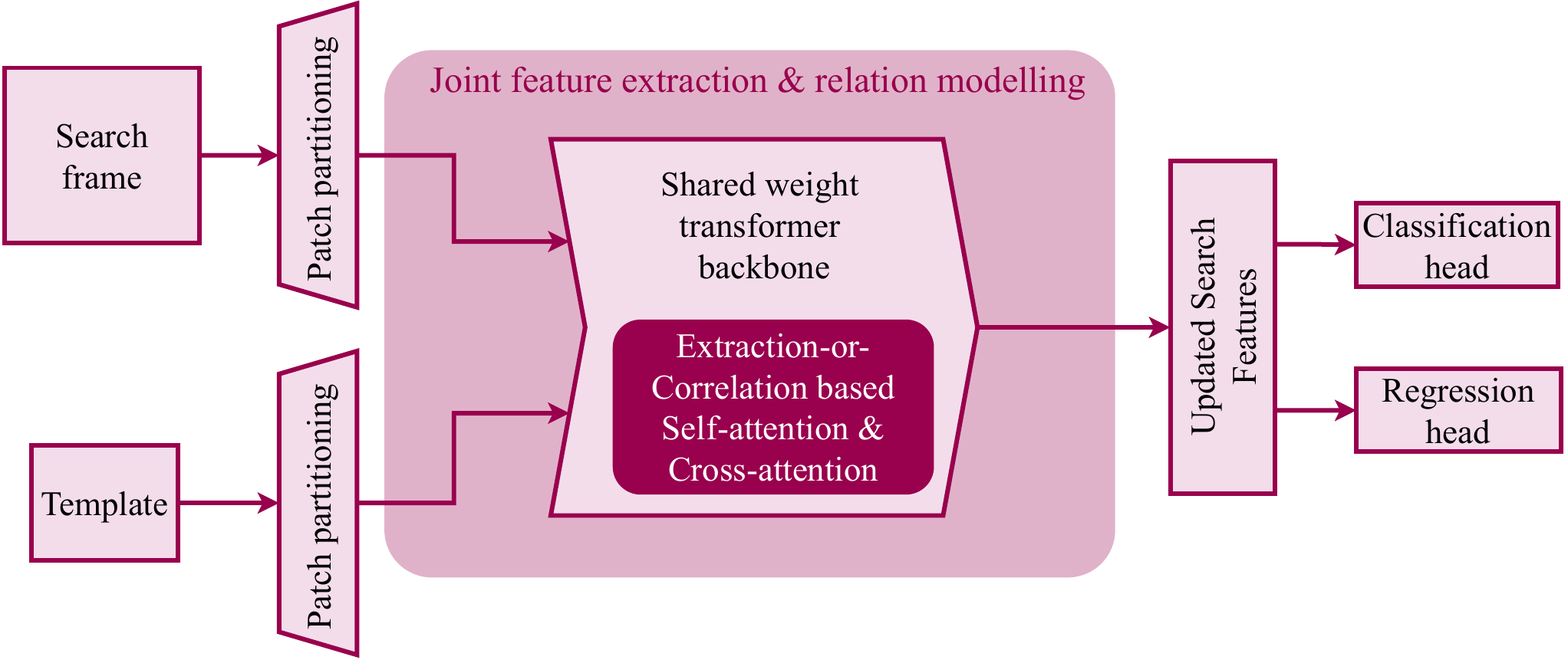}
\caption{Visual overview of single-branch SBT \cite{sbt} which applies joint feature extraction and relation modeling like OSTrack \cite{ostrack} but through a novel and more discriminative transformer-based backbone.} \label{SBT}
\end{figure}

Most modern trackers depend on separate modules for feature extraction and correlation, which often introduces architectural complexity and limits the discriminative power of extracted features, especially in the presence of distractors. To address this, \citet{sbt} introduced the Single Branch Transformer (SBT) with a novel target-dependent feature network that deeply embeds correlation through hierarchical self-attention and cross-attention blocks during feature extraction. By unifying the processing of template and search images in a single-stream transformer backbone, SBT enables deep interaction between the two inputs, resulting in dynamic and instance-specific feature representations. This effectively enhances target coherence while suppressing distractor interference. At the core of SBT is the Extract-or-Correlation (EoC) block, which alternates between self-attention and cross-attention operations. The self-attention modules improve intra-image features, while cross-attention modules progressively align inter-image features to filter out irrelevant regions and refine the representation for robust matching. This joint processing mechanism allows SBT to differentiate the target from distractors while maintaining temporal and spatial consistency. At the prediction level, the fully fused features of the search image are directly fed into a classification and regression head to generate the target's localization and size embeddings. This eliminates the need for an explicit correlation step found in prior trackers. The architecture of SBT is provided in Figure.~\ref{SBT}.

\begin{figure}[t!]
\centering
\includegraphics[width=1\textwidth]{ 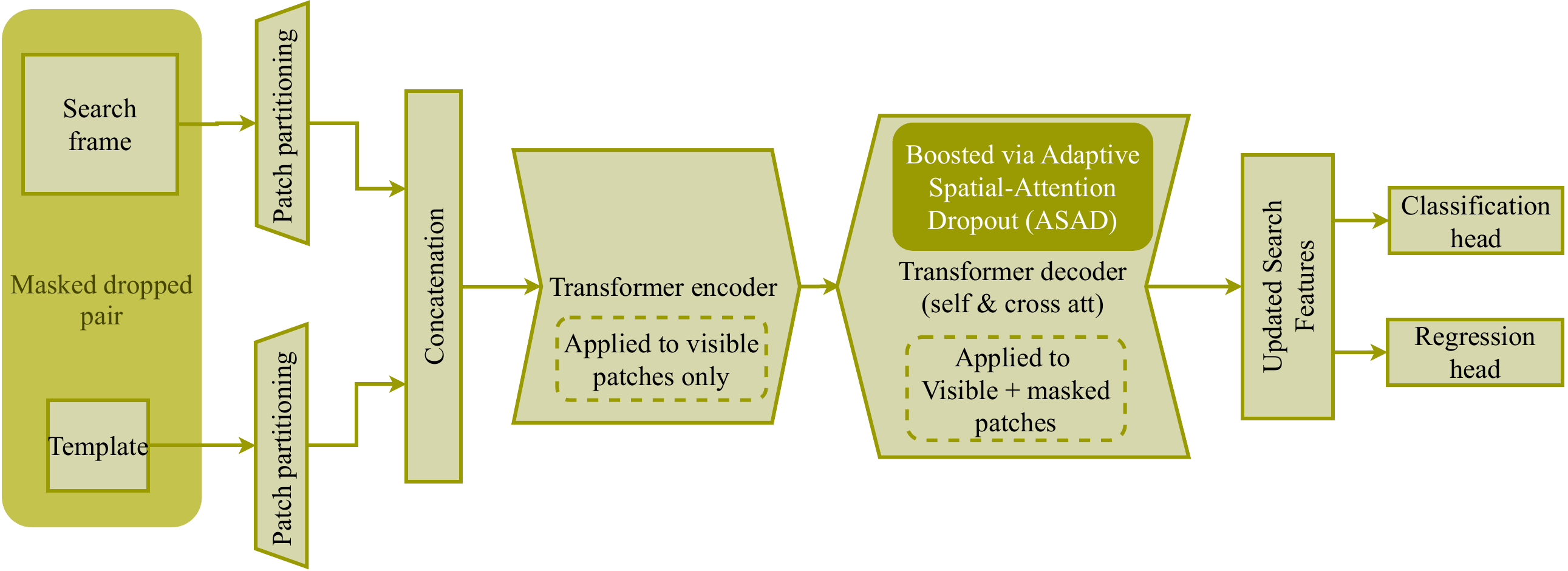}
\caption{Visual overview of DropMAE \cite{dropmae} emphasizes on applying drop masking to achieve more discriminative search features} \label{dropmae}
\end{figure}

Most masked autoencoder (MAE)-based ViT trackers \cite{ostrack,simtrack} rely heavily on spatial cues from static images, which limits their ability to capture temporal correspondences crucial for robust video object tracking. To address this limitation, DropMAE \cite{dropmae} in Figure.~\ref{dropmae} introduces a novel self-supervised video pre-training strategy via Adaptive Spatial-Attention Dropout (ASAD). ASAD enhances temporal correspondence learning during masked patch reconstruction by selectively dropping spatial attention weights from within-frame token interactions in order to force the model to depend more on between-frame cues. This encourages the encoder to learn temporally aligned representations without architectural modifications to the transformer backbone. DropMAE operates on video frame pairs by incorporating frame identity embeddings to distinguish between temporally adjacent frames. It is also compatible with existing ViT-based trackers. The authors of this paper highlight that pre-training on videos with diverse motion patterns is more beneficial than scene diversity for temporal matching tasks. Another example of applying a masked autoencoder to tracking is MAT (\cite{mat}), which uses random masking to encourage the learning of discriminative features. However, it introduces a masked appearance transfer framework that jointly encodes the template and search region and reconstructs the template as it appears within the search image. This nontrivial reconstruction objective enables the model to learn more discriminative, target-aware features, highlighting its potential for improving feature representations in trackers.

\begin{figure}[t!]
\centering
\includegraphics[width=1.0\textwidth]{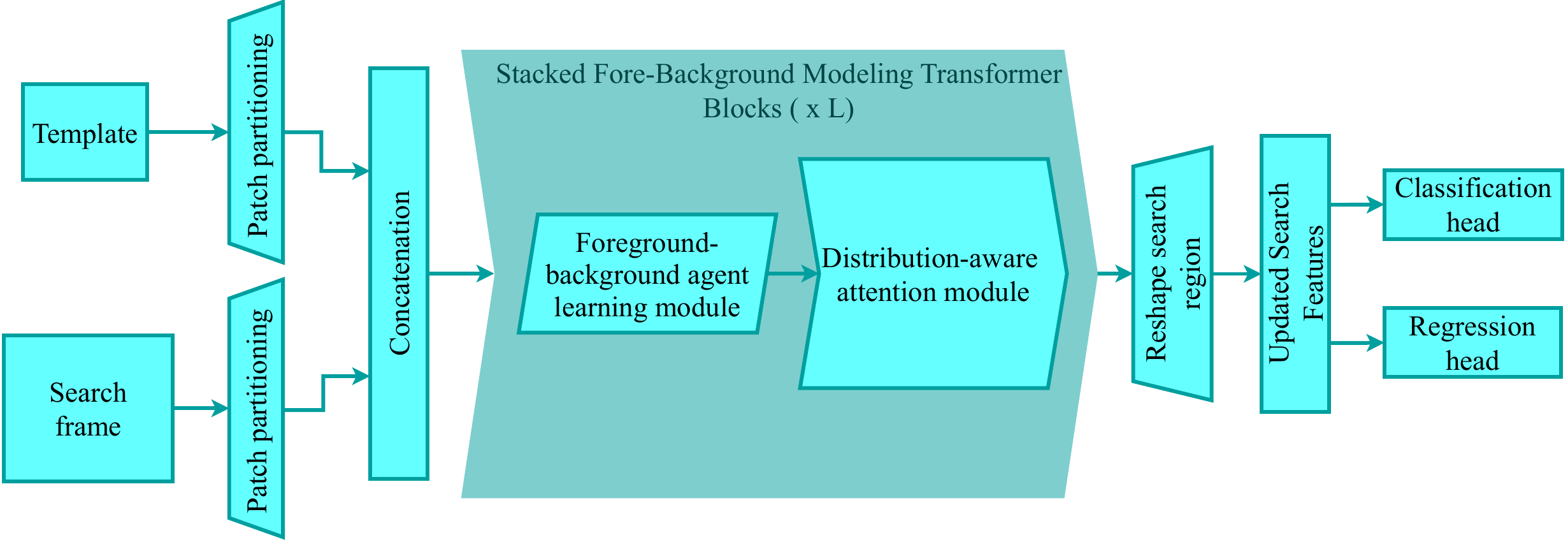}
\caption{ F-BDMTrack \cite{f-bdmtrack} architecture via Foreground-Background Distribution Modeling Transformer to achieve more discriminative power in tracking.} \label{fbdmtrack}
\end{figure}

While pure transformer-based trackers offer strong representation and interaction capabilities, they are often vulnerable to background clutter due to their reliance on appearance-based attention, which leads to inaccurate feature aggregation when foreground and background regions are visually similar. F-BDMTrack \citet{f-bdmtrack}, shown in Figure.~\ref{fbdmtrack}, solved this by introducing a Foreground-Background Distribution Modeling Transformer that incorporates two novel components of the Fore-Background Agent Learning (FBAL) module and the Distribution-Aware Attention (DA2) module. The FBAL module learns dynamic fore-background agents from both the template and the search region using a pseudo-bounding box generation technique in order to model object-background separability. Rather than relying solely on direct feature similarity, the subsequent DA2 module improves attention computation by incorporating distribution-level comparisons between foreground and background representations. This enhances the aggregation of target-specific features in cluttered scenes. The overall framework requires no additional supervision or auxiliary modules and achieves improved discrimination and context awareness, particularly in challenging tracking scenarios.

\begin{figure}[t!]
\centering
\includegraphics[width=0.9\textwidth]{ 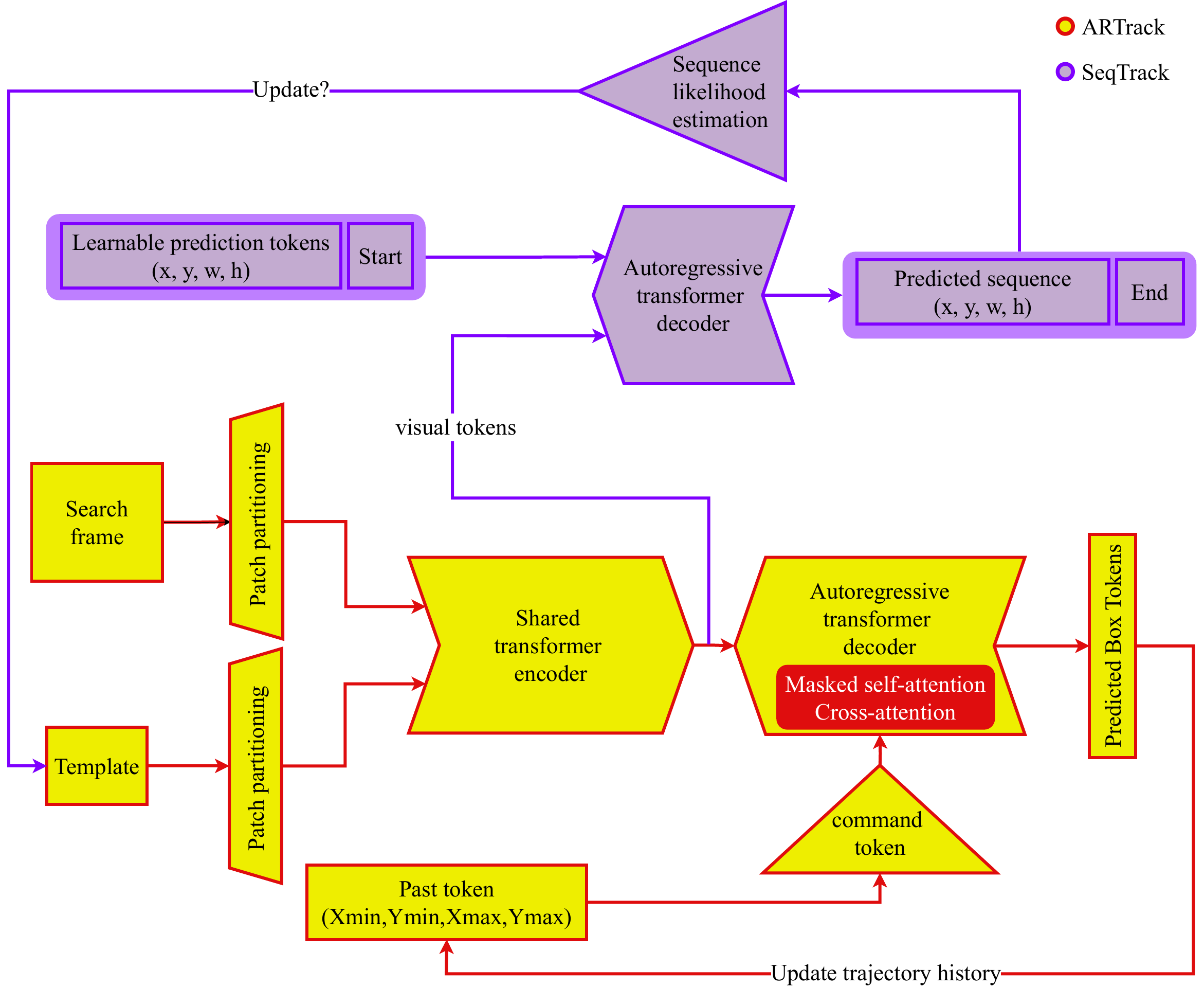}
\caption{Visual overview of sequence-level pure attention fully transformer trackers. This figure illustrates a coherent motion and spatio-temporal modeling in ARTrack \cite{artrack} and an autoregressive sequence-t-sequence learning in SeqTrack \cite{seqtrack} .} \label{puregroup2}
\end{figure}

To better leverage temporal coherence in tracking, ARTrack \cite{artrack} reformulates visual tracking as a coordinate sequence interpretation problem instead of conventional per-frame template matching, shown in Figure.~\ref{puregroup2}. More specifically, it proposes a novel, simple autoregressive framework that models object trajectories directly across frames. Inspired by language modeling, this model discretizes bounding box coordinates into token sequences and then leverages a transformer-based encoder-decoder architecture. ARTrack also conditions its predictions on spatio-temporal prompts, including past trajectory tokens and current frame features, allowing it to propagate motion dynamics for consistent localization. This sequence-level modeling unifies training and inference by maximizing sequence-level likelihood with a structured loss function. It also eliminates the need for handcrafted localization heads or complicated post-processing modules. The introduced design for ARTrack enables coherent motion modeling and consistent localization, making it an elegant and effective alternative to conventional frame-by-frame approaches.

Most previous GOT trackers decompose the task into two subtasks of classification and regression, each handled by separate head networks and loss functions increasing architectural complexity and training overhead. SeqTrack \cite{seqtrack}, presented in Figure.~\ref{puregroup2}, overcomes this challenge by introducing a novel sequence-to-sequence learning framework that formulates object tracking as an autoregressive sequence generation task. Instead of predicting bounding boxes through handcrafted heads, this method discretizes bounding box coordinates into token sequences and learns to generate them using a plain encoder-decoder transformer. The encoder jointly extracts features from both template and search images, while the causal decoder autoregressively predicts the bounding box tokens. This design is trained end-to-end with a simple cross-entropy loss, eliminating the need for complex supervision. SeqTrack also incorporates online template update using a confidence-driven token likelihood mechanism and applies a window penalty during inference to enhance localization stability.

\begin{figure}[t!]
\centering
\includegraphics[width=1.0\textwidth]{ 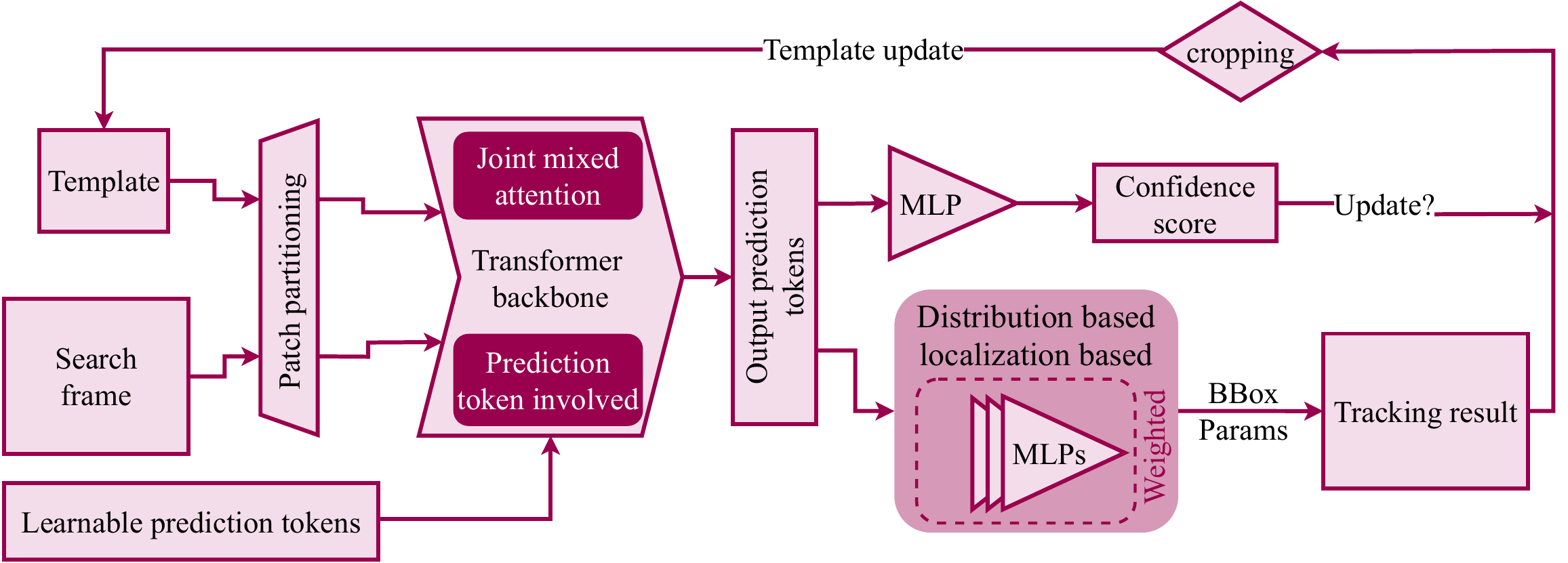}
\caption{Visual overview of effecient MixFormer2 based on learnable bounding box prediction tokens \cite{mixformerv2} .} \label{mixformerv2}
\end{figure}

\citet{mixformerv2}, tried to improve the deployment efficiency of transformer-based trackers by introducing MixFormer2. Notably, shown in Figure.~\ref{mixformerv2}, it is the first fully transformer-based tracking framework that eliminates dense convolutional heads and complex score prediction modules. It employs a set of learnable prediction tokens which are integrated with the template and search tokens through a prediction-token-involved mixed attention backbone. This unified architecture allows a significant reduction of computational overhead via direct regression of bounding box coordinates and confidence scores using lightweight MLP heads. To further improve efficiency and enable real-time performance, MixFormerV2 introduces a distillation-based model reduction strategy. This includes dense-to-sparse distillation for transferring knowledge from dense corner-head models and deep-to-shallow distillation for progressively pruning backbone layers. As a result, MixFormerV2 achieves a strong balance between tracking accuracy and speed for tracking tasks.

\begin{figure}[t!]
\centering
\includegraphics[width=1.0\textwidth]{ 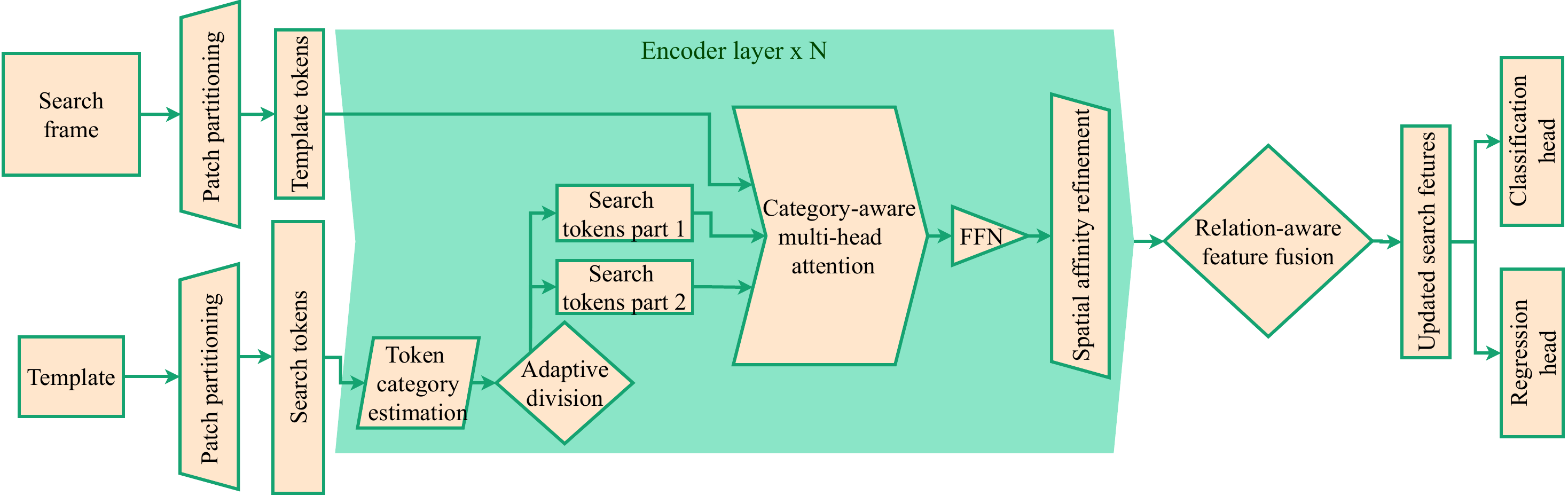}
\caption{Visual overview of GRM \cite{grm} with category-aware attention layers offering robustness against background clutter.} \label{grm}
\end{figure}

GRM \cite{grm} focused on increasing model discriminability in both one-stream and two-stream trackers by introducing a generalized relation modeling strategy that adaptively controls the token-level interaction between template and search features. Shown in Figure.~\ref{grm}, the model categorizes tokens into three groups: template tokens, interactive search tokens, and isolated search tokens. A lightweight token division module, guided by a target-aware representation and optimized via the Gumbel-Softmax trick, dynamically assigns search tokens to these groups at each encoder layer. This adaptive formulation enables the model to selectively perform cross-relation modeling only where beneficial, thus preventing confusion from background clutter and seamlessly unifying the strengths of two-stream and one-stream pipelines. To facilitate efficient computation, it uses an attention masking strategy that merges multiple attention operations into a single parallelizable step. 

\begin{figure}[t!]
\centering
\includegraphics[width=1.0\textwidth]{ 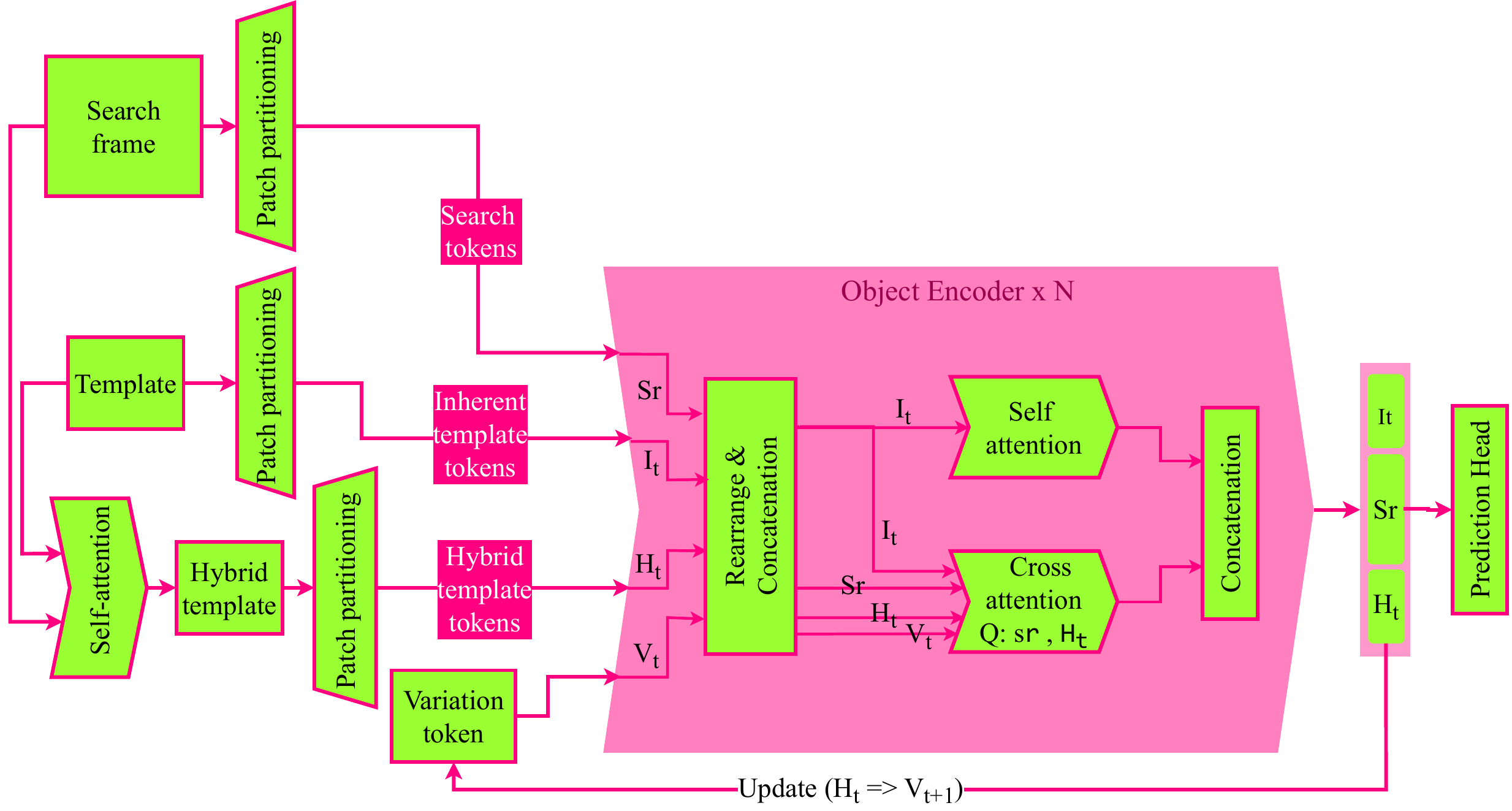}
\caption{Visual overview of three-stream ROMTrack \cite{romtrack} for temporally-aware modeling without the need for explicit online updates.} \label{romtrack}
\end{figure}

One-stream and two-stream transformer trackers have challenges because of background distraction and limited adaptability to dynamic appearance changes, respectively. To handle these issues, ROMTrack \cite{romtrack} in Figure.~\ref{romtrack} proposes a robust object modeling framework that integrates the advantages of both paradigms through a novel three-stream architecture. This tracker includes an inherent template that encodes stable and clean object features via self-attention, a hybrid template that enables dynamic fusion with the search region. In addition, there is a set of variation tokens that capture short-term temporal appearance variations across frames, which are derived from the hybrid template and injected into the attention mechanism in order to enable adaptive and temporally-aware modeling without the need for explicit online updates. ROMTrack also employs a lightweight fully convolutional center-based localization head to reduce complexity compared to corner-based regression heads. This unified design allows ROMTrack to handle appearance variations and background interference more effectively.

\begin{figure}[h!]
\centering
\includegraphics[width=1.0\textwidth]{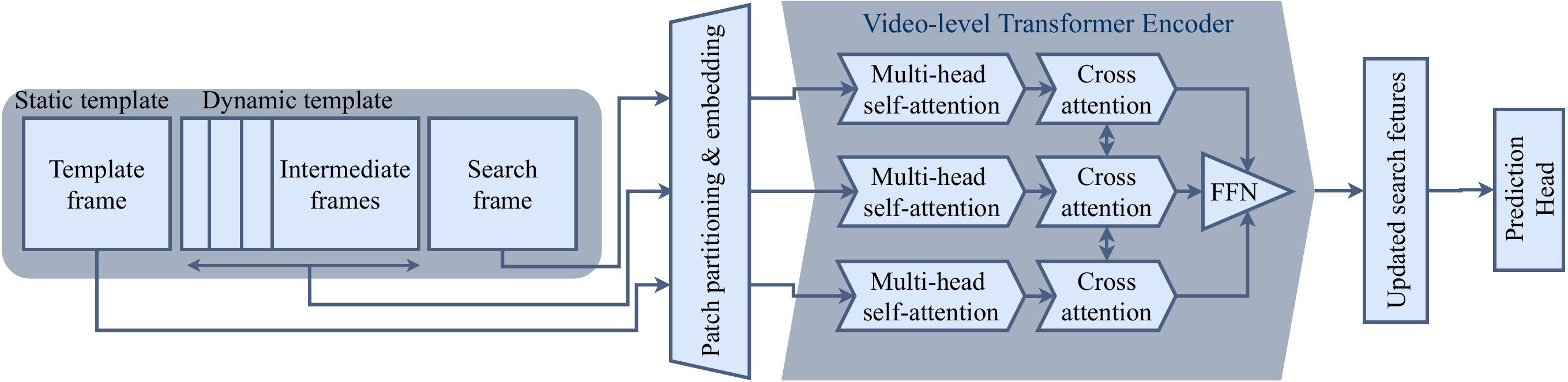}
\caption{Visual overview of VideoTrack \cite{videotrack} introducing video-level transformer for rich temporal context aggregation.} \label{videotrack}
\end{figure}

To effectively model spatiotemporal information across video sequences, VideoTrack \cite{videotrack} introduces a video-level transformer tracking framework that performs sequence-level target matching using a hierarchical triplet-block architecture. This design simultaneously attends to the initial template, a set of intermediate frames, and the current search frame, enabling rich temporal context aggregation without relying on handcrafted online updates or memory-based designs. A key innovation is the disentangled dual-template mechanism, which separates static appearance cues in the first-frame template and dynamic appearance variations captured from intermediate frames. This decomposition reduces feature redundancy and enhances temporal coherence in matching. Furthermore, to maintain compatibility with standard ViT backbones, VideoTrack leverages modified attention patterns and separated embedding strategies. A lightweight corner-based prediction head is employed for accurate localization. The resulting model performs efficient, feedforward temporal modeling without requiring complex temporal cues or motion priors. The architecture illustrates in Figure.~\ref{videotrack}.

\begin{figure}[t!]
\centering
\includegraphics[width=1.0\textwidth]{ 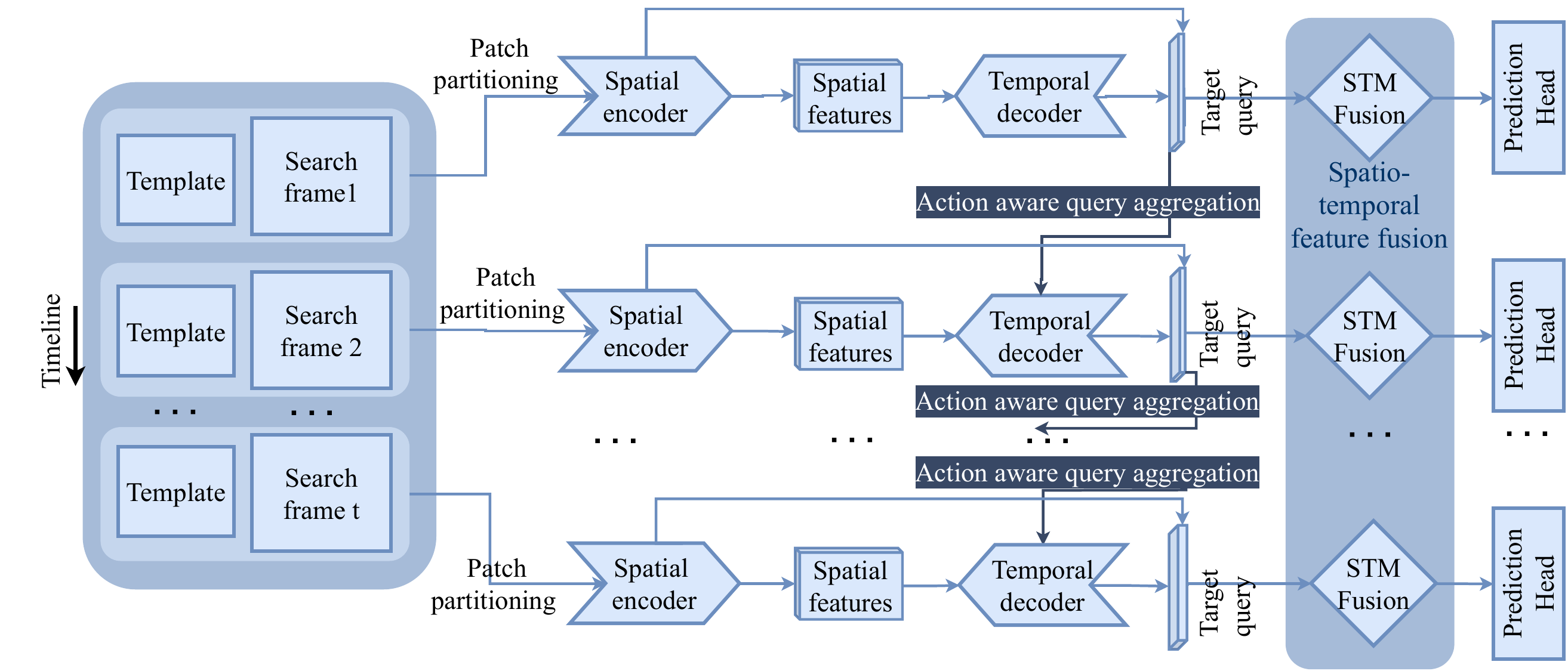}
\caption{Visual overview of AQA-Track \cite{aqa-track} highlighting its spatiotemporal sequence modeling based on temporal decoder.} \label{aqatrack}
\end{figure}

AQA-Track \cite{aqa-track} shown in Figure.~\ref{aqatrack} is another paper working on rich spatiotemporal modeling for accurate tracker against complicated target appearance variations. Instead of depending on conventional manually defined update rules or memory networks, this model introduces an adaptive transformer-based tracker that learns spatio-temporal information using autoregressive target queries. AQA-Track employs a temporal decoder that recursively refines queries over time operating in a sliding window fashion to allow the tracker to capture instantaneous appearance variations while maintaining temporal consistency. The autoregressive queries interact and accumulate spatiotemporal knowledge through a temporal attention mechanism, enabling the model to learn motion trends and appearance dynamics directly across frames. To guide localization with temporally-aware features, the model integrates a spatio-temporal fusion module (STM), which highlights spatial regions based on their temporal relevance. The backbone of AQA-Track is a lightweight hierarchical vision transformer (HiViT) \cite{hivit} that enables efficient representation learning across scales, and a center-based head is used for direct bounding box prediction. This architecture results in a strong balance between adaptability, accuracy, and computational efficiency. 

\begin{figure}[t!]
\centering
\includegraphics[width=.85\textwidth]{ 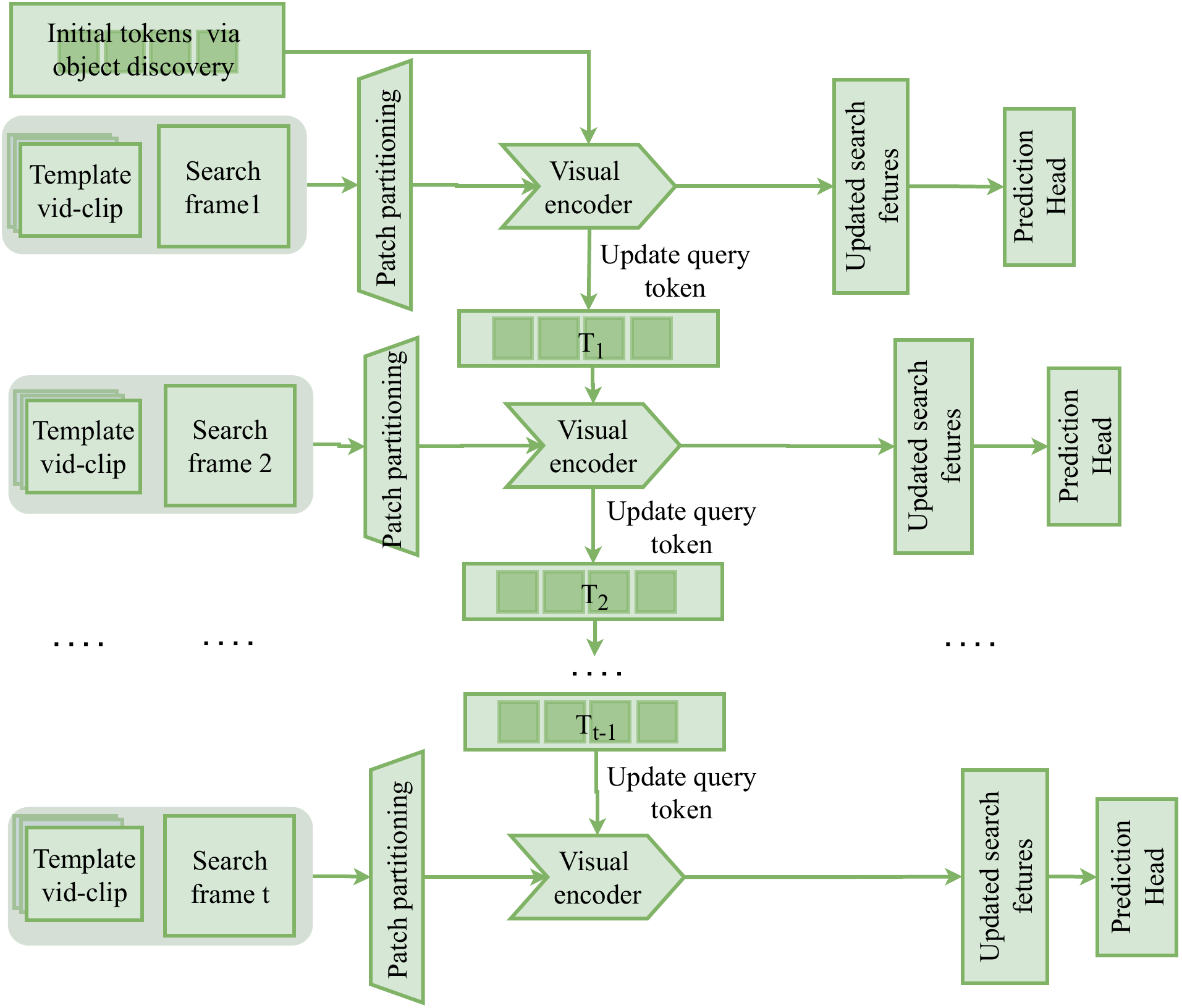}
\caption{Visual overview of ODTrack \cite{odtrack} with spatiotemporal sequence modeling using iterative token propagation strategy.} \label{odtrack}
\end{figure}

ODTrack \cite{odtrack} focuses on addressing the limitations of sparse temporal modeling in visual tracking by introducing a simple yet effective video-level tracking framework that performs online dense contextual association via iterative token propagation. Instead of relying on traditional image-pair matching or handcrafted online updates, ODTrack reformulates tracking as a sequence-level task that compresses target appearance and localization cues into compact temporal tokens. These tokens serve as dynamic prompts which are propagated frame-by-frame to enable spatiotemporal trajectory modeling across arbitrarily long video clips. A key component in this architecture is the temporal token propagation attention mechanism, which facilitates efficient online reasoning without requiring specialized optimization procedures or complex update modules. In addition, to accommodate long-term motion variation, ODTrack employs a video sequence sampling strategy that extracts sparse but informative frame sets. The architecture on this paper is illustrated in Figure.~\ref{odtrack}.

\begin{figure}[h!]
\centering
\includegraphics[width=1\textwidth]{ 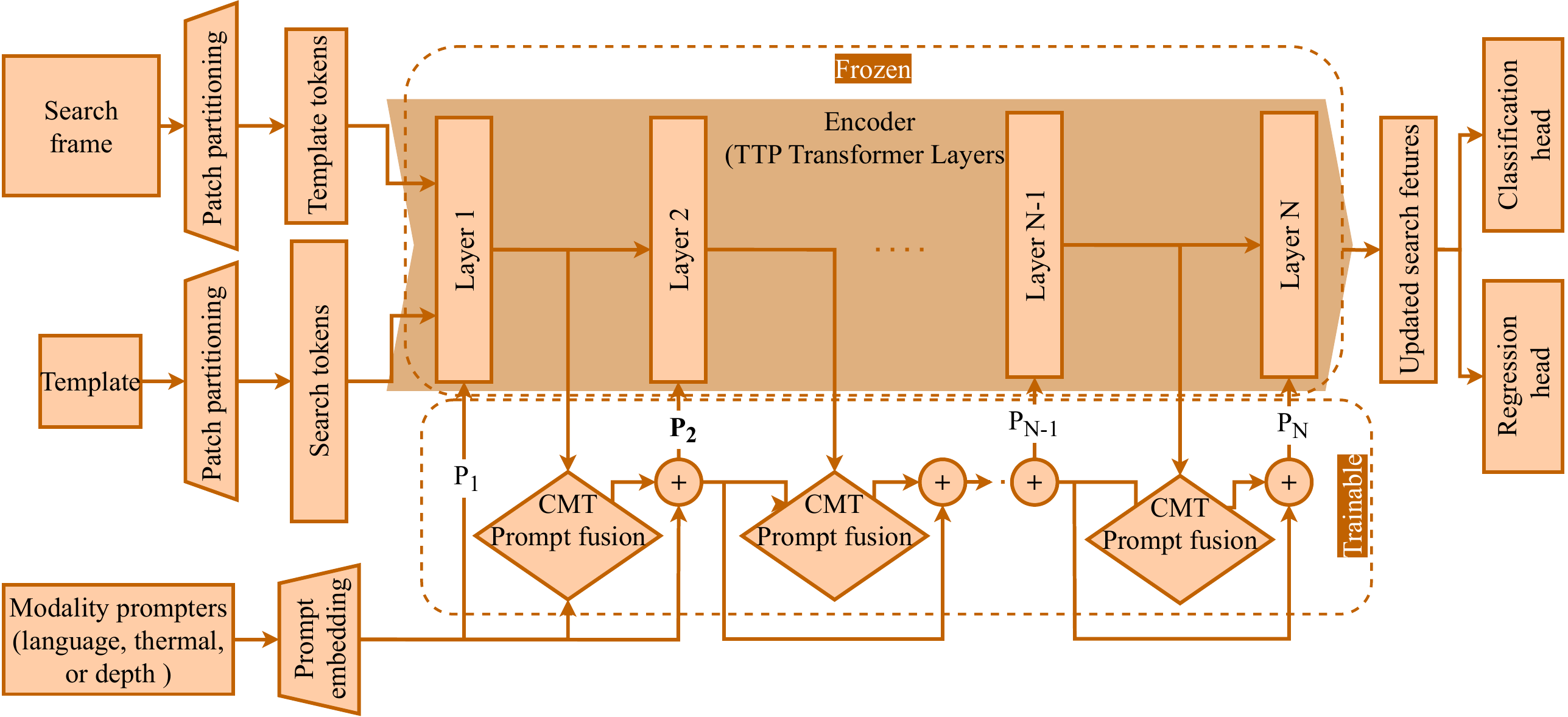}
\caption{Visual overview of OneTracker \cite{onetracker} with prompt-based modeling for improved generalization across modalities.} \label{onetracker}
\end{figure}

Concventional GOT methods often relied on modality-specific designs by using customized architectures with redundant parameters and limited performance. OneTracker \cite{onetracker} addresses this limitation by introducing a unified and efficient framework for both RGB and multimodal (RGB+X) tracking using a modular two-stage design. At its core lies the Foundation Tracker which is a transformer-based model pretrained on large-scale RGB tracking datasets to develop generalizable temporal matching capabilities. Shown in Figure.~\ref{onetracker}, to extend the model to other modalities, OneTracker integrates a Prompt Tracker module that treats extra inputs as task prompts. These extra inputs can be depth, thermal, segmentation masks, or language. This is achieved through the introduction of Cross-Modality Tracking Prompters (CMT-Prompters) and Tracking Task Perception (TTP) Transformer layers, which allow parameter-efficient fine-tuning by updating only lightweight adapters while keeping the main foundation model frozen. This design supports prompt-based multimodal fusion and enables task-specific adaptability without modifying the core model structure, making OneTracker an effective and extensible solution for diverse tracking scenarios across multiple input modalities.

\begin{figure}[t!]
\centering
\includegraphics[width=1\textwidth]{ 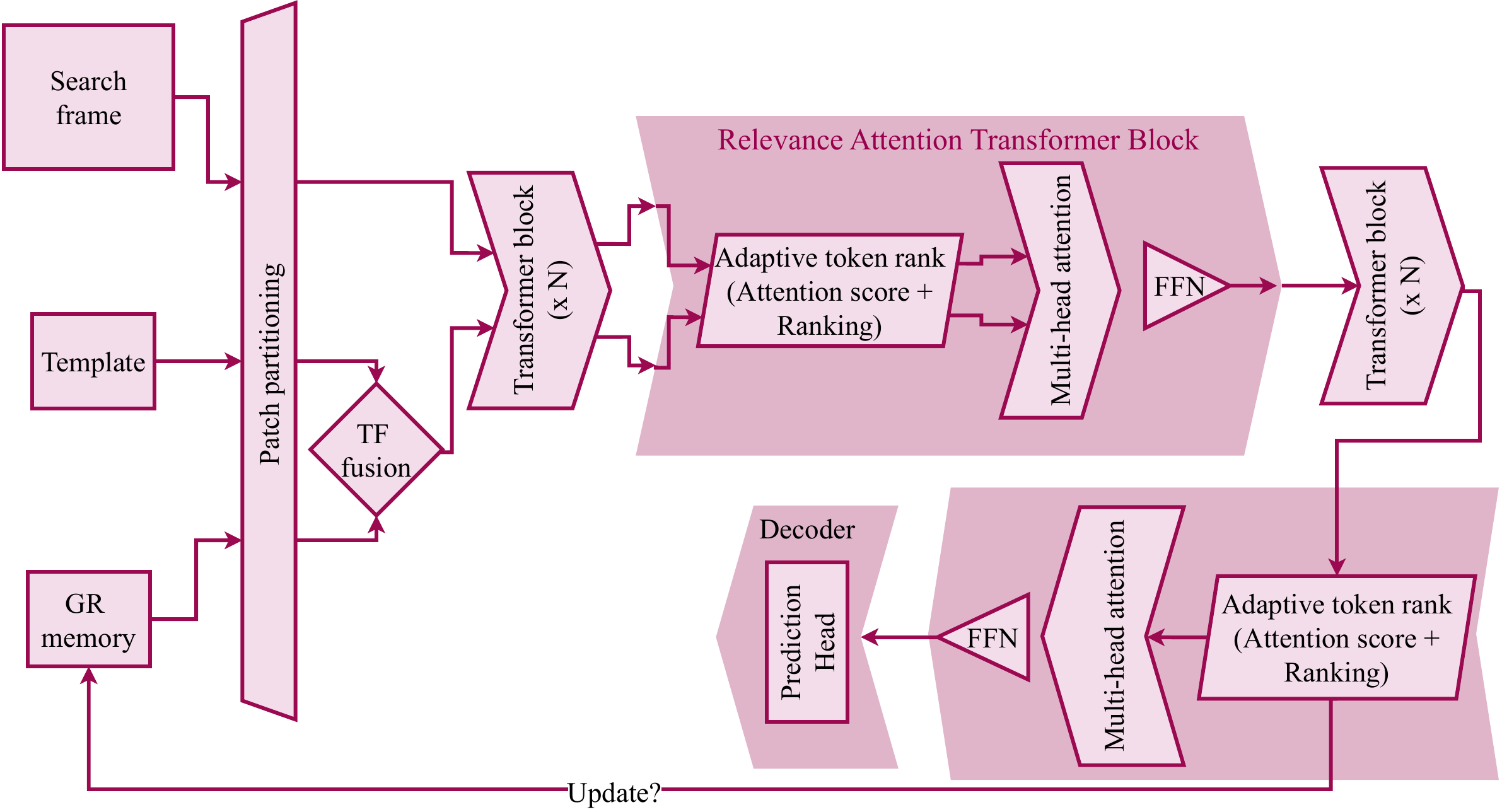}
\caption{Visual overview of RFGM \cite{rfgm} highlighting its memory adaptation technique for long-term tracking.} \label{rfgm}
\end{figure}

Most transformer-based trackers suffer from the accumulation of redundant or irrelevant information when integrating features from historical frames, especially in long-term tracking scenarios with significant appearance variation. \citet{rfgm} addresses this limitation by introducing RFGM in Figure.~\ref{rfgm} (Reading Relevant Feature from Global Representation Memory) with a global memory-based tracking paradigm that dynamically retrieves only the most relevant features for each frame. The core design of RFGM is the Global Representation (GR) memory, which stores feature tokens from previous templates, and a novel Relevance Attention mechanism that adaptively ranks and filters these tokens based on their similarity to the current search frame. Unlike conventional methods that apply cross-attention uniformly across all tokens, this approach learns to adaptively rank and filter memory tokens based on their relevance to the current search frame, thus preserving critical target features while discarding distractors. Additionally, a token filter module is used to selectively update the GR memory at the token level, ensuring memory compactness and relevance over time. To maintain computational efficiency, relevance attention is only applied at specific transformer layers. This design improves long-term tracking robustness while avoiding the cost of full memory attention at every stage.

\begin{figure}[t!]
\centering
\includegraphics[width=1\textwidth]{ 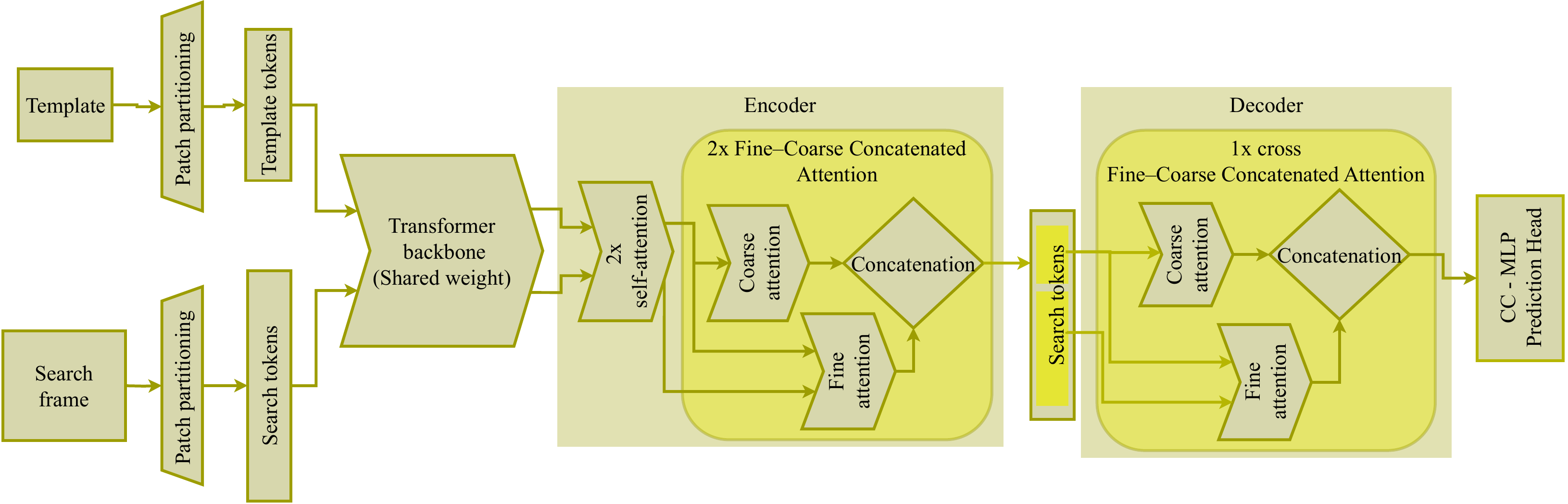}
\caption{Visual overview of FCAT \cite{fcat} composed of Fine–Coarse Concatenated Attention (FCA) and Cross-Concatenation MLP (CC-MLP).} \label{fcat}
\end{figure}

FCAT \cite{fcat}, shown in Figure.~\ref{fcat} (Fully Concatenated Attentional Tracker) focuses on handling multi-scale variations and local interactions to improve the accuracy in transformer-based trackers. This model introduces a fully attentional tracking framework composed of two key modules: Fine–Coarse Concatenated Attention (FCA) and Cross-Concatenation MLP (CC-MLP). The FCA module learns both fine-grained and coarse-grained feature representations simultaneously by applying multi-scale convolution before the attention operation in order to enable robust tracking under scale variations and occlusion. The CC-MLP further enriches feature representation by embedding depth-wise convolutions within the feed-forward layers, enabling more effective modeling of local token interactions. Together, these modules form an encoder-decoder transformer that unifies the template and search regions, followed by a dual-branch prediction head that performs classification and bounding box regression. FCAT thus achieves strong spatial sensitivity while maintaining the flexibility of a transformer-based framework. 

\begin{figure}[h!]
\centering
\includegraphics[width=1\textwidth]{ 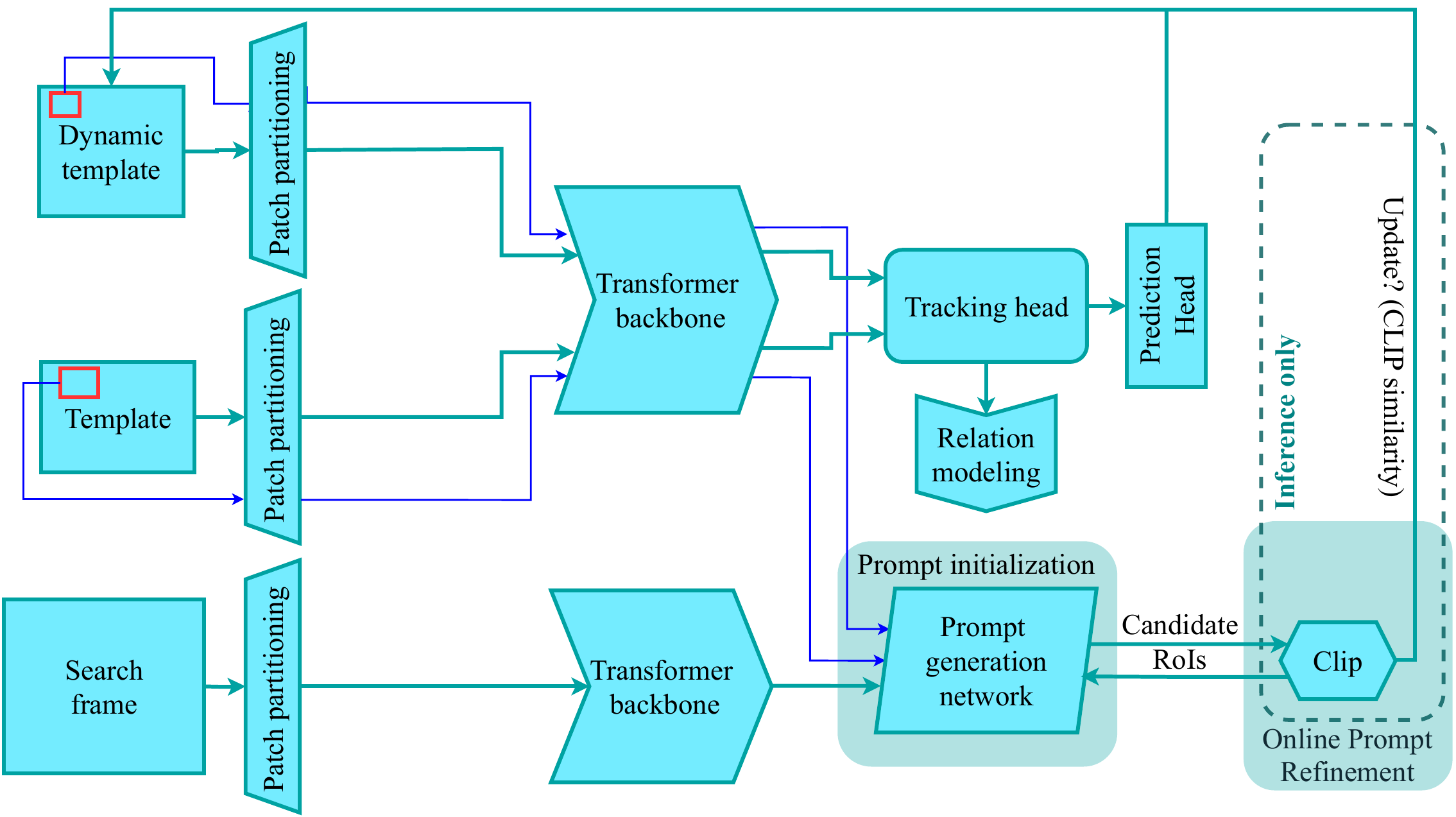}
\caption{Visual overview of promptable PiVOT \cite{pivot} tracker integrating CLIP \cite{clip} into visual tracking.} \label{pivot}
\end{figure}

Shown in Figure.~\ref{pivot} In order to establish a more discriminative tracker, PiVOT \cite{pivot} proposes a promptable tracking framework that integrates the strong visual-semantic priors of the CLIP \cite{clip} foundation model into visual tracking via learnable visual prompting. The architecture consists of a Prompt Generation Network (PGN) that generates score maps highlighting potential target regions and a Relation Modeling (RM) module that fuses these prompts with frame-level features to guide target localization. During inference, PiVOT employs a Test-time Prompt Refinement (TPR) strategy that leverages CLIP’s zero-shot visual capability to refine candidate object regions based on their similarity to reference templates. This mechanism enables the tracker to dynamically suppress distractors and focus on the correct target, even under severe occlusion, appearance variation, or semantic ambiguity. Unlike prior works that fine-tune large transformer backbones, PiVOT freezes the ViT-L backbone and uses a lightweight adapter module for efficient training and inference, drastically reducing training complexity while preserving generalization.

\begin{sidewaystable}
\tiny
\centering
\caption{Detailed Comparison Of Fully Transformer-Based Trackers with Pure Attention Architectures}
\label{fullytab}
\begin{tabular}{%
    >{\centering\arraybackslash}p{0.5cm}|  
    p{1.4cm}  
    p{0.5cm}  
    p{1.3cm}  
    p{2.5cm}  
    p{2.5cm}  
    p{2.5cm}  
    p{2.5cm}  
    p{1cm}    
    p{2cm}    
}

\midrule
\multirow{22}{*}{\rotatebox{90}{\parbox[c]{14cm}{\centering \textbf{Transformer-based Appearance Model} \\ \textbf{ Pure-Attention Architectures}}}} & 
\textbf{Method} & \textbf{Year} & \textbf{Backbone Network} & \textbf{Design Highlight} & \textbf{Focus} & \textbf{Novelty} & \textbf{Drawbacks} & \textbf{Template Update} & \textbf{Architectural-Level of Contribution} \\
\cmidrule{2-10}
& SwinTrack \cite{swintrack} & 2021 & Swin-Tiny/ Swin-Base \cite{swintransformer}  & Fully attentional tracker using Swin Transformer; Concatenation-based feature fusion; Motion token encodes historical target trajectory in decoder; No query-based decoder & Joint transformer-based feature representation; Motion-aware target localization & First to unify representation learning and fusion via Swin Transformer; Motion token to efficiently enhance robustness & No memory or explicit adaptation; Performance may degrade under drastic appearance shifts & Yes & Feature representation \\
\cmidrule{2-10}
& SimTrack \cite{simtrack} & 2022 & ViT \cite{vitbackbone} & One-branch transformer backbone with serialized input; Joint feature learning and interaction; Foveal window strategy enhances patch diversity and central target focus & Simplified and generalizable transformer framework with target-sensitive representation & First to unify feature learning and interaction within backbone; Introduces foveal patch sampling for improved target detail & No update strategy; Rely heavily on spatial cues from static images and fail to capture temporal correspondences, Miss fore-background relationship for the template and search region & No & \begin{tabular}[c]{@{}l@{}}Feature representation\\ Sampling strategy\end{tabular} \\
\cmidrule{2-10}
& OSTrack \cite{ostrack} & 2022 & ViT \cite{vitbackbone} & One-stream one-stage framework ; Joint feature learning and relation modeling; Early candidate elimination discards background tokens via attention-derived similarity score & Increase discriminability and target awareness; Unified target-aware representation and efficient relation modeling & Combines feature extraction and relation modeling via only self-attention; Early candidate elimination & No template update mechanism; Rely heavily on spatial cues from static images and fail to capture temporal correspondences; Miss fore-background relationship for the template and search region & No & \begin{tabular}[c]{@{}l@{}}Feature representation\\ Relation modeling\end{tabular} \\
\cmidrule{2-10}
& SBT \cite{sbt} & 2022 & Proposed attention-based backbone & Correlation-aware target-dependent feature extraction via Extract-or-Correlation (EoC) blocks; Novel self-/cross-attention single-stream backbone; Prediction on processed search features without separate correlation step & More discriminative instance-specific representation learning and distractor suppression & Single-branch attention based architecture unifying feature extraction and relation modeling; Extract-or-Correlation (EoC) blocks; Faster convergence & No explicit online update & No & Feature representation; Relation modeling \\
\cmidrule{2-10}
& DropMAE \cite{dropmae} & 2023 & ViT\cite{vitbackbone} & first to investigate masked autoencoder video pre-training for tracking, Introduce adaptive spatial-attention dropout to enforce temporal correspondence learning & Effective temporal correspondence learning in videos & First temporal-matching MAE video pretraining; Apply adaptive spatial-attention dropout to improve temporal cues by suppressing spatial co-adaptation & Sensitivity to dropout ratio; Absence of online adaptation mechanisms & No & Relation modeling \\
\cmidrule{2-10}
& F-BDMTrack \cite{f-bdmtrack} & 2023 & Swin-Tiny/Swin-Base \cite{swintransformer} & Includes fore-background agent learning module and a distribution-aware attention module in a unified transformer for distribution-aware feature aggregation; Uses pseudo-bounding box strategy to model foreground-background agents & Robust feature discrimination in cluttered scenes and dynamic appearance changes to discriminate foreground-background & ; Enables precise target localization under complex distractors & Computational overhead from dual-agent modeling and pseudo box generation; No online template update & No & Feature representation; Relation modeling \\
\midrule
\end{tabular}
\end{sidewaystable}

\begin{sidewaystable}
\ContinuedFloat
\tiny
\centering
\caption{Detailed Comparison Of Fully Transformer-Based Trackers with Pure Attention Architectures}
\begin{tabular}{%
    >{\centering\arraybackslash}p{0.5cm}|  
    p{1.4cm}  
    p{0.5cm}  
    p{1.3cm}  
    p{2.5cm}  
    p{2.5cm}  
    p{2.5cm}  
    p{2.5cm}  
    p{1cm}    
    p{2cm}    
}
\midrule
\multirow{22}{*}{\rotatebox{90}{\parbox[c]{14cm}{\centering \textbf{Transformer-based Appearance Model} \\ \textbf{ Pure-Attention Architectures}}}} & 
\textbf{Method} & \textbf{Year} & \textbf{Backbone Network} & \textbf{Design Highlight} & \textbf{Focus} & \textbf{Novelty} & \textbf{Drawbacks} & \textbf{Template Update} & \textbf{Architectural-Level of Contribution} \\
\cmidrule{2-10}
& ARTrack \cite{artrack} & 2023 & ViT \cite{vitbackbone} & based on general encoder-decoder architecture, Autoregressive framework with sequence decoding; Uses previous estimates as spatio-temporal prompts for Trajectory modeling; No post-processing or specialized heads & Sequence-level modeling of motion dynamics for for coherence temporal modeling; Exclude complicated postprocessing & First transformer tracker to unify tracking with autoregressive language modeling; Simultaneous visual template matching and motion information modeling. & Lack of online template updating; Limited efficiency due to sequential decoding & No & Relation modeling \\
\cmidrule{2-10}
& MixFormer2 \cite{mixformerv2} & 2023 & ViT \cite{vitbackbone} & Fully transformer framework with prediction-token-involved mixed attention; direct box and score regression via MLP heads; Distillation-based model reduction for lightweight variants & Efficient tracking algorithm with efficient token-based regression and score estimation & First convolution-free transformer tracker; Uses dense-to-sparse and deep-to-shallow distillation; Prediction-token-involved mixed attention in backbone & Requires multi-stage distillation for compact model training & Yes & Feature representation, Target state estimation \\
\cmidrule{2-10}
& SeqTrack \cite{seqtrack} & 2023 & ViT \cite{vitbackbone} & Models tracking as a sequence-to-sequence task using an encoder-decoder transformer; Discretizes bounding boxes into tokens; Decodes bounding autoregressively & End-to-end token-based bounding box prediction; Getting rid of complicated head networks and loss functions & First fully sequence-to-sequence tracker; Eliminates classification/regression heads using token-level generation; Integrates online template update via likelihood-based selection & Lacks explicit re-detection module; Sensitive to long-term occlusion & Yes & Target state estimation \\
\cmidrule{2-10}
& GRM \cite{grm} & 2023 & ViT \cite{vitbackbone} & Generalized attention-based relation modeling with adaptive token division; Dynamically controls token-level attention interaction between template and search via Gumbel-Softmax; Apply attention masking strategy and the Gumbel-Softmax in token division module & Extract more discriminative feature representation with background suppression & First unified formulation of token-wise relation modeling; Learnable token division and attention masking for more flexible and efficient relation modeling; Unifies the strengths of two-stream and one-stream pipelines & Increased model complexity; Lack of online adaptation & No & Feature representation; Relation modeling \\
\cmidrule{2-10}
& ROMTrack \cite{romtrack} & 2023 & ViT\cite{vitbackbone} & Robust object modeling with inherent and hybrid template streams and variation tokens; integrates appearance adaptation via token-level attention fusion, Center-based regression & Discriminative and adaptive target modeling for robust tracking & First tracker to unify inherent and hybrid template modeling; Apply variation tokens for temporal robustness & Increased Complexity in Token Design & No & Feature representation; Relation modeling \\
\cmidrule{2-10}
& VideoTrack \cite{videotrack} & 2023 & ViT\cite{vitbackbone} & Video transformer architecture; Enable spatiotemporal feature learning via sequential multi-branch triplet blocks; Dual-template design for separate static and dynamic appearance clues; No need for model update & Efficient spatiotemporal modeling without online update or memory & Dual-template video transformer tracker; Novel triplet attention blocks for temporal context encoding & Fixed reference set; Computational cost of triplet attention & No & Feature representation; Relation modeling \\
\midrule
\end{tabular}
\end{sidewaystable}

\begin{sidewaystable}
\ContinuedFloat
\tiny
\centering
\caption{Detailed Comparison Of Fully Transformer-Based Trackers with Pure Attention Architectures}
\begin{tabular}{%
    >{\centering\arraybackslash}p{0.5cm}|  
    p{1.4cm}  
    p{0.5cm}  
    p{1.3cm}  
    p{2.5cm}  
    p{2.5cm}  
    p{2.5cm}  
    p{2.5cm}  
    p{1cm}    
    p{2cm}    
}
\midrule
\multirow{22}{*}{\rotatebox{90}{\parbox[c]{14cm}{\centering \textbf{Transformer-based Appearance Model} \\ \textbf{ Pure-Attention Architectures}}}} & 
\textbf{Method} & \textbf{Year} & \textbf{Backbone Network} & \textbf{Design Highlight} & \textbf{Focus} & \textbf{Novelty} & \textbf{Drawbacks} & \textbf{Template Update} & \textbf{Architectural-Level of Contribution} \\
\cmidrule{2-10}
& AQA-Track \cite{aqa-track} & 2024 & HiViT\cite{hivit} & Autoregressive target queries refined via temporal decoder; temporal attention for motion modeling; STM module for spatial-temporal fusion & Adaptive spatio-temporal modeling without handcrafted updates or memory & First to use autoregressive queries for spatio-temporal learning; Sliding window formulation enables motion trend estimation & Limited long-term modeling due to query window length; Scalability restricted by memory constraints & No & Target state estimation \\
\cmidrule{2-10}
& ODTrack \cite{odtrack} & 2024 & ViT\cite{vitbackbone} & Video-level transformer framework; Dense online temporal token propagation; Compresses appearance and trajectory information into token sequences, Propagates token sequences autoregressively & Long-range temporal modeling through token-based context propagation without complex online updates & Reformulates tracking as token propagation across frames; Introduces two lightweight temporal attention mechanisms; Avoids handcrafted online update strategies & Slow inference due to computational complexity, Accumulated Error in Autoregressive Token Propagation & No & Relation modeling\\
\cmidrule{2-10} 
& OneTracker \cite{onetracker} & 2024 & ViT\cite{vitbackbone} & Unified RGB and RGB+X tracking via Foundation and Prompt Trackers;  Cross modality tracking prompters and Transformer layers for efficient multimodal adaptation & Generalizable prompt-based framework for RGB and multimodal tracking tasks & First general framework to unify various tracking tasks via prompt-tuning framework for tracking; Integrates language, mask, depth, and thermal modalities via tokenized prompts & Prompt complexity and optimization sensitivity & Yes & Appearance model \\
\cmidrule{2-10}
& RFGM \cite{rfgm} & 2024 & ViT \cite{vitbackbone} & Construct a
global representation (GR) memory; Dynamically selects relevant reference features from token-level GR memory; Using relevance attention to choose tokens based on dynamic ranking; Memory updated via adaptive token ranking and filtering & Fine-grained memory integration and temporal adaptability and robustness in long-term tracking & First to apply adaptive token-ranking-based memory reading and updating; Relevance attention enhances feature selection across frames & Potential error accumulation from noisy updates; Memory growth and token management complexity & Yes & Online update; Relation modeling \\
\cmidrule{2-10}
& FCAT \cite{fcat} & 2024 & Swin-T/ Swin-B \cite{swintransformer} & Fully attentional encoder-decoder with Fine–Coarse Concatenated Attention (FCA) and Cross-Concatenation MLP (CC-MLP); Captures multi-scale and local interactions simultaneously & Scale-robust and spatially discriminative representation learning & Fine-coarse dual granularity attention and local interaction modeling; Improves both self-attention and MLP modules for tracking-specific challenges & Lacks template update mechanism; Performance may degrade under long-term appearance drift & No & Feature representation; Relation modeling \\
\cmidrule{2-10}
& PiVOT \cite{pivot} & 2025 & ViT \cite{vitbackbone} & Prompt generation network and relation modeling module for prompt-based feature fusion; CLIP-guided online refinement enhances target-awareness without backbone tuning & More discriminative model and  distractor suppression via promptable tracking using zero-shot knowledge transfer & First tracker to integrate CLIP-based test-time visual prompting with frozen foundation model backbone; Supports dynamic prompt refinement via token similarity & Inference involves dual backbones increasing runtime cost, Reliance on CLIP’s\cite{clip} generalization capabilities & Yes & Appearance model, relation modeling \\
\midrule
\end{tabular}
\end{sidewaystable}

\section{Experimental Comparison}\label{exp}

In this section, an experimental comparison based on widely accepted benchmarks and evaluation protocols is presented in order to provide a comprehensive and objective understanding of the performance characteristics of the reviewed methods. The aim is to highlight the practical strengths and limitations of each tracking paradigm in real-world scenarios by systematically analyzing results across standard datasets and performance metrics. This prepares a fair assessment of accuracy, robustness, and computational efficiency. The following subsections detail the benchmark datasets used, the evaluation metrics adopted, and the performance outcomes reported by recent studies.

\subsection{Tracking Datasets}\label{dataset}



GOT datasets are designed to evaluate algorithms under diverse and realistic conditions. Below, we categorize these datasets by their temporal scope (short-term vs. long-term) and highlight their unique attributes, challenges, and contributions to advancing tracking research.

\subsubsection{Short-Term Tracking Datasets}
Short-term benchmarks focus on continuous tracking in sequences where targets remain visible or experience brief occlusions. Early benchmarks like OTB2013 \cite{otb2013} and its successor OTB2015 \cite{otb2015} laid the foundation for fair comparisons in VOT. OTB2013 introduced 50 video sequences annotated with attributes such as illumination variation and occlusion, while OTB2015 expanded this to 100 sequences, addressing biases in initial conditions and adding challenges like fast motion. These datasets became cornerstones for evaluating robustness but were limited in scale, prompting the creation of more diverse benchmarks.

The Temple-Color 128 (TC128) dataset \cite{TC128} emerged to address color sensitivity in tracking, offering 129 sequences, 78 of which were distinct from OTB, to study how trackers perform under color variations and aspect ratio changes. Meanwhile, the Amsterdam Library of Ordinary Videos (ALOV) \cite{ALOV} compiled 314 YouTube-sourced videos with 13 difficulty levels, emphasizing real-world challenges like viewpoint changes. However, ALOV’s per-sequence single-attribute annotations limited its utility for studying overlapping challenges.

The VOT challenges revolutionized evaluation protocols by introducing per-frame rotatable bounding boxes and the TraX protocol, which automated failure detection and tracker reinitialization. VOT’s yearly iterations refined these protocols, but its small size (60-360 sequences) restricted its use for training deep models. This gap was filled by TrackingNet \cite{trackingnet}, a large-scale dataset with 500 YouTube videos and over 14 million bounding boxes, enabling end-to-end training of data-hungry deep trackers.

For occlusion analysis, the NUS People and Rigid Objects (NUS-PRO) dataset \cite{nuspro} provided 365 sequences with frame-level occlusion labels (none/partial/full), making it invaluable for pedestrian tracking studies. The Need for Speed (NfS) dataset \cite{NFS} introduced high-frame-rate (240 FPS) videos to explore real-time tracking under fast motion and motion blur, while GOT-10k \cite{got10k} broke new ground with 10,000+ videos spanning 563 object classes and labels to evaluate robustness to temporary target disappearances. Additionally, the TracKlinic \cite{tracklinic} isolated specific challenges (e.g., occlusion with rotation) per sequence, offering a toolkit for targeted performance analysis.

Most short-term datasets prioritize common challenges (e.g., occlusion, scale variation) but lack annotations for compound attributes (e.g., occlusion during fast motion). Additionally, few include segmentation masks, limiting studies on precise target localization.

\subsubsection{Long-Term Tracking Benchmarks}
Long-term tracking demands resilience to frequent target disappearances and reappearances, mimicking real-world surveillance or wildlife monitoring. The OxUvA dataset \cite{oxuva}, derived from 14 hours of YouTube-BoundingBoxes videos, pioneered absent labels to assess re-detection capabilities. However, its sparse annotations limited fine-grained analysis. The TLP dataset \cite{tlp} improved temporal consistency studies with high-resolution, long-duration videos but lacked frequent target disappearances.

This shortcoming was addressed by LTB-35 \cite{LTB35}, which averaged 12 target disappearances per sequence, stressing tracker recovery. The Large-Scale Single Object Tracking (LaSOT) benchmark \cite{lasot} set a new standard with 1,400 sequences (2.3 million frames) and balanced object categories from ImageNet. LaSOT’s dense annotations and class balance reduced evaluation bias, though its focus on single-target scenarios overlooked multi-object challenges. Long-term datasets often neglect temporal consistency (e.g., gradual appearance changes over hours) and rarely include multi-target scenarios, limiting their utility for real-world applications like crowd monitoring. (see Table~\ref{tab:tracking-benchmarks} for a structured comparison).

\begin{table}[ht]
\centering
\caption{Overview of widely used visual tracking datasets. The table summarizes dataset scale, diversity, and characteristics relevant for training and evaluation.}
\label{tab:tracking-benchmarks}
\begin{tabular}{p{2.8cm}p{1.1cm}p{1.2cm}p{1.1cm}p{1cm}lp{1cm}p{0.7cm}}
\toprule
Dataset & \# Seqs & Total Frames & Avg. Length & Object Classes & Frame Resolution & Attr. Count & Track Type \\
\midrule
OTB-2015 \cite{otb2015}             & 100     & 59,000     & 598     & 16   & –              & 11  & Short \\
VOT2015 \cite{vot2015}            & 60      & 21,455     & 357     & 20   & –              & 11  & Short \\
VOT2016 \cite{vot2016}             & 60      & 21,455     & 357     & 20   & –              & 5   & Short \\
VOT2018 \cite{vot2018}            & 60      & 21,356     & 356     & 24   & –              & 5   & Short \\
TLP \cite{tlp}              & 50      & 676,000    & 13,000  & 17   & 1280$\times$720 & 6   & Long \\
UAV123 \cite{uav123}             & 123     & 113,000    & 915     & 9    & –              & 12  & Short \\
ALOV300++ \cite{ALOV}         & 315     & 8,936      & 483     & –    & –              & 14  & Short \\
TC-128 \cite{TC128}             & 129     & 55,000     & 431     & 27   & –              & 11  & Short \\
OXUVa \cite{oxuva}           & 366     & 1.55M      & 4,200   & 22   & –              & 6   & Long \\
LTB35 \cite{LTB35}           & 35      & 146,000    & 4,000   & 19   & 1280$\times$720 $\sim$ 290$\times$217 & 10  & Long \\
GOT-10k \cite{got10k}          & 10,000  & 1.5M       & 149     & 563  & –              & 6   & Short \\
LaSOT \cite{lasot}           & 1,400   & 3.52M      & 2,506   & 70   & 1280$\times$720 & 14  & Long \\
TrackingNet \cite{trackingnet}      & 30,000  & 14M        & 471     & 27   & –              & 15  & Short \\
NUS-PRO \cite{nuspro}      & 365     & 109,000    & 370     & 8    & 1280$\times$720 & 12  & Short \\
\bottomrule
\end{tabular}

\end{table}



\subsection{Evaluation Metric}\label{metric}

There are several standard evaluation metrics widely adopted in the literature in order to provide a consistent and objective performance assessment across tracking methods. These metrics focus on critical aspects of tracking performance, such as target localization accuracy, robustness to tracking failures, and adaptability to various conditions. Precision-based metrics are important for evaluating spatial accuracy. They quantify the proportion of frames in which the predicted target center falls within a predetermined threshold of the ground-truth center, making spatial accuracy sensitive to image resolution and object scale. To overcome this limitation, normalized precision adjusts the threshold based on its relation to the target size to enable scale-invariant evaluation. Additionally, Center Location Error (CLE) reports the average Euclidean distance between predicted and ground-truth centers, providing a raw but informative measure of tracking accuracy.

IoU-based metrics provide a more region-aware assessment. For instance, the success rate indicates the percentage of frames where the Intersection over Union (IoU) between predicted and ground-truth bounding boxes exceeds a given threshold. Over varying IoU thresholds, the Area Under the Curve (AUC) is computed, which is often used in OTB and LaSOT benchmarks to summarize overall tracking performance. In addition, the Expected Average Overlap (EAO), which is primarily used in the VOT challenge, combines accuracy and robustness into a single measure by estimating the expected IoU over a sequence while penalizing tracking failures.

These evaluation metrics are typically chosen based on the benchmark dataset and the goals of the evaluation. For instance, the OTB dataset mainly reports precision and CLE, and the LaSOT benchmark emphasizes normalized precision and AUC. This is while the VOT dataset adopts EAO for short-term tracking evaluation. The comparative analysis across different trackers remains fair, interpretable, and reproducible via employing these standardized metrics.

\subsection{Performance Evaluation}\label{perf}

\begin{figure}[h!]
\centering
\includegraphics[width=1\textwidth]{ 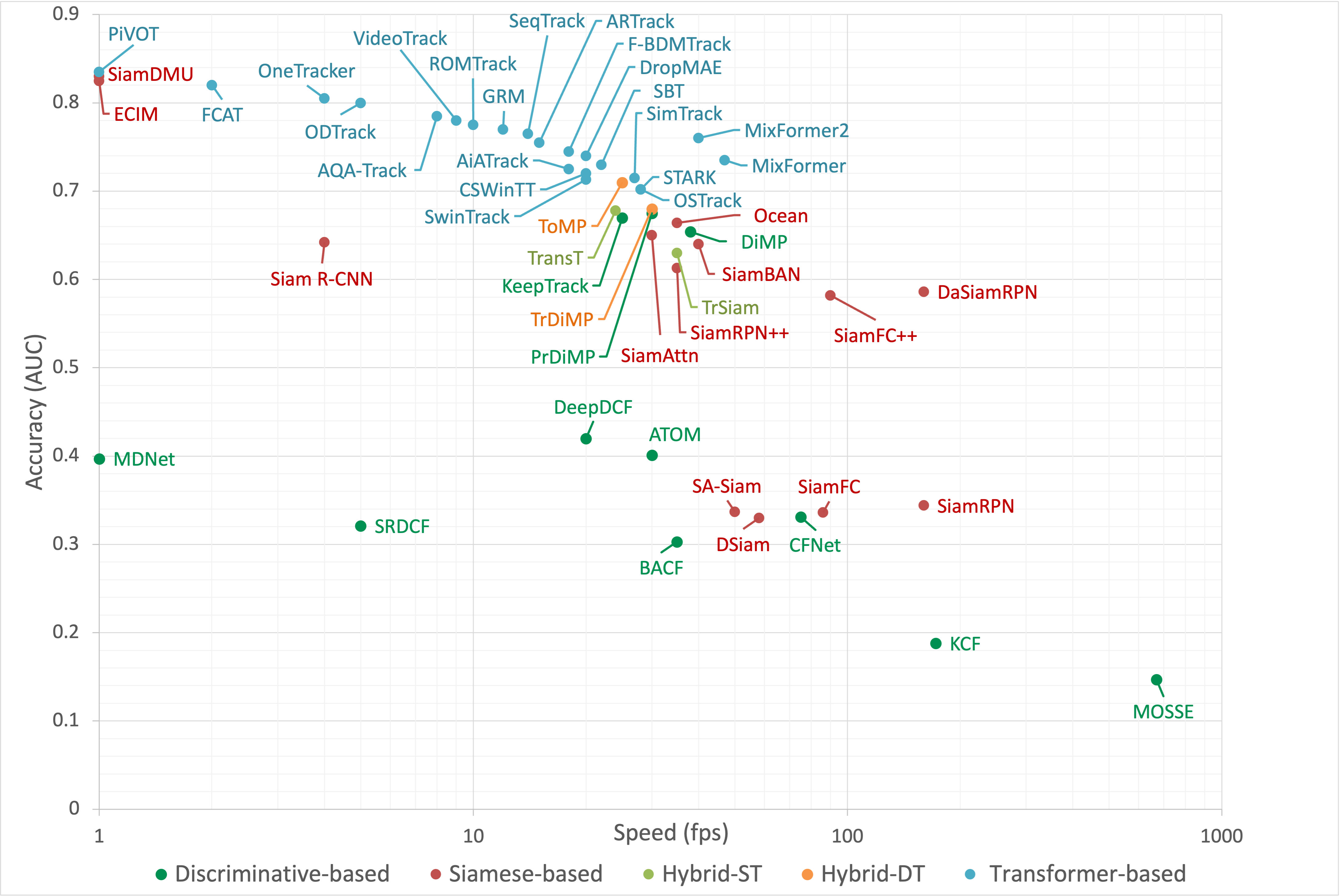}
\caption{Performance comparison of reviewed GOT trackers categorized by appearance model based on AUC vs. FPS on LaSOT \cite{lasot} dataset. Trackers are color-coded by their appearance model type (Discriminative-based, Siamese-based, Transformer-based, and Hybrid variants). The upper-left region, near the center of the plot is constantly occupied by fully transformer-based trackers (blue dots), which demonstrates their high accuracy at moderately fast speeds.}
\label{fig:auc_fps}
\end{figure}

Figure.~\ref{fig:auc_fps} presents a comparative analysis of GOT trackers, grouped by their underlying appearance model, in terms of AUC and runtime speed (FPS, log scale) on LaSOT datatset \cite{lasot}. Discriminative-based trackers shown in green exhibit moderate to low accuracy with relatively slow speeds because of the computational cost of their online learning mechanisms. Siamese-based trackers, depicted in dark red, perform noticeably faster during inference but with lower AUC values due to their poor discriminative quality. The hybrid models (orange and light green), which combine transformer modules with either Siamese (ST) or discriminative (DT) backbones, are placed in the mid-range of both accuracy and speed. This demonstrates how effectively they strike a balance between temporal modeling and efficiency. The upper-left region, near the center of the plot is constantly occupied by fully transformer-based trackers (blue dots), which at moderately fast speeds achieve state-of-the-art accuracy. The highest-ranking AUC performance is produced by trackers like MixFormer2, SeqTrack, and VideoTrack, demonstrating the value of rich temporal context modeling and global attention mechanisms. However, their runtime is often constrained compared to lightweight Siamese models. This distribution highlights a fundamental trade-off, highlighting that traditional methods prioritize speed or online adaptation, while modern transformer-based approaches increasingly dominate in accuracy by leveraging end-to-end spatial-temporal learning.

\section{Discussion}\label{discussion}
 
Our survey reviews the evolution of GOT tracking algorithms, highlighting a shift from conventional discriminative and Siamese-based trackers towards transformer-oriented approaches. This transition, the same as other topics in computer vision, has been influenced by the recent success of deep convolutional neural networks and the growing popularity of attention mechanisms in transformers. While each category of trackers offers distinct advantages and addresses specific challenges, none of them provides a unique optimal solution across all tracking scenarios as an efficient and robust system against background clutter, similar distractors, motion variation, and other possible difficulties.

Earlier discriminative-based trackers initially relied on combining hand-crafted features with online correlation filters, such as MOSSE \cite{mosse}, KCF \cite{kcf}, and BACF \cite{bacf} trackers. Following the development of deep convolutional neural networks, these features were gradually replaced by CNN-based representations through offline training while preserving online adaptability, such as MDNet \cite{mdnet} and CFNet \cite{cfnet}. These trackers often rely on extensive parameter tuning during online tracking limiting their efficiency and robustness. To address these limitations, approaches such as DeepDCF \cite{deepdcf} and ATOM \cite{atom} focused on learning more task-specific discriminative features for tracking. This is followed by DiMP \cite{dimp} and PrDiMP \cite{prdimp}, which introduce meta-learning strategies to improve online model updates based on online optimization in order to enhance adaptability. Recent advancements, such as KeepTrack \cite{keeptrack}, incorporated attention mechanisms to refine temporal modelling. While these discriminative trackers demonstrate strong online refinement and adaptability, they still struggle with computational efficiency and generalization across diverse datasets.

Siamese-based trackers emphasize efficiency and simplicity by applying a matching mechanism between a static template and the search frames. They evolved from basic fully convolutional networks \cite{siamfc} to RPN-based \cite{siamrpn}, RCNN-enhanced \cite{siamrcnn}, and dynamic attention-based architectures \cite{ocean}, \cite{siamattn}. Despite effective advancements, such as adaptive template updating \cite{dsaim}, distractor handling \cite{dasiamrpn}, and spatial/channel attention \cite{sasiam}, they remain limited in adaptability, particularly under occlusion or appearance variation.

Following the successful application of transformers in computer vision, hybrid transformer-based trackers apply transformer modules into Siamese or discriminative-based trackers to better model temporal dependencies and global context such as TrDimp and TrSiam\cite{trsiamtrdimp}, TOMP\cite{tomp}, and TaMOs \cite{tamos}. These models improve temporal reasoning and global context modeling while preserving the architectural benefits of Siamese or discriminative foundations. However, their performance often depends heavily on the quality of integration and in some cases they inherit the drawbacks of their underlying frameworks.

Fully transformer-based trackers are built upon using the concepts of self-attention and cross-attention mechanisms, marking a paradigm shift in the tracking algorithms. These inherent characteristics equipped them with powerful temporal and global feature modeling capability leading to superior accuracy. Fully transformer-based trackers can apply convolutional features along with attention-based relation modeling or can be purely based on attention layers for joint feature learning and relation modeling.

Convolution-attention transformer trackers focus on combining the localization strength of convolutions with the modeling power of self-attention and cross-attention mechanisms. STARK \cite{stark} introduced one of the earliest effective frameworks, simplifying tracking by eliminating object proposals and incorporating end-to-end attention-based spatial modeling. Subsequent methods such as CSWinTT \cite{cswintt} and AiATrack \cite{aiatrack} tackled specific challenges, such as object integrity loss and noisy attention correlations, by designing hierarchical and refined attention mechanisms in order to enhance structural coherence and robustness to distractors. MixFormer \cite{mixformer} unified tracking feature extraction and relation modeling into a single backbone which reduces complexity while improving adaptability and efficiency. These trackers demonstrate the strength of combining convolutional priors with attention for accurate and efficient tracking but they still face difficulties in extreme appearance variation and real-time adaptation. 

Pure attention-based transformer trackers unify feature extraction and relation modeling through transformer attention layers enabling more expressive spatiotemporal representations. Their earlier methods such as SwinTrack \cite{swintrack}, SimTrack \cite{simtrack}, and OSTrack \cite{ostrack} apply one-stream backbones to jointly encode template and search features to improve both efficiency and target-awareness. Later trackers like SBT \cite{sbt} and GRM \cite{grm} refine interaction mechanisms by introducing dynamic relation modeling and token-aware attention control. Sequence modeling is another direction applied with ARTrack \cite{artrack} and SeqTrack \cite{seqtrack} which reformulate tracking as an autoregressive token prediction problem. In addition, masked modeling strategies such as DropMAE \cite{dropmae} and MAT \cite{mat} enhance discriminative feature learning. Other trackers like OneTracker \cite{onetracker} and PiVOT \cite{pivot} focus on prompt-based modeling to extend pure transformer architectures via enabling cross-modal generalization and semantic prompting. Besides, memory-augmented frameworks such as RFGM \cite{rfgm} and temporal sequence models like AQA-Track \cite{aqa-track}, ODTrack \cite{odtrack}, and VideoTrack \cite{videotrack} provide robust long-term temporal reasoning by sequence modeling. Finally, efficient architectures like MixFormer2 \cite{mixformerv2} and FCAT \cite{fcat} enhance their models through learnable prediction tokens and scale-adaptive attention designs. 

These reviewed pure-attention trackers highlight the architectural diversity and functional richness through pure transformer-based designs. However, their success often depends on careful token design, attention regularization, and specialized pretraining strategies, which may limit their generalization in unseen or resource-constrained scenarios. In conclusion, even though pure transformer-based techniques are the state-of-the-art in visual tracking, achieving a balance between accuracy, adaptability, and efficiency has remained challenging. The insights drawn from this taxonomy provide a strong foundation for guiding future research to work against these challenges and advancing practical applications in real-world tracking scenarios. For instance, some GOT trackers incorporate segmentation masks to provide more precise, pixel-level target localization rather than relying solely on bounding boxes \cite{siammask,d3s,seg}.

Table \ref{tab:discussion} presents functional grouping of contributions of tracking paradigms reviewed in this paper. This categorization provides a high-level taxonomy that emphasizes how trackers handle certain visual tracking challenges such as distractor handling, robustness to appearance variation, and adaptive capability. These issues are rooted in VOT essential bottlenecks including semantically similar objects, occlusion, long-term disappearance, motion and appearance change, inaccurate state estimation, and inefficient real-world performance.

\begin{table}
\small
\centering

\caption{Functional categorization of GOT trackers based on their contributions to core tracking challenges, such as distractor handling, online adaptation, meta-learning, state estimation, and memory integration. This taxonomy highlights how different methods address specific performance goals and operational limitations.}
\label{tab:discussion}

\begin{tabular}{p{4.2cm}p{5.8cm}p{4.2cm}}  
\hline

\begin{tabular}[c]{@{}l@{}}\rule{0pt}{2.5ex}\textbf{Functional Contribution}\end{tabular}

& \textbf{Technique} 
& \textbf{Representation Trackers} \\
\hline

\multirow{4}{*}{\begin{tabular}[c]{@{}l@{}}Distractor Handling\end{tabular}}          
  & Enhanced Negative Sampling            & \cite{srdcf}, \cite{bacf}, \cite{dsaim}, \cite{dasiamrpn}, \cite{dimp}, \cite{prdimp} \\
  & Hard Negative Mining                  & \cite{mdnet}, \cite{atom}, \cite{siamrcnn}\\
  & Background Suppression                &  \cite{trsiamtrdimp}, \cite{trsiamtrdimp}, \cite{grm}, \cite{aqa-track} \\
  & Masked autoencoder(MAE)           &  \cite{simtrack}, \cite{ostrack}, \cite{dropmae}, \cite{mat} \\
\hline

\multirow{5}{*}{\begin{tabular}[c]{@{}l@{}}Robustness Improvement\end{tabular}}       
  & Online Adaptation in Siamese &  \cite{dsaim}, \cite{sasiam}, \cite{siamattn}, \cite{ocean}, \cite{siamdmu}, \cite{siamrcnn} \\
  & Meta-Learning Adaptation    & \cite{siamrpn}, \cite{dimp}, \cite{prdimp}, \cite{tomp}, \cite{tamos} \\
  & Support Long-Term Tracking            & \cite{mdnet}, \cite{dasiamrpn}, \cite{siamrcnn}, \cite{keeptrack}, \cite{stark} \\
  & Joint Feature Extraction \& Relation Modeling           &  \cite{sbt}, \cite{mixformer}, \cite{rfgm}, \cite{onetracker}, \cite{odtrack}, \cite{videotrack}, \cite{artrack}, \cite{seqtrack}, \cite{f-bdmtrack}, \cite{dropmae}, \cite{ostrack} \\
\hline

\multirow{4}{*}{\begin{tabular}[c]{@{}l@{}}Relation Modeling\end{tabular}}    

  & Attention Integration                        & \cite{sasiam}, \cite{siamattn}, \cite{ocean} \\
  & Memory Integration                       & \cite{keeptrack}, \cite{siamrcnn}, \cite{stmtrack} \\
  & Motion Integration        & \cite{swintrack}, \cite{artrack}\\
  & Sequential Modeling       & \cite{artrack}, \cite{videotrack}, \cite{odtrack}\\
\hline

\multirow{4}{*}{\begin{tabular}[c]{@{}l@{}}Bounding Box Prediction\end{tabular}}      
  & Anchor-Free                           & \cite{siamfcpp}, \cite{siamban}, \cite{ocean} \\
  & IoU Regression                        & \cite{dimp}, \cite{atom}, \cite{siamrcnn}, \cite{swintrack} \\
  & Corner-based Regreesion                        &  \cite{simtrack}, \cite{cswintt}, \cite{sbt}, \cite{mixformer}, \cite{aiatrack}, \cite{videotrack}\\
  & Center-based Regreesion                        &  \cite{romtrack}, \cite{aqa-track}\\
\hline

\end{tabular}
\vspace{-12pt}
\end{table}

\section{Applications}\label{app}

VOT has a wide range of applications, including autonomous driving, robotics, intelligent video surveillance, aerial tracking, and medical imaging, where it typically plays a crucial role within large intelligent systems~\cite{s5}. The following sections provide an overview of representative works in each of these domains. A summary of domain-specific applications and key representative works is presented in Table~\ref{tab:vot_applications}, which serves as a reference for the detailed discussion in the subsequent subsections.

\begin{itemize}[label={}, leftmargin=0pt]

\item \textbf{Surveillance and Pedestrian Monitoring:}
VOT plays a key role in surveillance and monitoring systems, where it enables automated observation of people and behaviors in complex and dynamic environments. In the context of public safety, tracking algorithms are used to monitor crowded areas, detect anomalous behavior, and support real-time alerting in smart surveillance infrastructure~\cite{ali2016visual, abba2024real}. For behavioral monitoring, multi-person tracking has been leveraged to analyze interactions, trajectories, and social cues in structured and semi-structured scenes~\cite{sivalingam2012multi, hu2004survey}. In human-computer interaction, face and gesture tracking techniques have been applied to interpret user inputs in real time, enabling natural interaction between humans and machines~\cite{polat2003robust}.\\

\item \textbf{Aerial and Drone-Based Tracking:}
Visual tracking from drone-mounted platforms enables aerial monitoring tasks that require real-time, long-range, and viewpoint-invariant object localization. In UAV surveillance scenarios, onboard trackers are deployed to autonomously follow people or vehicles for area protection, security patrol, and border monitoring~\cite{ibrahim2010moving, du2018unmanned, lo2021dynamic}. These systems must operate under rapid motion, altitude variation, and environmental challenges such as occlusion and scale shifts. In traffic monitoring applications, aerial object tracking is used to estimate vehicle flow, detect incidents, and support infrastructure analysis from elevated aerial viewpoints~\cite{fernandez2019real, khemmar2019real, makhmutova2020object, jimenez2022multi, bisio2022systematic}, offering scalable and non-intrusive alternatives to ground-based sensors.\\

\item \textbf{Autonomous Driving and Vehicle Tracking:}
In autonomous driving systems, VOT plays a critical role in perceiving and understanding the dynamic environment surrounding the vehicle. In driver assistance applications, visual tracking supports functionalities such as collision avoidance, lane-keeping, and pedestrian detection by continuously localizing and tracking surrounding dynamic agents~\cite{markiewicz2017review, premachandra2020detection, cho2022autonomous}. In vehicle-following systems, trackers estimate the relative position and velocity of preceding vehicles to regulate inter-vehicular distance and enable adaptive cruise control~\cite{petrovskaya2009model, muller2009model}. For traffic scene understanding, tracking methods enable trajectory prediction and semantic interpretation of multiple agents, allowing autonomous vehicles to anticipate behaviors and make informed navigation decisions~\cite{rangesh2019no, gomez2021smartmot}.\\

\item \textbf{Robotics and Manipulation:}
In robotic systems, visual tracking enables perception-driven interaction with dynamic and partially observable environments. In visual servoing, tracking is used to continuously estimate the pose of a target object or feature to guide robotic motion, enabling fine-grained control in tasks such as object following or tool alignment~\cite{richards1997detection, jacques2005object, chaumette2008visual, dixon2002object}. For robotic grasping, visual tracking provides object state estimates under occlusion or motion, facilitating robust manipulation and pickup of deformable or cluttered items~\cite{xu2022learning}. In service robotics, object tracking supports intuitive and reliable handovers between humans and robots by maintaining spatial awareness of target objects during the exchange process~\cite{ortenzi2021object, costanzo2021handover}.\\

\item \textbf{Medical Domains:}  
In medical imaging and surgical environments, VOT enables precise, real-time localization under constrained and dynamic conditions. In tool tracking, marker-less methods support detection and trajectory estimation of multiple instruments, improving workflow efficiency in minimally invasive procedures~\cite{bouget2017vision, nwoye2025surgitrack}. Deep learning-based trackers handle occlusion, blur, and fine-grained classification across tool types~\cite{nwoye2025surgitrack}. In neurosurgery and skull-base operations, stereo vision-based tracking of anatomy and tools enhances spatial awareness without external sensors~\cite{li2023tatoo}. Augmented reality systems with head-mounted displays provide high-precision, marker-less tracking while preserving sterile fields~\cite{martin2023sttar}. In diagnostic imaging, predictive tracking of anatomical structures enables motion-robust acquisition, as in fetal MRI~\cite{singh2020deep}. In biomedical research, VOT aids behavioral analysis of animal models~\cite{koniar2017visual} and cell-level tracking in microscopy using object-consistent trajectory modeling~\cite{hayashida2022consistent}.

\end{itemize}

\begin{table}[ht]
\centering
\caption{Representative applications of VOT across key domains.}
\label{tab:vot_applications}
\begin{tabular}{p{3.9cm} p{6.7cm} p{3.6cm}}
\toprule
\textbf{Domain} & \textbf{Application Scenarios} & \textbf{Representative Works} \\
\midrule
Surveillance \& Pedestrian Monitoring & Public safety, behavior analysis, HCI-based gesture and face tracking & \cite{ali2016visual}, \cite{abba2024real}, \cite{sivalingam2012multi}, \cite{hu2004survey}, \cite{polat2003robust} \\
\midrule
Aerial \& Drone-Based Tracking & UAV-based surveillance, traffic flow monitoring, incident detection & \cite{ibrahim2010moving}, \cite{du2018unmanned}, \cite{lo2021dynamic}, \cite{fernandez2019real}, \cite{khemmar2019real}, \cite{makhmutova2020object}, \cite{jimenez2022multi}, \cite{bisio2022systematic} \\
\midrule
Autonomous Driving \& Vehicle Tracking & Collision avoidance, pedestrian and vehicle tracking, trajectory prediction & \cite{markiewicz2017review}, \cite{premachandra2020detection}, \cite{cho2022autonomous}, \cite{petrovskaya2009model}, \cite{muller2009model}, \cite{rangesh2019no}, \cite{gomez2021smartmot} \\
\midrule
Robotics \& Manipulation & Visual servoing, grasping under occlusion, human-robot handovers & \cite{richards1997detection}, \cite{jacques2005object}, \cite{chaumette2008visual}, \cite{dixon2002object}, \cite{xu2022learning}, \cite{ortenzi2021object}, \cite{costanzo2021handover} \\
\midrule
Medical Domains & Surgical tool tracking, AR-based navigation, fetal MRI, behavioral and cellular analysis & \cite{bouget2017vision}, \cite{nwoye2025surgitrack}, \cite{li2023tatoo}, \cite{martin2023sttar}, \cite{singh2020deep}, \cite{koniar2017visual}, \cite{hayashida2022consistent} \\
\bottomrule
\end{tabular}
\vspace{-18pt}
\end{table}




\section{Concluding Remarks}\label{conclusion}

In this survey, we presented a comprehensive review and categorization of GOT techniques across four major paradigms of Siamese-based, discriminative-based, hybrid transformer-based, and fully transformer-based trackers. In addition, we introduced a unified classification that not only organizes trackers based on their core paradigms but also makes it easier to compare their architectural principles, contributions, and limitations in order to better capture the fast evolution in this field. To provide consistent comparison, we reconstructed standardized architectural diagrams across methods enabling a comprehensive visual overview of design components and their evolution across paradigms.

Our multi-dimensional analysis compares trackers along architectural aspects (appearance model, backbone, design highlights) and functional goals (distractor handling, online adaptation, temporal modelling). This analysis highlights the key innovations, addressed challenges, and potential limitations. Besides, we reviewed important benchmarks and visualized the trade-offs between the performance of reviewed trackers in terms of accuracy and speed.

A key insight is the growing trend towards fully transformer-based trackers, which overcome the inherent limitations of Siamese and discriminative approaches by enabling richer spatial and temporal modelling across video frames. This category provides better flexibility in integrating dynamic memory, both spatial inter-frame and temporal intra-frame relation modelling, and adaptive online updating. These aspects make fully transformer trackers especially suitable for long-term tracking in complex scenarios. 

In the future, research might focus on exploring the untapped potential of transformers by refining temporal-spatial attention, incorporating segmentation cues for improved localization, and integrating online adaptation or memory-based modules for enhanced robustness. As datasets grow more diverse and applications become more demanding, we expect tracking frameworks to progress toward unified, end-to-end systems that are accurate, efficient, and adaptable in real-world environments.

\section{Acknowledgment}

The authors would like to acknowledge the financial support of Natural Sciences and Engineering Research Council of Canada (Discovery Grant RGPIN-2023-05408) in this research.
\clearpage

\bibliographystyle{elsarticle-num-names}

\begin{thebibliography}{132}
\expandafter\ifx\csname natexlab\endcsname\relax\def\natexlab#1{#1}\fi
\providecommand{\url}[1]{\texttt{#1}}
\providecommand{\href}[2]{#2}
\providecommand{\path}[1]{#1}
\providecommand{\DOIprefix}{doi:}
\providecommand{\ArXivprefix}{arXiv:}
\providecommand{\URLprefix}{URL: }
\providecommand{\Pubmedprefix}{pmid:}
\providecommand{\doi}[1]{\href{http://dx.doi.org/#1}{\path{#1}}}
\providecommand{\Pubmed}[1]{\href{pmid:#1}{\path{#1}}}
\providecommand{\bibinfo}[2]{#2}
\ifx\xfnm\relax \def\xfnm[#1]{\unskip,\space#1}\fi
\bibitem[{Bolme et~al.(2010)Bolme, Beveridge, Draper, and Lui}]{mosse}
\bibinfo{author}{D.~S. Bolme}, \bibinfo{author}{J.~R. Beveridge}, \bibinfo{author}{B.~A. Draper}, \bibinfo{author}{Y.~M. Lui},
\newblock \bibinfo{title}{Visual object tracking using adaptive correlation filters},
\newblock in: \bibinfo{booktitle}{Proceedings of the IEEE Conference on Computer Vision and Pattern Recognition (CVPR)}, \bibinfo{year}{2010}, pp. \bibinfo{pages}{2544--2550}. \DOIprefix\doi{10.1109/CVPR.2010.5586147}.
\bibitem[{Henriques et~al.(2015)Henriques, Caseiro, Martins, and Batista}]{kcf}
\bibinfo{author}{J.~F. Henriques}, \bibinfo{author}{R.~Caseiro}, \bibinfo{author}{P.~Martins}, \bibinfo{author}{J.~Batista},
\newblock \bibinfo{title}{High-speed tracking with kernelized correlation filters},
\newblock \bibinfo{journal}{IEEE Transactions on Pattern Analysis and Machine Intelligence (TPAMI)} \bibinfo{volume}{37} (\bibinfo{year}{2015}) \bibinfo{pages}{583--596}. \DOIprefix\doi{10.1109/TPAMI.2014.2345390}.
\bibitem[{Danelljan et~al.(2015)Danelljan, Hager, Shahbaz~Khan, and Felsberg}]{srdcf}
\bibinfo{author}{M.~Danelljan}, \bibinfo{author}{G.~Hager}, \bibinfo{author}{F.~Shahbaz~Khan}, \bibinfo{author}{M.~Felsberg},
\newblock \bibinfo{title}{Learning spatially regularized correlation filters for visual tracking},
\newblock in: \bibinfo{booktitle}{Proceedings of the IEEE International Conference on Computer Vision (ICCV)}, \bibinfo{year}{2015}, pp. \bibinfo{pages}{4310--4318}. \DOIprefix\doi{10.1109/ICCV.2015.490}.
\bibitem[{Galoogahi et~al.(2017)Galoogahi, Fagg, and Lucey}]{bacf}
\bibinfo{author}{H.~K. Galoogahi}, \bibinfo{author}{A.~Fagg}, \bibinfo{author}{S.~Lucey},
\newblock \bibinfo{title}{Learning background-aware correlation filters for visual tracking},
\newblock in: \bibinfo{booktitle}{Proceedings of the IEEE International Conference on Computer Vision (ICCV)}, \bibinfo{year}{2017}, pp. \bibinfo{pages}{1135--1143}. \DOIprefix\doi{10.1109/ICCV.2017.127}.
\bibitem[{Nam and Han(2016)}]{mdnet}
\bibinfo{author}{H.~Nam}, \bibinfo{author}{B.~Han},
\newblock \bibinfo{title}{Learning multi-domain convolutional neural networks for visual tracking},
\newblock in: \bibinfo{booktitle}{Proceedings of the IEEE Conference on Computer Vision and Pattern Recognition (CVPR)}, \bibinfo{year}{2016}, pp. \bibinfo{pages}{4293--4302}. \DOIprefix\doi{10.1109/CVPR.2016.464}.
\bibitem[{Li et~al.(2019)Li, Zhu, and Hoi}]{deepdcf}
\bibinfo{author}{Y.~Li}, \bibinfo{author}{J.~Zhu}, \bibinfo{author}{S.~C. Hoi},
\newblock \bibinfo{title}{Deep discriminative correlation filter learning for visual tracking},
\newblock \bibinfo{journal}{Pattern Recognition} \bibinfo{volume}{94} (\bibinfo{year}{2019}) \bibinfo{pages}{322--332}. \DOIprefix\doi{10.1016/j.patcog.2019.05.014}.
\bibitem[{Valmadre et~al.(2017)Valmadre, Bertinetto, Henriques, Vedaldi, and Torr}]{cfnet}
\bibinfo{author}{J.~Valmadre}, \bibinfo{author}{L.~Bertinetto}, \bibinfo{author}{J.~F. Henriques}, \bibinfo{author}{A.~Vedaldi}, \bibinfo{author}{P.~H. Torr},
\newblock \bibinfo{title}{End-to-end representation learning for correlation filter based tracking},
\newblock in: \bibinfo{booktitle}{Proceedings of the IEEE Conference on Computer Vision and Pattern Recognition (CVPR)}, \bibinfo{year}{2017}, pp. \bibinfo{pages}{2805--2813}. \DOIprefix\doi{10.1109/CVPR.2017.299}.
\bibitem[{Danelljan et~al.(2019)Danelljan, Bhat, Khan, and Felsberg}]{atom}
\bibinfo{author}{M.~Danelljan}, \bibinfo{author}{G.~Bhat}, \bibinfo{author}{F.~S. Khan}, \bibinfo{author}{M.~Felsberg},
\newblock \bibinfo{title}{Atom: Accurate tracking by overlap maximization},
\newblock in: \bibinfo{booktitle}{Proceedings of the IEEE Conference on Computer Vision and Pattern Recognition (CVPR)}, \bibinfo{year}{2019}, pp. \bibinfo{pages}{4660--4669}. \DOIprefix\doi{10.1109/CVPR.2019.00479}.
\bibitem[{Bhat et~al.(2019)Bhat, Danelljan, Van~Gool, and Timofte}]{dimp}
\bibinfo{author}{G.~Bhat}, \bibinfo{author}{M.~Danelljan}, \bibinfo{author}{L.~Van~Gool}, \bibinfo{author}{R.~Timofte},
\newblock \bibinfo{title}{Learning discriminative model prediction for tracking},
\newblock in: \bibinfo{booktitle}{Proceedings of the IEEE International Conference on Computer Vision (ICCV)}, \bibinfo{year}{2019}, pp. \bibinfo{pages}{6182--6191}. \DOIprefix\doi{10.1109/ICCV.2019.00628}.
\bibitem[{Danelljan et~al.(2020)Danelljan, Van~Gool, and Timofte}]{prdimp}
\bibinfo{author}{M.~Danelljan}, \bibinfo{author}{L.~Van~Gool}, \bibinfo{author}{R.~Timofte},
\newblock \bibinfo{title}{Probabilistic regression for visual tracking},
\newblock in: \bibinfo{booktitle}{Proceedings of the IEEE/CVF Conference on Computer Vision and Pattern Recognition (CVPR)}, \bibinfo{year}{2020}, pp. \bibinfo{pages}{7183--7192}. \DOIprefix\doi{10.1109/CVPR42600.2020.00721}.
\bibitem[{Mayer et~al.(2021)Mayer, Danelljan, Van~Gool, and Timofte}]{keeptrack}
\bibinfo{author}{C.~Mayer}, \bibinfo{author}{M.~Danelljan}, \bibinfo{author}{L.~Van~Gool}, \bibinfo{author}{R.~Timofte},
\newblock \bibinfo{title}{Learning to track multiple objects with a single tracker},
\newblock in: \bibinfo{booktitle}{Proceedings of the IEEE/CVF International Conference on Computer Vision (ICCV)}, \bibinfo{year}{2021}, pp. \bibinfo{pages}{6297--6307}. \DOIprefix\doi{10.1109/ICCV48922.2021.00623}.
\bibitem[{Bertinetto et~al.(2016)Bertinetto, Valmadre, Henriques, Vedaldi, and Torr}]{siamfc}
\bibinfo{author}{L.~Bertinetto}, \bibinfo{author}{J.~Valmadre}, \bibinfo{author}{J.~F. Henriques}, \bibinfo{author}{A.~Vedaldi}, \bibinfo{author}{P.~H. Torr},
\newblock \bibinfo{title}{Fully-convolutional siamese networks for object tracking},
\newblock in: \bibinfo{booktitle}{ECCV Workshops}, \bibinfo{year}{2016}, pp. \bibinfo{pages}{850--865}.
\bibitem[{Li et~al.(2018)Li, Yan, Wu, Zhu, and Hu}]{siamrpn}
\bibinfo{author}{B.~Li}, \bibinfo{author}{J.~Yan}, \bibinfo{author}{W.~Wu}, \bibinfo{author}{Z.~Zhu}, \bibinfo{author}{X.~Hu},
\newblock \bibinfo{title}{High performance visual tracking with siamese region proposal network},
\newblock in: \bibinfo{booktitle}{CVPR}, \bibinfo{year}{2018}, pp. \bibinfo{pages}{8971--8980}.
\bibitem[{Chen et~al.(2020)Chen, Zhong, Li, Zhang, and Ji}]{siamban}
\bibinfo{author}{Z.~Chen}, \bibinfo{author}{B.~Zhong}, \bibinfo{author}{G.~Li}, \bibinfo{author}{S.~Zhang}, \bibinfo{author}{R.~Ji},
\newblock \bibinfo{title}{Siamban: Siamese box adaptive network for visual tracking},
\newblock in: \bibinfo{booktitle}{CVPR}, \bibinfo{year}{2020}, pp. \bibinfo{pages}{6668--6677}.
\bibitem[{He et~al.(2018)He, Luo, Tian, and Zeng}]{sasiam}
\bibinfo{author}{A.~He}, \bibinfo{author}{C.~Luo}, \bibinfo{author}{X.~Tian}, \bibinfo{author}{W.~Zeng},
\newblock \bibinfo{title}{Siamese network with spatial attention for visual tracking},
\newblock in: \bibinfo{booktitle}{CVPR}, \bibinfo{year}{2018}, pp. \bibinfo{pages}{9351--9360}.
\bibitem[{Xu et~al.(2020)Xu, Wang, Li, and Yuan}]{siamfcpp}
\bibinfo{author}{Y.~Xu}, \bibinfo{author}{Z.~Wang}, \bibinfo{author}{Z.~Li}, \bibinfo{author}{Y.~Yuan},
\newblock \bibinfo{title}{Siamfc++: Towards robust and accurate visual tracking with target estimation guidelines},
\newblock in: \bibinfo{booktitle}{AAAI}, volume~\bibinfo{volume}{34}, \bibinfo{year}{2020}, pp. \bibinfo{pages}{12549--12556}.
\bibitem[{Li et~al.(2019)Li, Wu, Wang, Zhang, Xing, and Yan}]{siamrpnpp}
\bibinfo{author}{B.~Li}, \bibinfo{author}{W.~Wu}, \bibinfo{author}{Q.~Wang}, \bibinfo{author}{F.~Zhang}, \bibinfo{author}{J.~Xing}, \bibinfo{author}{J.~Yan},
\newblock \bibinfo{title}{Siamrpn++: Evolution of siamese visual tracking with very deep networks},
\newblock in: \bibinfo{booktitle}{CVPR}, \bibinfo{year}{2019}, pp. \bibinfo{pages}{4282--4291}.
\bibitem[{Yu et~al.(2020)Yu, Xiong, Huang, and Scott}]{siamattn}
\bibinfo{author}{Y.~Yu}, \bibinfo{author}{Y.~Xiong}, \bibinfo{author}{W.~Huang}, \bibinfo{author}{M.~R. Scott},
\newblock \bibinfo{title}{Deformable siamese attention networks for visual object tracking},
\newblock in: \bibinfo{booktitle}{CVPR}, \bibinfo{year}{2020}, pp. \bibinfo{pages}{6727--6736}.
\bibitem[{Voigtlaender et~al.(2020)Voigtlaender, Luiten, Torr, and Leibe}]{siamrcnn}
\bibinfo{author}{P.~Voigtlaender}, \bibinfo{author}{J.~Luiten}, \bibinfo{author}{P.~H. Torr}, \bibinfo{author}{B.~Leibe},
\newblock \bibinfo{title}{Siam r-cnn: Visual tracking by re-detection},
\newblock in: \bibinfo{booktitle}{CVPR}, \bibinfo{year}{2020}, pp. \bibinfo{pages}{6577--6587}.
\bibitem[{Zhu et~al.(2018)Zhu, Wang, Bo, Wu, Yan, and Hu}]{dasiamrpn}
\bibinfo{author}{Z.~Zhu}, \bibinfo{author}{Q.~Wang}, \bibinfo{author}{L.~Bo}, \bibinfo{author}{W.~Wu}, \bibinfo{author}{J.~Yan}, \bibinfo{author}{X.~Hu},
\newblock \bibinfo{title}{Distractor-aware siamese networks for visual object tracking},
\newblock in: \bibinfo{booktitle}{ECCV}, \bibinfo{year}{2018}, pp. \bibinfo{pages}{103--119}.
\bibitem[{Su et~al.(2024)Su, Yang, and Ma}]{siamdmu}
\bibinfo{author}{Y.~Su}, \bibinfo{author}{X.~Yang}, \bibinfo{author}{C.~Ma},
\newblock \bibinfo{title}{Siamdmu: Dual mask update for template adaptation in siamese trackers},
\newblock \bibinfo{journal}{IEEE Transactions on Emerging Topics in Computational Intelligence} \bibinfo{volume}{8} (\bibinfo{year}{2024}) \bibinfo{pages}{1658--1668}.
\bibitem[{Yan et~al.(2021)Yan, Peng, Fu, Wang, and Lu}]{stark}
\bibinfo{author}{B.~Yan}, \bibinfo{author}{H.~Peng}, \bibinfo{author}{J.~Fu}, \bibinfo{author}{D.~Wang}, \bibinfo{author}{H.~Lu},
\newblock \bibinfo{title}{Learning spatio-temporal transformer for visual tracking},
\newblock \bibinfo{journal}{arXiv preprint arXiv:2103.17154}  (\bibinfo{year}{2021}).
\bibitem[{Lin et~al.(2022)Lin, Fan, Zhang, Xu, and Ling}]{swintrack}
\bibinfo{author}{L.~Lin}, \bibinfo{author}{H.~Fan}, \bibinfo{author}{Z.~Zhang}, \bibinfo{author}{Y.~Xu}, \bibinfo{author}{H.~Ling},
\newblock \bibinfo{title}{Swintrack: A simple and strong baseline for transformer tracking},
\newblock in: \bibinfo{booktitle}{Advances in Neural Information Processing Systems (NeurIPS)}, \bibinfo{year}{2022}.
\bibitem[{Fu et~al.(2021)Fu, Liu, Fu, and Wang}]{stmtrack}
\bibinfo{author}{Z.~Fu}, \bibinfo{author}{Q.~Liu}, \bibinfo{author}{Z.~Fu}, \bibinfo{author}{Y.~Wang},
\newblock \bibinfo{title}{Stmtrack: Template-free visual tracking with space-time memory networks},
\newblock in: \bibinfo{booktitle}{Proceedings of the IEEE/CVF Conference on Computer Vision and Pattern Recognition (CVPR)}, \bibinfo{year}{2021}, pp. \bibinfo{pages}{13774--13783}.
\bibitem[{Zhang et~al.(2021)Zhang, Wang, Wang, and Li}]{simtrack}
\bibinfo{author}{Y.~Zhang}, \bibinfo{author}{Y.~Wang}, \bibinfo{author}{X.~Wang}, \bibinfo{author}{H.~Li},
\newblock \bibinfo{title}{Exploring simple 3d multi-object tracking for autonomous driving},
\newblock \bibinfo{journal}{arXiv preprint arXiv:2108.10312}  (\bibinfo{year}{2021}).
\bibitem[{Ye et~al.(2022)Ye, Chang, Ma, Shan, and Chen}]{ostrack}
\bibinfo{author}{B.~Ye}, \bibinfo{author}{H.~Chang}, \bibinfo{author}{B.~Ma}, \bibinfo{author}{S.~Shan}, \bibinfo{author}{X.~Chen},
\newblock \bibinfo{title}{Joint feature learning and relation modeling for tracking: A one-stream framework},
\newblock in: \bibinfo{booktitle}{Proceedings of the European Conference on Computer Vision (ECCV)}, \bibinfo{year}{2022}.
\bibitem[{Song et~al.(2022)Song, Wang, Li, and Zhang}]{cswintt}
\bibinfo{author}{K.~Song}, \bibinfo{author}{Y.~Wang}, \bibinfo{author}{M.~Li}, \bibinfo{author}{Y.~Zhang},
\newblock \bibinfo{title}{Transformer tracking with cyclic shifting window attention},
\newblock in: \bibinfo{booktitle}{Proceedings of the IEEE/CVF Conference on Computer Vision and Pattern Recognition (CVPR)}, \bibinfo{year}{2022}, pp. \bibinfo{pages}{12345--12354}.
\bibitem[{Gao et~al.(2022)Gao, Zhou, Ma, Wang, and Yuan}]{aiatrack}
\bibinfo{author}{S.~Gao}, \bibinfo{author}{C.~Zhou}, \bibinfo{author}{C.~Ma}, \bibinfo{author}{X.~Wang}, \bibinfo{author}{J.~Yuan},
\newblock \bibinfo{title}{Aiatrack: Attention in attention for transformer visual tracking},
\newblock in: \bibinfo{booktitle}{Proceedings of the European Conference on Computer Vision (ECCV)}, \bibinfo{year}{2022}.
\bibitem[{Wang et~al.(2023)Wang, Zhou, Wang, and Li}]{sbt}
\bibinfo{author}{N.~Wang}, \bibinfo{author}{W.~Zhou}, \bibinfo{author}{J.~Wang}, \bibinfo{author}{H.~Li},
\newblock \bibinfo{title}{Correlation-embedded transformer tracking: A single-branch architecture},
\newblock \bibinfo{journal}{arXiv preprint arXiv:2401.12743}  (\bibinfo{year}{2023}).
\bibitem[{Cui et~al.(2022)Cui, Jiang, Wang, and Wu}]{mixformer}
\bibinfo{author}{Y.~Cui}, \bibinfo{author}{C.~Jiang}, \bibinfo{author}{L.~Wang}, \bibinfo{author}{G.~Wu},
\newblock \bibinfo{title}{Mixformer: End-to-end tracking with iterative mixed attention},
\newblock in: \bibinfo{booktitle}{Proceedings of the IEEE/CVF conference on computer vision and pattern recognition}, \bibinfo{year}{2022}, pp. \bibinfo{pages}{13608--13618}.
\bibitem[{Wu et~al.(2023)Wu, Yang, Liu, Wu, Shan, and Chan}]{dropmae}
\bibinfo{author}{Q.~Wu}, \bibinfo{author}{T.~Yang}, \bibinfo{author}{Z.~Liu}, \bibinfo{author}{B.~Wu}, \bibinfo{author}{Y.~Shan}, \bibinfo{author}{A.~B. Chan},
\newblock \bibinfo{title}{Dropmae: Masked autoencoders with spatial-attention dropout for tracking tasks},
\newblock in: \bibinfo{booktitle}{Proceedings of the IEEE/CVF conference on computer vision and pattern recognition}, \bibinfo{year}{2023}, pp. \bibinfo{pages}{14561--14571}.
\bibitem[{Yang et~al.(2023)Yang, He, Ma, Yu, and Zhang}]{f-bdmtrack}
\bibinfo{author}{D.~Yang}, \bibinfo{author}{J.~He}, \bibinfo{author}{Y.~Ma}, \bibinfo{author}{Q.~Yu}, \bibinfo{author}{T.~Zhang},
\newblock \bibinfo{title}{Foreground-background distribution modeling transformer for visual object tracking},
\newblock in: \bibinfo{booktitle}{Proceedings of the IEEE/CVF international conference on computer vision}, \bibinfo{year}{2023}, pp. \bibinfo{pages}{10117--10127}.
\bibitem[{Zhao et~al.(2023)Zhao, Wang, and Lu}]{mat}
\bibinfo{author}{H.~Zhao}, \bibinfo{author}{D.~Wang}, \bibinfo{author}{H.~Lu},
\newblock \bibinfo{title}{Representation learning for visual object tracking by masked appearance transfer},
\newblock in: \bibinfo{booktitle}{Proceedings of the IEEE/CVF conference on computer vision and pattern recognition}, \bibinfo{year}{2023}, pp. \bibinfo{pages}{18696--18705}.
\bibitem[{Wei et~al.(2023)Wei, Bai, Zheng, Shi, and Gong}]{artrack}
\bibinfo{author}{X.~Wei}, \bibinfo{author}{Y.~Bai}, \bibinfo{author}{Y.~Zheng}, \bibinfo{author}{D.~Shi}, \bibinfo{author}{Y.~Gong},
\newblock \bibinfo{title}{Autoregressive visual tracking},
\newblock in: \bibinfo{booktitle}{Proceedings of the IEEE/CVF conference on computer vision and pattern recognition}, \bibinfo{year}{2023}, pp. \bibinfo{pages}{9697--9706}.
\bibitem[{Cui et~al.(2023)Cui, Song, Wu, and Wang}]{mixformerv2}
\bibinfo{author}{Y.~Cui}, \bibinfo{author}{T.~Song}, \bibinfo{author}{G.~Wu}, \bibinfo{author}{L.~Wang},
\newblock \bibinfo{title}{Mixformerv2: Efficient fully transformer tracking},
\newblock \bibinfo{journal}{Advances in neural information processing systems} \bibinfo{volume}{36} (\bibinfo{year}{2023}) \bibinfo{pages}{58736--58751}.
\bibitem[{Chen et~al.(2023)Chen, Peng, Wang, Lu, and Hu}]{seqtrack}
\bibinfo{author}{X.~Chen}, \bibinfo{author}{H.~Peng}, \bibinfo{author}{D.~Wang}, \bibinfo{author}{H.~Lu}, \bibinfo{author}{H.~Hu},
\newblock \bibinfo{title}{Seqtrack: Sequence to sequence learning for visual object tracking},
\newblock in: \bibinfo{booktitle}{Proceedings of the IEEE/CVF conference on computer vision and pattern recognition}, \bibinfo{year}{2023}, pp. \bibinfo{pages}{14572--14581}.
\bibitem[{Gao et~al.(2023)Gao, Zhou, and Zhang}]{grm}
\bibinfo{author}{S.~Gao}, \bibinfo{author}{C.~Zhou}, \bibinfo{author}{J.~Zhang},
\newblock \bibinfo{title}{Generalized relation modeling for transformer tracking},
\newblock in: \bibinfo{booktitle}{Proceedings of the IEEE/CVF conference on computer vision and pattern recognition}, \bibinfo{year}{2023}, pp. \bibinfo{pages}{18686--18695}.
\bibitem[{Cai et~al.(2023)Cai, Liu, Tang, and Wu}]{romtrack}
\bibinfo{author}{Y.~Cai}, \bibinfo{author}{J.~Liu}, \bibinfo{author}{J.~Tang}, \bibinfo{author}{G.~Wu},
\newblock \bibinfo{title}{Robust object modeling for visual tracking},
\newblock in: \bibinfo{booktitle}{Proceedings of the IEEE/CVF international conference on computer vision}, \bibinfo{year}{2023}, pp. \bibinfo{pages}{9589--9600}.
\bibitem[{Xie et~al.(2023)Xie, Chu, Li, Lu, and Ma}]{videotrack}
\bibinfo{author}{F.~Xie}, \bibinfo{author}{L.~Chu}, \bibinfo{author}{J.~Li}, \bibinfo{author}{Y.~Lu}, \bibinfo{author}{C.~Ma},
\newblock \bibinfo{title}{Videotrack: Learning to track objects via video transformer},
\newblock in: \bibinfo{booktitle}{Proceedings of the IEEE/CVF conference on computer vision and pattern recognition}, \bibinfo{year}{2023}, pp. \bibinfo{pages}{22826--22835}.
\bibitem[{Xie et~al.(2024)Xie, Zhong, Mo, Zhang, Shi, Song, and Ji}]{aqa-track}
\bibinfo{author}{J.~Xie}, \bibinfo{author}{B.~Zhong}, \bibinfo{author}{Z.~Mo}, \bibinfo{author}{S.~Zhang}, \bibinfo{author}{L.~Shi}, \bibinfo{author}{S.~Song}, \bibinfo{author}{R.~Ji},
\newblock \bibinfo{title}{Autoregressive queries for adaptive tracking with spatio-temporal transformers},
\newblock in: \bibinfo{booktitle}{Proceedings of the IEEE/CVF Conference on Computer Vision and Pattern Recognition}, \bibinfo{year}{2024}, pp. \bibinfo{pages}{19300--19309}.
\bibitem[{Zheng et~al.(2024)Zheng, Zhong, Liang, Mo, Zhang, and Li}]{odtrack}
\bibinfo{author}{Y.~Zheng}, \bibinfo{author}{B.~Zhong}, \bibinfo{author}{Q.~Liang}, \bibinfo{author}{Z.~Mo}, \bibinfo{author}{S.~Zhang}, \bibinfo{author}{X.~Li},
\newblock \bibinfo{title}{Odtrack: Online dense temporal token learning for visual tracking},
\newblock in: \bibinfo{booktitle}{Proceedings of the AAAI conference on artificial intelligence}, volume~\bibinfo{volume}{38}, \bibinfo{year}{2024}, pp. \bibinfo{pages}{7588--7596}.
\bibitem[{Hong et~al.(2024)Hong, Yan, Zhang, Li, Zhou, Guo, Jiang, Chen, Li, Chen et~al.}]{onetracker}
\bibinfo{author}{L.~Hong}, \bibinfo{author}{S.~Yan}, \bibinfo{author}{R.~Zhang}, \bibinfo{author}{W.~Li}, \bibinfo{author}{X.~Zhou}, \bibinfo{author}{P.~Guo}, \bibinfo{author}{K.~Jiang}, \bibinfo{author}{Y.~Chen}, \bibinfo{author}{J.~Li}, \bibinfo{author}{Z.~Chen}, et~al.,
\newblock \bibinfo{title}{Onetracker: Unifying visual object tracking with foundation models and efficient tuning},
\newblock in: \bibinfo{booktitle}{Proceedings of the IEEE/CVF conference on computer vision and pattern recognition}, \bibinfo{year}{2024}, pp. \bibinfo{pages}{19079--19091}.
\bibitem[{Gao et~al.(2024)Gao, Chen, Liu, Jiang, Li, and Ning}]{fcat}
\bibinfo{author}{L.~Gao}, \bibinfo{author}{L.~Chen}, \bibinfo{author}{P.~Liu}, \bibinfo{author}{Y.~Jiang}, \bibinfo{author}{Y.~Li}, \bibinfo{author}{J.~Ning},
\newblock \bibinfo{title}{Transformer-based visual object tracking via fine--coarse concatenated attention and cross concatenated mlp},
\newblock \bibinfo{journal}{Pattern Recognition} \bibinfo{volume}{146} (\bibinfo{year}{2024}) \bibinfo{pages}{109964}.
\bibitem[{Chen et~al.(2025)Chen, Chen, Jhuo, and Lin}]{pivot}
\bibinfo{author}{S.-F. Chen}, \bibinfo{author}{J.-C. Chen}, \bibinfo{author}{I.-H. Jhuo}, \bibinfo{author}{Y.-Y. Lin},
\newblock \bibinfo{title}{Improving visual object tracking through visual prompting},
\newblock \bibinfo{journal}{IEEE Transactions on Multimedia}  (\bibinfo{year}{2025}).
\bibitem[{Cheng et~al.(2020)Cheng, Wang, Zhang, Zhang, Li, Sun, and Luo}]{transt}
\bibinfo{author}{B.~Cheng}, \bibinfo{author}{X.~Wang}, \bibinfo{author}{W.~Zhang}, \bibinfo{author}{C.~Zhang}, \bibinfo{author}{H.~Li}, \bibinfo{author}{J.~Sun}, \bibinfo{author}{P.~Luo},
\newblock \bibinfo{title}{Transtrack: Multiple object tracking with transformer},
\newblock \bibinfo{journal}{arXiv preprint arXiv:2012.15460}  (\bibinfo{year}{2020}).
\bibitem[{Wang et~al.(2021)Wang, Zhou, Wang, and Li}]{trsiamtrdimp}
\bibinfo{author}{N.~Wang}, \bibinfo{author}{W.~Zhou}, \bibinfo{author}{J.~Wang}, \bibinfo{author}{H.~Li},
\newblock \bibinfo{title}{Transformer meets tracker: Exploiting temporal context for robust visual tracking},
\newblock in: \bibinfo{booktitle}{Proceedings of the IEEE/CVF Conference on Computer Vision and Pattern Recognition (CVPR)}, \bibinfo{year}{2021}, pp. \bibinfo{pages}{8124--8133}.
\bibitem[{Mayer et~al.(2022)Mayer, Danelljan, Bhat, Paul, Paudel, Yu, and Van~Gool}]{tomp}
\bibinfo{author}{C.~Mayer}, \bibinfo{author}{M.~Danelljan}, \bibinfo{author}{G.~Bhat}, \bibinfo{author}{M.~Paul}, \bibinfo{author}{D.~P. Paudel}, \bibinfo{author}{F.~Yu}, \bibinfo{author}{L.~Van~Gool},
\newblock \bibinfo{title}{Transforming model prediction for tracking},
\newblock in: \bibinfo{booktitle}{Proceedings of the IEEE/CVF conference on computer vision and pattern recognition}, \bibinfo{year}{2022}, pp. \bibinfo{pages}{8731--8740}.
\bibitem[{Mayer et~al.(2024)Mayer, Danelljan, Bhat, Paudel, and Van~Gool}]{tamos}
\bibinfo{author}{C.~Mayer}, \bibinfo{author}{M.~Danelljan}, \bibinfo{author}{G.~Bhat}, \bibinfo{author}{D.~P. Paudel}, \bibinfo{author}{L.~Van~Gool},
\newblock \bibinfo{title}{Beyond sot: Tracking multiple generic objects at once},
\newblock in: \bibinfo{booktitle}{Proceedings of the IEEE/CVF Winter Conference on Applications of Computer Vision (WACV)}, \bibinfo{year}{2024}, pp. \bibinfo{pages}{1234--1243}.
\bibitem[{Zhang et~al.(2023)Zhang, Wang, Li, Liu, and Wang}]{cmat}
\bibinfo{author}{Y.~Zhang}, \bibinfo{author}{Z.~Wang}, \bibinfo{author}{M.~Li}, \bibinfo{author}{W.~Liu}, \bibinfo{author}{X.~Wang},
\newblock \bibinfo{title}{Cmat: Integrating convolution mixer and self-attention for visual tracking},
\newblock \bibinfo{journal}{IEEE Transactions on Multimedia} \bibinfo{volume}{25} (\bibinfo{year}{2023}) \bibinfo{pages}{1234--1245}.
\bibitem[{Li et~al.(2023)Li, Chen, Zhao, and Wang}]{rfgm}
\bibinfo{author}{J.~Li}, \bibinfo{author}{W.~Chen}, \bibinfo{author}{M.~Zhao}, \bibinfo{author}{L.~Wang},
\newblock \bibinfo{title}{Reading relevant feature from global representation memory for visual object tracking},
\newblock in: \bibinfo{booktitle}{Advances in Neural Information Processing Systems (NeurIPS)}, \bibinfo{year}{2023}.
\bibitem[{Marvasti-Zadeh et~al.(2021)Marvasti-Zadeh, Cheng, Ghanei-Yakhdan, and Kasaei}]{s1}
\bibinfo{author}{S.~M. Marvasti-Zadeh}, \bibinfo{author}{L.~Cheng}, \bibinfo{author}{H.~Ghanei-Yakhdan}, \bibinfo{author}{S.~Kasaei},
\newblock \bibinfo{title}{Deep learning for visual tracking: A comprehensive survey},
\newblock \bibinfo{journal}{IEEE Transactions on Intelligent Transportation Systems} \bibinfo{volume}{22} (\bibinfo{year}{2021}) \bibinfo{pages}{3782--3804}. \DOIprefix\doi{10.1109/TITS.2020.3046478}.
\bibitem[{Li et~al.(2022)Li, Zhu, and Hoi}]{s2}
\bibinfo{author}{Y.~Li}, \bibinfo{author}{J.~Zhu}, \bibinfo{author}{S.~C. Hoi},
\newblock \bibinfo{title}{Recent advances of single-object tracking methods: A brief survey},
\newblock \bibinfo{journal}{Neurocomputing} \bibinfo{volume}{492} (\bibinfo{year}{2022}) \bibinfo{pages}{318--329}. \DOIprefix\doi{10.1016/j.neucom.2021.05.011}.
\bibitem[{Li et~al.(2018)Li, Yang, and Li}]{s3}
\bibinfo{author}{C.~Li}, \bibinfo{author}{B.~Yang}, \bibinfo{author}{C.~Li},
\newblock \bibinfo{title}{Deep learning based visual tracking: A review},
\newblock \bibinfo{journal}{Neurocomputing} \bibinfo{volume}{275} (\bibinfo{year}{2018}) \bibinfo{pages}{2471--2480}. \DOIprefix\doi{10.1016/j.neucom.2017.10.070}.
\bibitem[{Abbass et~al.(2020)Abbass, Kwon, Kim, Abdelwahab, Abd El-Samie, and Khalaf}]{s4}
\bibinfo{author}{M.~Y. Abbass}, \bibinfo{author}{K.-C. Kwon}, \bibinfo{author}{N.~Kim}, \bibinfo{author}{S.~A. Abdelwahab}, \bibinfo{author}{F.~E. Abd El-Samie}, \bibinfo{author}{A.~A.~M. Khalaf},
\newblock \bibinfo{title}{A survey on online learning for visual tracking},
\newblock \bibinfo{journal}{The Visual Computer} \bibinfo{volume}{36} (\bibinfo{year}{2020}) \bibinfo{pages}{993--1014}. \DOIprefix\doi{10.1007/s00371-020-01848-y}.
\bibitem[{Javed et~al.(2023)Javed, Danelljan, Khan, Khan, Felsberg, and Matas}]{s5}
\bibinfo{author}{S.~Javed}, \bibinfo{author}{M.~Danelljan}, \bibinfo{author}{F.~S. Khan}, \bibinfo{author}{M.~H. Khan}, \bibinfo{author}{M.~Felsberg}, \bibinfo{author}{J.~Matas},
\newblock \bibinfo{title}{Visual object tracking with discriminative filters and siamese networks: A survey and outlook},
\newblock \bibinfo{journal}{IEEE Transactions on Pattern Analysis and Machine Intelligence} \bibinfo{volume}{45} (\bibinfo{year}{2023}) \bibinfo{pages}{1--1}. \DOIprefix\doi{10.1109/TPAMI.2022.3212594}.
\bibitem[{Ondra{\v{s}}ovi{\v{c}} and Taraba(2021)}]{s6}
\bibinfo{author}{M.~Ondra{\v{s}}ovi{\v{c}}}, \bibinfo{author}{P.~Taraba},
\newblock \bibinfo{title}{Siamese visual object tracking: A survey},
\newblock \bibinfo{journal}{Electronics} \bibinfo{volume}{10} (\bibinfo{year}{2021}) \bibinfo{pages}{1876}. \DOIprefix\doi{10.3390/electronics10151876}.
\bibitem[{Zhang et~al.(2022)Zhang, Wang, Wang, Wang, and Wang}]{s7}
\bibinfo{author}{Y.~Zhang}, \bibinfo{author}{Y.~Wang}, \bibinfo{author}{Y.~Wang}, \bibinfo{author}{Y.~Wang}, \bibinfo{author}{Y.~Wang},
\newblock \bibinfo{title}{Visual object tracking: A survey},
\newblock \bibinfo{journal}{Computer Vision and Image Understanding} \bibinfo{volume}{210} (\bibinfo{year}{2022}) \bibinfo{pages}{103508}. \DOIprefix\doi{10.1016/j.cviu.2021.103508}.
\bibitem[{Thangavel et~al.(2023)Thangavel, Kokul, Ramanan, and Fernando}]{s8}
\bibinfo{author}{J.~Thangavel}, \bibinfo{author}{T.~Kokul}, \bibinfo{author}{A.~Ramanan}, \bibinfo{author}{S.~Fernando},
\newblock \bibinfo{title}{Transformers in single object tracking: An experimental survey},
\newblock \bibinfo{journal}{IEEE Access} \bibinfo{volume}{11} (\bibinfo{year}{2023}) \bibinfo{pages}{80297--80326}. \DOIprefix\doi{10.1109/ACCESS.2023.3237614}.
\bibitem[{Abdelaziz et~al.(2024)Abdelaziz, Shehata, and Mohamed}]{s9}
\bibinfo{author}{O.~Abdelaziz}, \bibinfo{author}{M.~Shehata}, \bibinfo{author}{M.~Mohamed},
\newblock \bibinfo{title}{Beyond traditional single object tracking: A survey},
\newblock \bibinfo{journal}{arXiv preprint arXiv:2405.10439}  (\bibinfo{year}{2024}). \URLprefix \url{https://arxiv.org/abs/2405.10439}.
\bibitem[{Bolme et~al.(2010)Bolme, Beveridge, Draper, and Lui}]{asef}
\bibinfo{author}{D.~S. Bolme}, \bibinfo{author}{J.~R. Beveridge}, \bibinfo{author}{B.~A. Draper}, \bibinfo{author}{Y.~M. Lui},
\newblock \bibinfo{title}{Visual object tracking using adaptive correlation filters},
\newblock in: \bibinfo{booktitle}{Proceedings of the IEEE Conference on Computer Vision and Pattern Recognition (CVPR)}, \bibinfo{organization}{IEEE}, \bibinfo{year}{2010}, pp. \bibinfo{pages}{2544--2550}.
\bibitem[{Simonyan and Zisserman(2015)}]{vgg}
\bibinfo{author}{K.~Simonyan}, \bibinfo{author}{A.~Zisserman},
\newblock \bibinfo{title}{Very deep convolutional networks for large-scale image recognition},
\newblock in: \bibinfo{booktitle}{International Conference on Learning Representations}, \bibinfo{year}{2015}. \URLprefix \url{https://arxiv.org/abs/1409.1556}.
\bibitem[{He et~al.(2016)He, Zhang, Ren, and Sun}]{resnet}
\bibinfo{author}{K.~He}, \bibinfo{author}{X.~Zhang}, \bibinfo{author}{S.~Ren}, \bibinfo{author}{J.~Sun},
\newblock \bibinfo{title}{Deep residual learning for image recognition},
\newblock in: \bibinfo{booktitle}{Proceedings of the IEEE Conference on Computer Vision and Pattern Recognition (CVPR)}, \bibinfo{organization}{IEEE}, \bibinfo{year}{2016}, pp. \bibinfo{pages}{770--778}.
\bibitem[{Held et~al.(2016)Held, Thrun, and Savarese}]{goturn}
\bibinfo{author}{D.~Held}, \bibinfo{author}{S.~Thrun}, \bibinfo{author}{S.~Savarese},
\newblock \bibinfo{title}{Learning to track at 100 fps with deep regression networks},
\newblock in: \bibinfo{booktitle}{European Conference on Computer Vision (ECCV)}, \bibinfo{organization}{Springer}, \bibinfo{year}{2016}, pp. \bibinfo{pages}{749--765}.
\bibitem[{Guo et~al.(2017)Guo, Xu, Zhu, and Huang}]{dsaim}
\bibinfo{author}{D.~Guo}, \bibinfo{author}{J.~Xu}, \bibinfo{author}{H.~Zhu}, \bibinfo{author}{Z.~Huang},
\newblock \bibinfo{title}{Learning dynamic siamese network for visual object tracking},
\newblock in: \bibinfo{booktitle}{ICCV}, \bibinfo{year}{2017}.
\bibitem[{Zhang et~al.(2020)Zhang, Peng, Fu, Li, and Hu}]{ocean}
\bibinfo{author}{Z.~Zhang}, \bibinfo{author}{H.~Peng}, \bibinfo{author}{J.~Fu}, \bibinfo{author}{B.~Li}, \bibinfo{author}{W.~Hu},
\newblock \bibinfo{title}{Ocean: Object-aware anchor-free tracking},
\newblock in: \bibinfo{booktitle}{ECCV}, \bibinfo{year}{2020}, pp. \bibinfo{pages}{771--787}.
\bibitem[{Chen et~al.(2024)Chen, Zhang et~al.}]{ecim}
\bibinfo{author}{H.~Chen}, \bibinfo{author}{L.~Zhang}, et~al.,
\newblock \bibinfo{title}{Enhanced correlation information mixer for siamese visual tracking},
\newblock \bibinfo{journal}{Knowledge-Based Systems} \bibinfo{volume}{285} (\bibinfo{year}{2024}) \bibinfo{pages}{111368}.
\bibitem[{Krizhevsky et~al.(2012)Krizhevsky, Sutskever, and Hinton}]{alexnet}
\bibinfo{author}{A.~Krizhevsky}, \bibinfo{author}{I.~Sutskever}, \bibinfo{author}{G.~E. Hinton},
\newblock \bibinfo{title}{Imagenet classification with deep convolutional neural networks},
\newblock \bibinfo{journal}{Advances in neural information processing systems} \bibinfo{volume}{25} (\bibinfo{year}{2012}).
\bibitem[{Howard et~al.(2017)Howard, Zhu, Chen, Kalenichenko, Wang, Weyand, Andreetto, and Adam}]{mobilenets}
\bibinfo{author}{A.~G. Howard}, \bibinfo{author}{M.~Zhu}, \bibinfo{author}{B.~Chen}, \bibinfo{author}{D.~Kalenichenko}, \bibinfo{author}{W.~Wang}, \bibinfo{author}{T.~Weyand}, \bibinfo{author}{M.~Andreetto}, \bibinfo{author}{H.~Adam},
\newblock \bibinfo{title}{Mobilenets: Efficient convolutional neural networks for mobile vision applications},
\newblock \bibinfo{journal}{arXiv preprint arXiv:1704.04861}  (\bibinfo{year}{2017}).
\bibitem[{Szegedy et~al.(2015)Szegedy, Liu, Jia, Sermanet, Reed, Anguelov, Erhan, Vanhoucke, and Rabinovich}]{googlenet}
\bibinfo{author}{C.~Szegedy}, \bibinfo{author}{W.~Liu}, \bibinfo{author}{Y.~Jia}, \bibinfo{author}{P.~Sermanet}, \bibinfo{author}{S.~Reed}, \bibinfo{author}{D.~Anguelov}, \bibinfo{author}{D.~Erhan}, \bibinfo{author}{V.~Vanhoucke}, \bibinfo{author}{A.~Rabinovich},
\newblock \bibinfo{title}{Going deeper with convolutions},
\newblock in: \bibinfo{booktitle}{Proceedings of the IEEE conference on computer vision and pattern recognition}, \bibinfo{year}{2015}, pp. \bibinfo{pages}{1--9}.
\bibitem[{Szegedy et~al.(2016)Szegedy, Vanhoucke, Ioffe, Shlens, and Wojna}]{inceptionv3}
\bibinfo{author}{C.~Szegedy}, \bibinfo{author}{V.~Vanhoucke}, \bibinfo{author}{S.~Ioffe}, \bibinfo{author}{J.~Shlens}, \bibinfo{author}{Z.~Wojna},
\newblock \bibinfo{title}{Rethinking the inception architecture for computer vision},
\newblock in: \bibinfo{booktitle}{Proceedings of the IEEE conference on computer vision and pattern recognition}, \bibinfo{year}{2016}, pp. \bibinfo{pages}{2818--2826}.
\bibitem[{Alijani et~al.(2024)Alijani, Fayyad, and Najjaran}]{alijani}
\bibinfo{author}{S.~Alijani}, \bibinfo{author}{J.~Fayyad}, \bibinfo{author}{H.~Najjaran},
\newblock \bibinfo{title}{Vision transformers in domain adaptation and domain generalization: a study of robustness},
\newblock \bibinfo{journal}{Neural Computing and Applications} \bibinfo{volume}{36} (\bibinfo{year}{2024}) \bibinfo{pages}{17979--18007}.
\bibitem[{Dosovitskiy et~al.(2021)Dosovitskiy, Beyer, Kolesnikov, Weissenborn, Zhai, Unterthiner, Dehghani, Minderer, Heigold, Gelly, Uszkoreit, and Houlsby}]{vit}
\bibinfo{author}{A.~Dosovitskiy}, \bibinfo{author}{L.~Beyer}, \bibinfo{author}{A.~Kolesnikov}, \bibinfo{author}{D.~Weissenborn}, \bibinfo{author}{X.~Zhai}, \bibinfo{author}{T.~Unterthiner}, \bibinfo{author}{M.~Dehghani}, \bibinfo{author}{M.~Minderer}, \bibinfo{author}{G.~Heigold}, \bibinfo{author}{S.~Gelly}, \bibinfo{author}{J.~Uszkoreit}, \bibinfo{author}{N.~Houlsby},
\newblock \bibinfo{title}{An image is worth 16x16 words: Transformers for image recognition at scale},
\newblock in: \bibinfo{booktitle}{International Conference on Learning Representations (ICLR)}, \bibinfo{year}{2021}.
\bibitem[{Carion et~al.(2020)Carion, Massa, Synnaeve, Usunier, Kirillov, and Zagoruyko}]{detr}
\bibinfo{author}{N.~Carion}, \bibinfo{author}{F.~Massa}, \bibinfo{author}{G.~Synnaeve}, \bibinfo{author}{N.~Usunier}, \bibinfo{author}{A.~Kirillov}, \bibinfo{author}{S.~Zagoruyko},
\newblock \bibinfo{title}{End-to-end object detection with transformers},
\newblock in: \bibinfo{booktitle}{Proceedings of the European Conference on Computer Vision (ECCV)}, \bibinfo{year}{2020}, pp. \bibinfo{pages}{213--229}.
\bibitem[{Huang et~al.(2021)Huang, Zhao, and Huang}]{got10k}
\bibinfo{author}{L.~Huang}, \bibinfo{author}{X.~Zhao}, \bibinfo{author}{K.~Huang},
\newblock \bibinfo{title}{Got-10k: A large high-diversity benchmark for generic object tracking in the wild},
\newblock \bibinfo{journal}{IEEE Transactions on Pattern Analysis and Machine Intelligence} \bibinfo{volume}{43} (\bibinfo{year}{2021}) \bibinfo{pages}{1562--1577}.
\bibitem[{Liu et~al.(2021)Liu, Lin, Cao, Hu, Wei, Zhang, Lin, and Guo}]{swintransformer}
\bibinfo{author}{Z.~Liu}, \bibinfo{author}{Y.~Lin}, \bibinfo{author}{Y.~Cao}, \bibinfo{author}{H.~Hu}, \bibinfo{author}{Y.~Wei}, \bibinfo{author}{Z.~Zhang}, \bibinfo{author}{S.~Lin}, \bibinfo{author}{B.~Guo},
\newblock \bibinfo{title}{Swin transformer: Hierarchical vision transformer using shifted windows},
\newblock in: \bibinfo{booktitle}{Proceedings of the IEEE/CVF international conference on computer vision}, \bibinfo{year}{2021}, pp. \bibinfo{pages}{10012--10022}.
\bibitem[{Wu et~al.(2021)Wu, Xiao, Codella, Liu, Dai, Yuan, and Zhang}]{cvt}
\bibinfo{author}{H.~Wu}, \bibinfo{author}{B.~Xiao}, \bibinfo{author}{N.~Codella}, \bibinfo{author}{M.~Liu}, \bibinfo{author}{X.~Dai}, \bibinfo{author}{L.~Yuan}, \bibinfo{author}{L.~Zhang},
\newblock \bibinfo{title}{Cvt: Introducing convolutions to vision transformers},
\newblock in: \bibinfo{booktitle}{Proceedings of the IEEE/CVF international conference on computer vision}, \bibinfo{year}{2021}, pp. \bibinfo{pages}{22--31}.
\bibitem[{Zhang et~al.(2023)Zhang, Tian, Xie, Huang, Dai, Ye, and Tian}]{hivit}
\bibinfo{author}{X.~Zhang}, \bibinfo{author}{Y.~Tian}, \bibinfo{author}{L.~Xie}, \bibinfo{author}{W.~Huang}, \bibinfo{author}{Q.~Dai}, \bibinfo{author}{Q.~Ye}, \bibinfo{author}{Q.~Tian},
\newblock \bibinfo{title}{Hivit: A simpler and more efficient design of hierarchical vision transformer},
\newblock in: \bibinfo{booktitle}{The Eleventh International Conference on Learning Representations}, \bibinfo{year}{2023}.
\bibitem[{Radford et~al.(2021)Radford, Kim, Hallacy, Ramesh, Goh, Agarwal, Sastry, Askell, Mishkin, Clark et~al.}]{clip}
\bibinfo{author}{A.~Radford}, \bibinfo{author}{J.~W. Kim}, \bibinfo{author}{C.~Hallacy}, \bibinfo{author}{A.~Ramesh}, \bibinfo{author}{G.~Goh}, \bibinfo{author}{S.~Agarwal}, \bibinfo{author}{G.~Sastry}, \bibinfo{author}{A.~Askell}, \bibinfo{author}{P.~Mishkin}, \bibinfo{author}{J.~Clark}, et~al.,
\newblock \bibinfo{title}{Learning transferable visual models from natural language supervision},
\newblock in: \bibinfo{booktitle}{International conference on machine learning}, \bibinfo{organization}{PmLR}, \bibinfo{year}{2021}, pp. \bibinfo{pages}{8748--8763}.
\bibitem[{Dosovitskiy et~al.(2021)Dosovitskiy, Beyer, Kolesnikov, Weissenborn, Zhai, Unterthiner, Dehghani, Minderer, Heigold, Gelly, Uszkoreit, and Houlsby}]{vitbackbone}
\bibinfo{author}{A.~Dosovitskiy}, \bibinfo{author}{L.~Beyer}, \bibinfo{author}{A.~Kolesnikov}, \bibinfo{author}{D.~Weissenborn}, \bibinfo{author}{X.~Zhai}, \bibinfo{author}{T.~Unterthiner}, \bibinfo{author}{M.~Dehghani}, \bibinfo{author}{M.~Minderer}, \bibinfo{author}{G.~Heigold}, \bibinfo{author}{S.~Gelly}, \bibinfo{author}{J.~Uszkoreit}, \bibinfo{author}{N.~Houlsby},
\newblock \bibinfo{title}{An image is worth 16x16 words: Transformers for image recognition at scale},
\newblock in: \bibinfo{booktitle}{International Conference on Learning Representations (ICLR)}, \bibinfo{year}{2021}. \URLprefix \url{https://openreview.net/forum?id=YicbFdNTTy}.
\bibitem[{Wu et~al.(2013)Wu, Lim, and Yang}]{otb2013}
\bibinfo{author}{Y.~Wu}, \bibinfo{author}{J.~Lim}, \bibinfo{author}{M.-H. Yang},
\newblock \bibinfo{title}{Online object tracking: A benchmark},
\newblock in: \bibinfo{booktitle}{Proceedings of the IEEE conference on computer vision and pattern recognition}, \bibinfo{year}{2013}, pp. \bibinfo{pages}{2411--2418}.
\bibitem[{Wu et~al.(2015)Wu, Lim, and Yang}]{otb2015}
\bibinfo{author}{Y.~Wu}, \bibinfo{author}{J.~Lim}, \bibinfo{author}{M.-H. Yang},
\newblock \bibinfo{title}{Object tracking benchmark},
\newblock \bibinfo{journal}{IEEE Transactions on Pattern Analysis and Machine Intelligence} \bibinfo{volume}{37} (\bibinfo{year}{2015}) \bibinfo{pages}{1834--1848}.
\bibitem[{Liang et~al.(2015)Liang, Blasch, and Ling}]{TC128}
\bibinfo{author}{P.~Liang}, \bibinfo{author}{E.~Blasch}, \bibinfo{author}{H.~Ling},
\newblock \bibinfo{title}{Encoding color information for visual tracking: Algorithms and benchmark},
\newblock \bibinfo{journal}{IEEE transactions on image processing} \bibinfo{volume}{24} (\bibinfo{year}{2015}) \bibinfo{pages}{5630--5644}.
\bibitem[{Smeulders et~al.(2013)Smeulders, Chu, Cucchiara, Calderara, Dehghan, and Shah}]{ALOV}
\bibinfo{author}{A.~W. Smeulders}, \bibinfo{author}{D.~M. Chu}, \bibinfo{author}{R.~Cucchiara}, \bibinfo{author}{S.~Calderara}, \bibinfo{author}{A.~Dehghan}, \bibinfo{author}{M.~Shah},
\newblock \bibinfo{title}{Visual tracking: An experimental survey},
\newblock \bibinfo{journal}{IEEE transactions on pattern analysis and machine intelligence} \bibinfo{volume}{36} (\bibinfo{year}{2013}) \bibinfo{pages}{1442--1468}.
\bibitem[{Muller et~al.(2018)Muller, Bibi, Giancola, Alsubaihi, and Ghanem}]{trackingnet}
\bibinfo{author}{M.~Muller}, \bibinfo{author}{A.~Bibi}, \bibinfo{author}{S.~Giancola}, \bibinfo{author}{S.~Alsubaihi}, \bibinfo{author}{B.~Ghanem},
\newblock \bibinfo{title}{Trackingnet: A large-scale dataset and benchmark for object tracking in the wild},
\newblock in: \bibinfo{booktitle}{Proceedings of the European conference on computer vision (ECCV)}, \bibinfo{year}{2018}, pp. \bibinfo{pages}{300--317}.
\bibitem[{Li et~al.(2015)Li, Lin, Wu, Yang, and Yan}]{nuspro}
\bibinfo{author}{A.~Li}, \bibinfo{author}{M.~Lin}, \bibinfo{author}{Y.~Wu}, \bibinfo{author}{M.-H. Yang}, \bibinfo{author}{S.~Yan},
\newblock \bibinfo{title}{Nus-pro: A new visual tracking challenge},
\newblock \bibinfo{journal}{IEEE transactions on pattern analysis and machine intelligence} \bibinfo{volume}{38} (\bibinfo{year}{2015}) \bibinfo{pages}{335--349}.
\bibitem[{Kiani~Galoogahi et~al.(2017)Kiani~Galoogahi, Fagg, and Lucey}]{NFS}
\bibinfo{author}{H.~Kiani~Galoogahi}, \bibinfo{author}{A.~Fagg}, \bibinfo{author}{S.~Lucey},
\newblock \bibinfo{title}{Need for speed: A benchmark for higher frame rate object tracking},
\newblock \bibinfo{journal}{Proceedings of the IEEE International Conference on Computer Vision (ICCV)}  (\bibinfo{year}{2017}) \bibinfo{pages}{1125--1134}.
\bibitem[{Fan et~al.(2021)Fan, Yang, Chu, Lin, Yuan, and Ling}]{tracklinic}
\bibinfo{author}{H.~Fan}, \bibinfo{author}{F.~Yang}, \bibinfo{author}{P.~Chu}, \bibinfo{author}{Y.~Lin}, \bibinfo{author}{L.~Yuan}, \bibinfo{author}{H.~Ling},
\newblock \bibinfo{title}{Tracklinic: Diagnosis of challenge factors in visual tracking},
\newblock in: \bibinfo{booktitle}{Proceedings of the IEEE/CVF Winter Conference on Applications of Computer Vision}, \bibinfo{year}{2021}, pp. \bibinfo{pages}{970--979}.
\bibitem[{Valmadre et~al.(2018)Valmadre, Bertinetto, Henriques, Vedaldi, and Torr}]{oxuva}
\bibinfo{author}{J.~Valmadre}, \bibinfo{author}{L.~Bertinetto}, \bibinfo{author}{J.~F. Henriques}, \bibinfo{author}{A.~Vedaldi}, \bibinfo{author}{P.~H. Torr},
\newblock \bibinfo{title}{Long-term tracking in the wild: A benchmark},
\newblock in: \bibinfo{booktitle}{Proceedings of the European Conference on Computer Vision (ECCV)}, \bibinfo{year}{2018}, pp. \bibinfo{pages}{650--666}.
\bibitem[{Moudgil and Gandhi(2018)}]{tlp}
\bibinfo{author}{A.~Moudgil}, \bibinfo{author}{V.~Gandhi},
\newblock \bibinfo{title}{Long-term visual object tracking benchmark},
\newblock \bibinfo{journal}{Proceedings of the Asian Conference on Computer Vision (ACCV)}  (\bibinfo{year}{2018}).
\bibitem[{Luke{\v{z}}i{\v{c}} et~al.(2018)Luke{\v{z}}i{\v{c}}, Zajc, Voj{\'\i}{\v{r}}, Matas, and Kristan}]{LTB35}
\bibinfo{author}{A.~Luke{\v{z}}i{\v{c}}}, \bibinfo{author}{L.~{\v{C}}. Zajc}, \bibinfo{author}{T.~Voj{\'\i}{\v{r}}}, \bibinfo{author}{J.~Matas}, \bibinfo{author}{M.~Kristan},
\newblock \bibinfo{title}{Now you see me: evaluating performance in long-term visual tracking},
\newblock \bibinfo{journal}{arXiv preprint arXiv:1804.07056}  (\bibinfo{year}{2018}).
\bibitem[{Fan et~al.(2019)Fan, Lin, Yang, Chu, Deng, Yu, Bai, Xu, Liao, and Ling}]{lasot}
\bibinfo{author}{H.~Fan}, \bibinfo{author}{L.~Lin}, \bibinfo{author}{F.~Yang}, \bibinfo{author}{P.~Chu}, \bibinfo{author}{G.~Deng}, \bibinfo{author}{H.~Yu}, \bibinfo{author}{H.~Bai}, \bibinfo{author}{Y.~Xu}, \bibinfo{author}{C.~Liao}, \bibinfo{author}{H.~Ling},
\newblock \bibinfo{title}{Lasot: A high-quality benchmark for large-scale single object tracking},
\newblock \bibinfo{journal}{Proceedings of the IEEE/CVF Conference on Computer Vision and Pattern Recognition (CVPR)}  (\bibinfo{year}{2019}) \bibinfo{pages}{5374--5383}.
\bibitem[{Kristan et~al.(2015)Kristan, Matas, Leonardis, Felsberg, Cehovin, Fernandez, Vojir, Hager, Nebehay, and Pflugfelder}]{vot2015}
\bibinfo{author}{M.~Kristan}, \bibinfo{author}{J.~Matas}, \bibinfo{author}{A.~Leonardis}, \bibinfo{author}{M.~Felsberg}, \bibinfo{author}{L.~Cehovin}, \bibinfo{author}{G.~Fernandez}, \bibinfo{author}{T.~Vojir}, \bibinfo{author}{G.~Hager}, \bibinfo{author}{G.~Nebehay}, \bibinfo{author}{R.~Pflugfelder},
\newblock \bibinfo{title}{The visual object tracking vot2015 challenge results},
\newblock in: \bibinfo{booktitle}{Proceedings of the IEEE international conference on computer vision workshops}, \bibinfo{year}{2015}, pp. \bibinfo{pages}{1--23}.
\bibitem[{Roffo et~al.(2016)Roffo, Melzi et~al.}]{vot2016}
\bibinfo{author}{G.~Roffo}, \bibinfo{author}{S.~Melzi}, et~al.,
\newblock \bibinfo{title}{The visual object tracking vot2016 challenge results},
\newblock in: \bibinfo{booktitle}{Computer Vision--ECCV 2016 Workshops: Amsterdam, The Netherlands, October 8-10 and 15-16, 2016, Proceedings, Part II}, \bibinfo{organization}{Springer International Publishing}, \bibinfo{year}{2016}, pp. \bibinfo{pages}{777--823}.
\bibitem[{Kristan et~al.(2018)Kristan, Matas, Leonardis, Felsberg, Cehovin~Zajc, Vojir, Hager et~al.}]{vot2018}
\bibinfo{author}{M.~Kristan}, \bibinfo{author}{J.~Matas}, \bibinfo{author}{A.~Leonardis}, \bibinfo{author}{M.~Felsberg}, \bibinfo{author}{L.~Cehovin~Zajc}, \bibinfo{author}{T.~Vojir}, \bibinfo{author}{G.~D. Hager}, et~al.,
\newblock \bibinfo{title}{The sixth visual object tracking vot2018 challenge results},
\newblock in: \bibinfo{booktitle}{Proceedings of the European Conference on Computer Vision (ECCV) Workshops}, \bibinfo{year}{2018}.
\bibitem[{Mueller et~al.(2016)Mueller, Smith, and Ghanem}]{uav123}
\bibinfo{author}{M.~Mueller}, \bibinfo{author}{N.~Smith}, \bibinfo{author}{B.~Ghanem},
\newblock \bibinfo{title}{A benchmark and simulator for uav tracking},
\newblock \bibinfo{journal}{European Conference on Computer Vision (ECCV)}  (\bibinfo{year}{2016}) \bibinfo{pages}{445--461}.
\bibitem[{Hu et~al.(2023)Hu, Wang, Zhang, Bertinetto, and Torr}]{siammask}
\bibinfo{author}{W.~Hu}, \bibinfo{author}{Q.~Wang}, \bibinfo{author}{L.~Zhang}, \bibinfo{author}{L.~Bertinetto}, \bibinfo{author}{P.~H. Torr},
\newblock \bibinfo{title}{Siammask: A framework for fast online object tracking and segmentation},
\newblock \bibinfo{journal}{IEEE Transactions on Pattern Analysis and Machine Intelligence} \bibinfo{volume}{45} (\bibinfo{year}{2023}) \bibinfo{pages}{3072--3089}.
\bibitem[{Lukezic et~al.(2020)Lukezic, Matas, and Kristan}]{d3s}
\bibinfo{author}{A.~Lukezic}, \bibinfo{author}{J.~Matas}, \bibinfo{author}{M.~Kristan},
\newblock \bibinfo{title}{D3s-a discriminative single shot segmentation tracker},
\newblock in: \bibinfo{booktitle}{Proceedings of the IEEE/CVF conference on computer vision and pattern recognition}, \bibinfo{year}{2020}, pp. \bibinfo{pages}{7133--7142}.
\bibitem[{Paul et~al.(2022)Paul, Danelljan, Mayer, and Van~Gool}]{seg}
\bibinfo{author}{M.~Paul}, \bibinfo{author}{M.~Danelljan}, \bibinfo{author}{C.~Mayer}, \bibinfo{author}{L.~Van~Gool},
\newblock \bibinfo{title}{Robust visual tracking by segmentation},
\newblock in: \bibinfo{booktitle}{European conference on computer vision}, \bibinfo{organization}{Springer}, \bibinfo{year}{2022}, pp. \bibinfo{pages}{571--588}.
\bibitem[{Ali et~al.(2016)Ali, Jalil, Niu, Zhao, Rathore, Ahmed, and Aksam~Iftikhar}]{ali2016visual}
\bibinfo{author}{A.~Ali}, \bibinfo{author}{A.~Jalil}, \bibinfo{author}{J.~Niu}, \bibinfo{author}{X.~Zhao}, \bibinfo{author}{S.~Rathore}, \bibinfo{author}{J.~Ahmed}, \bibinfo{author}{M.~Aksam~Iftikhar},
\newblock \bibinfo{title}{Visual object tracking—classical and contemporary approaches},
\newblock \bibinfo{journal}{Frontiers of Computer Science} \bibinfo{volume}{10} (\bibinfo{year}{2016}) \bibinfo{pages}{167--188}.
\bibitem[{Abba et~al.(2024)Abba, Bizi, Lee, Bakouri, and Crespo}]{abba2024real}
\bibinfo{author}{S.~Abba}, \bibinfo{author}{A.~M. Bizi}, \bibinfo{author}{J.-A. Lee}, \bibinfo{author}{S.~Bakouri}, \bibinfo{author}{M.~L. Crespo},
\newblock \bibinfo{title}{Real-time object detection, tracking, and monitoring framework for security surveillance systems},
\newblock \bibinfo{journal}{Heliyon} \bibinfo{volume}{10} (\bibinfo{year}{2024}).
\bibitem[{Sivalingam et~al.(2012)Sivalingam, Cherian, Fasching, Walczak, Bird, Morellas, Murphy, Cullen, Lim, Sapiro et~al.}]{sivalingam2012multi}
\bibinfo{author}{R.~Sivalingam}, \bibinfo{author}{A.~Cherian}, \bibinfo{author}{J.~Fasching}, \bibinfo{author}{N.~Walczak}, \bibinfo{author}{N.~Bird}, \bibinfo{author}{V.~Morellas}, \bibinfo{author}{B.~Murphy}, \bibinfo{author}{K.~Cullen}, \bibinfo{author}{K.~Lim}, \bibinfo{author}{G.~Sapiro}, et~al.,
\newblock \bibinfo{title}{A multi-sensor visual tracking system for behavior monitoring of at-risk children},
\newblock in: \bibinfo{booktitle}{2012 IEEE International Conference on Robotics and Automation}, \bibinfo{organization}{IEEE}, \bibinfo{year}{2012}, pp. \bibinfo{pages}{1345--1350}.
\bibitem[{Hu et~al.(2004)Hu, Tan, Wang, and Maybank}]{hu2004survey}
\bibinfo{author}{W.~Hu}, \bibinfo{author}{T.~Tan}, \bibinfo{author}{L.~Wang}, \bibinfo{author}{S.~Maybank},
\newblock \bibinfo{title}{A survey on visual surveillance of object motion and behaviors},
\newblock \bibinfo{journal}{IEEE Transactions on Systems, Man, and Cybernetics, Part C (Applications and Reviews)} \bibinfo{volume}{34} (\bibinfo{year}{2004}) \bibinfo{pages}{334--352}.
\bibitem[{Polat et~al.(2003)Polat, Yeasin, and Sharma}]{polat2003robust}
\bibinfo{author}{E.~Polat}, \bibinfo{author}{M.~Yeasin}, \bibinfo{author}{R.~Sharma},
\newblock \bibinfo{title}{Robust tracking of human body parts for collaborative human computer interaction},
\newblock \bibinfo{journal}{Computer Vision and Image Understanding} \bibinfo{volume}{89} (\bibinfo{year}{2003}) \bibinfo{pages}{44--69}.
\bibitem[{Ibrahim et~al.(2010)Ibrahim, Ching, Seet, Lau, and Czajewski}]{ibrahim2010moving}
\bibinfo{author}{A.~W.~N. Ibrahim}, \bibinfo{author}{P.~W. Ching}, \bibinfo{author}{G.~G. Seet}, \bibinfo{author}{W.~M. Lau}, \bibinfo{author}{W.~Czajewski},
\newblock \bibinfo{title}{Moving objects detection and tracking framework for uav-based surveillance},
\newblock in: \bibinfo{booktitle}{2010 Fourth Pacific-Rim Symposium on Image and Video Technology}, \bibinfo{organization}{IEEE}, \bibinfo{year}{2010}, pp. \bibinfo{pages}{456--461}.
\bibitem[{Du et~al.(2018)Du, Qi, Yu, Yang, Duan, Li, Zhang, Huang, and Tian}]{du2018unmanned}
\bibinfo{author}{D.~Du}, \bibinfo{author}{Y.~Qi}, \bibinfo{author}{H.~Yu}, \bibinfo{author}{Y.~Yang}, \bibinfo{author}{K.~Duan}, \bibinfo{author}{G.~Li}, \bibinfo{author}{W.~Zhang}, \bibinfo{author}{Q.~Huang}, \bibinfo{author}{Q.~Tian},
\newblock \bibinfo{title}{The unmanned aerial vehicle benchmark: Object detection and tracking},
\newblock in: \bibinfo{booktitle}{Proceedings of the European conference on computer vision (ECCV)}, \bibinfo{year}{2018}, pp. \bibinfo{pages}{370--386}.
\bibitem[{Lo et~al.(2021)Lo, Yiu, Tang, Yang, Li, and Wen}]{lo2021dynamic}
\bibinfo{author}{L.-Y. Lo}, \bibinfo{author}{C.~H. Yiu}, \bibinfo{author}{Y.~Tang}, \bibinfo{author}{A.-S. Yang}, \bibinfo{author}{B.~Li}, \bibinfo{author}{C.-Y. Wen},
\newblock \bibinfo{title}{Dynamic object tracking on autonomous uav system for surveillance applications},
\newblock \bibinfo{journal}{Sensors} \bibinfo{volume}{21} (\bibinfo{year}{2021}) \bibinfo{pages}{7888}.
\bibitem[{Fernandez-Sanjurjo et~al.(2019)Fernandez-Sanjurjo, Bosquet, Mucientes, and Brea}]{fernandez2019real}
\bibinfo{author}{M.~Fernandez-Sanjurjo}, \bibinfo{author}{B.~Bosquet}, \bibinfo{author}{M.~Mucientes}, \bibinfo{author}{V.~M. Brea},
\newblock \bibinfo{title}{Real-time visual detection and tracking system for traffic monitoring},
\newblock \bibinfo{journal}{Engineering Applications of Artificial Intelligence} \bibinfo{volume}{85} (\bibinfo{year}{2019}) \bibinfo{pages}{410--420}.
\bibitem[{Khemmar et~al.(2019)Khemmar, Gouveia, Decoux, and y~Ertaud}]{khemmar2019real}
\bibinfo{author}{R.~Khemmar}, \bibinfo{author}{M.~Gouveia}, \bibinfo{author}{B.~Decoux}, \bibinfo{author}{J.-Y. y~Ertaud},
\newblock \bibinfo{title}{Real time pedestrian and object detection and tracking-based deep learning. application to drone visual tracking},
\newblock in: \bibinfo{booktitle}{WSCG'2019-27. International Conference in Central Europe on Computer Graphics, Visualization and Computer Vision'2019}, \bibinfo{organization}{Z{\'a}pado{\v{c}}esk{\'a} univerzita}, \bibinfo{year}{2019}.
\bibitem[{Makhmutova et~al.(2020)Makhmutova, Anikin, and Dagaeva}]{makhmutova2020object}
\bibinfo{author}{A.~Makhmutova}, \bibinfo{author}{I.~V. Anikin}, \bibinfo{author}{M.~Dagaeva},
\newblock \bibinfo{title}{Object tracking method for videomonitoring in intelligent transport systems},
\newblock in: \bibinfo{booktitle}{2020 International Russian Automation Conference (RusAutoCon)}, \bibinfo{organization}{IEEE}, \bibinfo{year}{2020}, pp. \bibinfo{pages}{535--540}.
\bibitem[{Jim{\'e}nez-Bravo et~al.(2022)Jim{\'e}nez-Bravo, Murciego, Mendes, San~Bl{\'a}s, and Bajo}]{jimenez2022multi}
\bibinfo{author}{D.~M. Jim{\'e}nez-Bravo}, \bibinfo{author}{{\'A}.~L. Murciego}, \bibinfo{author}{A.~S. Mendes}, \bibinfo{author}{H.~S. San~Bl{\'a}s}, \bibinfo{author}{J.~Bajo},
\newblock \bibinfo{title}{Multi-object tracking in traffic environments: A systematic literature review},
\newblock \bibinfo{journal}{Neurocomputing} \bibinfo{volume}{494} (\bibinfo{year}{2022}) \bibinfo{pages}{43--55}.
\bibitem[{Bisio et~al.(2022)Bisio, Garibotto, Haleem, Lavagetto, and Sciarrone}]{bisio2022systematic}
\bibinfo{author}{I.~Bisio}, \bibinfo{author}{C.~Garibotto}, \bibinfo{author}{H.~Haleem}, \bibinfo{author}{F.~Lavagetto}, \bibinfo{author}{A.~Sciarrone},
\newblock \bibinfo{title}{A systematic review of drone based road traffic monitoring system},
\newblock \bibinfo{journal}{Ieee Access} \bibinfo{volume}{10} (\bibinfo{year}{2022}) \bibinfo{pages}{101537--101555}.
\bibitem[{Markiewicz et~al.(2017)Markiewicz, D{\l}ugosz, and Skruch}]{markiewicz2017review}
\bibinfo{author}{P.~Markiewicz}, \bibinfo{author}{M.~D{\l}ugosz}, \bibinfo{author}{P.~Skruch},
\newblock \bibinfo{title}{Review of tracking and object detection systems for advanced driver assistance and autonomous driving applications with focus on vulnerable road users sensing},
\newblock in: \bibinfo{booktitle}{Polish Control Conference}, \bibinfo{organization}{Springer}, \bibinfo{year}{2017}, pp. \bibinfo{pages}{224--237}.
\bibitem[{Premachandra et~al.(2020)Premachandra, Ueda, and Suzuki}]{premachandra2020detection}
\bibinfo{author}{C.~Premachandra}, \bibinfo{author}{S.~Ueda}, \bibinfo{author}{Y.~Suzuki},
\newblock \bibinfo{title}{Detection and tracking of moving objects at road intersections using a 360-degree camera for driver assistance and automated driving},
\newblock \bibinfo{journal}{IEEE Access} \bibinfo{volume}{8} (\bibinfo{year}{2020}) \bibinfo{pages}{135652--135660}.
\bibitem[{Cho and Cho(2022)}]{cho2022autonomous}
\bibinfo{author}{K.~Cho}, \bibinfo{author}{D.~Cho},
\newblock \bibinfo{title}{Autonomous driving assistance with dynamic objects using traffic surveillance cameras},
\newblock \bibinfo{journal}{Applied Sciences} \bibinfo{volume}{12} (\bibinfo{year}{2022}) \bibinfo{pages}{6247}.
\bibitem[{Petrovskaya and Thrun(2009)}]{petrovskaya2009model}
\bibinfo{author}{A.~Petrovskaya}, \bibinfo{author}{S.~Thrun},
\newblock \bibinfo{title}{Model based vehicle detection and tracking for autonomous urban driving},
\newblock \bibinfo{journal}{Autonomous Robots} \bibinfo{volume}{26} (\bibinfo{year}{2009}) \bibinfo{pages}{123--139}.
\bibitem[{Muller et~al.(2009)Muller, Manz, Himmelsbach, and Wunsche}]{muller2009model}
\bibinfo{author}{A.~Muller}, \bibinfo{author}{M.~Manz}, \bibinfo{author}{M.~Himmelsbach}, \bibinfo{author}{H.~Wunsche},
\newblock \bibinfo{title}{A model-based object following system},
\newblock in: \bibinfo{booktitle}{2009 IEEE Intelligent Vehicles Symposium}, \bibinfo{organization}{IEEE}, \bibinfo{year}{2009}, pp. \bibinfo{pages}{242--249}.
\bibitem[{Rangesh and Trivedi(2019)}]{rangesh2019no}
\bibinfo{author}{A.~Rangesh}, \bibinfo{author}{M.~M. Trivedi},
\newblock \bibinfo{title}{No blind spots: Full-surround multi-object tracking for autonomous vehicles using cameras and lidars},
\newblock \bibinfo{journal}{IEEE Transactions on Intelligent Vehicles} \bibinfo{volume}{4} (\bibinfo{year}{2019}) \bibinfo{pages}{588--599}.
\bibitem[{G{\'o}mez-Hu{\'e}lamo et~al.(2021)G{\'o}mez-Hu{\'e}lamo, Bergasa, Guti{\'e}rrez, Arango, and D{\'\i}az}]{gomez2021smartmot}
\bibinfo{author}{C.~G{\'o}mez-Hu{\'e}lamo}, \bibinfo{author}{L.~M. Bergasa}, \bibinfo{author}{R.~Guti{\'e}rrez}, \bibinfo{author}{J.~F. Arango}, \bibinfo{author}{A.~D{\'\i}az},
\newblock \bibinfo{title}{Smartmot: Exploiting the fusion of hdmaps and multi-object tracking for real-time scene understanding in intelligent vehicles applications},
\newblock in: \bibinfo{booktitle}{2021 IEEE Intelligent Vehicles Symposium (IV)}, \bibinfo{organization}{IEEE}, \bibinfo{year}{2021}, pp. \bibinfo{pages}{710--715}.
\bibitem[{Richards and Papanikolopoulos(1997)}]{richards1997detection}
\bibinfo{author}{C.~A. Richards}, \bibinfo{author}{N.~P. Papanikolopoulos},
\newblock \bibinfo{title}{Detection and tracking for robotic visual servoing systems},
\newblock \bibinfo{journal}{Robotics and Computer-Integrated Manufacturing} \bibinfo{volume}{13} (\bibinfo{year}{1997}) \bibinfo{pages}{101--120}.
\bibitem[{Jacques et~al.(2005)Jacques, Rodrigo, McIsaac, and Samarabandu}]{jacques2005object}
\bibinfo{author}{D.~J. Jacques}, \bibinfo{author}{R.~Rodrigo}, \bibinfo{author}{K.~A. McIsaac}, \bibinfo{author}{J.~Samarabandu},
\newblock \bibinfo{title}{An object tracking and visual servoing system for the visually impaired},
\newblock in: \bibinfo{booktitle}{Proceedings of the 2005 IEEE International Conference on Robotics and Automation}, \bibinfo{organization}{IEEE}, \bibinfo{year}{2005}, pp. \bibinfo{pages}{3510--3515}.
\bibitem[{Chaumette and Hutchinson(2008)}]{chaumette2008visual}
\bibinfo{author}{F.~Chaumette}, \bibinfo{author}{S.~Hutchinson},
\newblock \bibinfo{title}{Visual servoing and visual tracking},
\newblock \bibinfo{journal}{Handbook of Robotics}  (\bibinfo{year}{2008}) \bibinfo{pages}{563--583}.
\bibitem[{Dixon et~al.(2002)Dixon, Zergeroglu, Fang, and Dawson}]{dixon2002object}
\bibinfo{author}{W.~E. Dixon}, \bibinfo{author}{E.~Zergeroglu}, \bibinfo{author}{Y.~Fang}, \bibinfo{author}{D.~M. Dawson},
\newblock \bibinfo{title}{Object tracking by a robot manipulator: a robust cooperative visual servoing approach},
\newblock in: \bibinfo{booktitle}{Proceedings 2002 IEEE International Conference on Robotics and Automation (Cat. No. 02CH37292)}, volume~\bibinfo{volume}{1}, \bibinfo{organization}{IEEE}, \bibinfo{year}{2002}, pp. \bibinfo{pages}{211--216}.
\bibitem[{Xu et~al.(2022)Xu, Chen, Ou, Wang, and Yang}]{xu2022learning}
\bibinfo{author}{S.~Xu}, \bibinfo{author}{K.~Chen}, \bibinfo{author}{Y.~Ou}, \bibinfo{author}{Z.~Wang}, \bibinfo{author}{C.~Yang},
\newblock \bibinfo{title}{A learning-based object tracking strategy using visual sensors and intelligent robot arm},
\newblock \bibinfo{journal}{IEEE Transactions on Automation Science and Engineering} \bibinfo{volume}{20} (\bibinfo{year}{2022}) \bibinfo{pages}{2280--2293}.
\bibitem[{Ortenzi et~al.(2021)Ortenzi, Cosgun, Pardi, Chan, Croft, and Kuli{\'c}}]{ortenzi2021object}
\bibinfo{author}{V.~Ortenzi}, \bibinfo{author}{A.~Cosgun}, \bibinfo{author}{T.~Pardi}, \bibinfo{author}{W.~P. Chan}, \bibinfo{author}{E.~Croft}, \bibinfo{author}{D.~Kuli{\'c}},
\newblock \bibinfo{title}{Object handovers: a review for robotics},
\newblock \bibinfo{journal}{IEEE Transactions on Robotics} \bibinfo{volume}{37} (\bibinfo{year}{2021}) \bibinfo{pages}{1855--1873}.
\bibitem[{Costanzo et~al.(2021)Costanzo, De~Maria, and Natale}]{costanzo2021handover}
\bibinfo{author}{M.~Costanzo}, \bibinfo{author}{G.~De~Maria}, \bibinfo{author}{C.~Natale},
\newblock \bibinfo{title}{Handover control for human-robot and robot-robot collaboration},
\newblock \bibinfo{journal}{Frontiers in Robotics and AI} \bibinfo{volume}{8} (\bibinfo{year}{2021}) \bibinfo{pages}{672995}.
\bibitem[{Bouget et~al.(2017)Bouget, Allan, Stoyanov, and Jannin}]{bouget2017vision}
\bibinfo{author}{D.~Bouget}, \bibinfo{author}{M.~Allan}, \bibinfo{author}{D.~Stoyanov}, \bibinfo{author}{P.~Jannin},
\newblock \bibinfo{title}{Vision-based and marker-less surgical tool detection and tracking: a review of the literature},
\newblock \bibinfo{journal}{Medical Image Analysis} \bibinfo{volume}{35} (\bibinfo{year}{2017}) \bibinfo{pages}{633--654}. \DOIprefix\doi{10.1016/j.media.2016.09.003}.
\bibitem[{Nwoye and Padoy(2025)}]{nwoye2025surgitrack}
\bibinfo{author}{C.~I. Nwoye}, \bibinfo{author}{N.~Padoy},
\newblock \bibinfo{title}{Surgitrack: Fine-grained multi-class multi-tool tracking in surgical videos},
\newblock \bibinfo{journal}{Medical Image Analysis} \bibinfo{volume}{101} (\bibinfo{year}{2025}) \bibinfo{pages}{103438}.
\bibitem[{Li et~al.(2023)Li, Shu, Liang, Goodridge, Sahu, Creighton, Taylor, and Unberath}]{li2023tatoo}
\bibinfo{author}{Z.~Li}, \bibinfo{author}{H.~Shu}, \bibinfo{author}{R.~Liang}, \bibinfo{author}{A.~Goodridge}, \bibinfo{author}{M.~Sahu}, \bibinfo{author}{F.~X. Creighton}, \bibinfo{author}{R.~H. Taylor}, \bibinfo{author}{M.~Unberath},
\newblock \bibinfo{title}{Tatoo: vision-based joint tracking of anatomy and tool for skull-base surgery},
\newblock \bibinfo{journal}{International journal of computer assisted radiology and surgery} \bibinfo{volume}{18} (\bibinfo{year}{2023}) \bibinfo{pages}{1303--1310}.
\bibitem[{Martin-Gomez et~al.(2023)Martin-Gomez, Li, Song, Yang, Wang, Ding, Navab, Zhao, and Armand}]{martin2023sttar}
\bibinfo{author}{A.~Martin-Gomez}, \bibinfo{author}{H.~Li}, \bibinfo{author}{T.~Song}, \bibinfo{author}{S.~Yang}, \bibinfo{author}{G.~Wang}, \bibinfo{author}{H.~Ding}, \bibinfo{author}{N.~Navab}, \bibinfo{author}{Z.~Zhao}, \bibinfo{author}{M.~Armand},
\newblock \bibinfo{title}{Sttar: surgical tool tracking using off-the-shelf augmented reality head-mounted displays},
\newblock \bibinfo{journal}{IEEE Transactions on Visualization and Computer Graphics}  (\bibinfo{year}{2023}).
\bibitem[{Singh et~al.(2020)Singh, Salehi, and Gholipour}]{singh2020deep}
\bibinfo{author}{A.~Singh}, \bibinfo{author}{S.~S.~M. Salehi}, \bibinfo{author}{A.~Gholipour},
\newblock \bibinfo{title}{Deep predictive motion tracking in magnetic resonance imaging: application to fetal imaging},
\newblock \bibinfo{journal}{IEEE transactions on medical imaging} \bibinfo{volume}{39} (\bibinfo{year}{2020}) \bibinfo{pages}{3523--3534}.
\bibitem[{Koniar et~al.(2017)Koniar, Harga{\v{s}}, Loncova, Simonova, Ducho{\v{n}}, and Be{\v{n}}o}]{koniar2017visual}
\bibinfo{author}{D.~Koniar}, \bibinfo{author}{L.~Harga{\v{s}}}, \bibinfo{author}{Z.~Loncova}, \bibinfo{author}{A.~Simonova}, \bibinfo{author}{F.~Ducho{\v{n}}}, \bibinfo{author}{P.~Be{\v{n}}o},
\newblock \bibinfo{title}{Visual system-based object tracking using image segmentation for biomedical applications},
\newblock \bibinfo{journal}{Electrical Engineering} \bibinfo{volume}{99} (\bibinfo{year}{2017}) \bibinfo{pages}{1349--1366}.
\bibitem[{Hayashida et~al.(2022)Hayashida, Nishimura, and Bise}]{hayashida2022consistent}
\bibinfo{author}{J.~Hayashida}, \bibinfo{author}{K.~Nishimura}, \bibinfo{author}{R.~Bise},
\newblock \bibinfo{title}{Consistent cell tracking in multi-frames with spatio-temporal context by object-level warping loss},
\newblock in: \bibinfo{booktitle}{Proceedings of the IEEE/CVF Winter Conference on Applications of Computer Vision}, \bibinfo{year}{2022}, pp. \bibinfo{pages}{1727--1736}.

\end{thebibliography}

\end{document}